\documentclass{article}

\usepackage[preprint,nonatbib]{neurips_data_2024}

\usepackage[utf8]{inputenc} 
\usepackage[T1]{fontenc}    
\usepackage{url}            
\usepackage{booktabs}       
\usepackage{amsfonts}       
\usepackage{nicefrac}       
\usepackage{microtype}      
\usepackage{xcolor}         
\usepackage{cite}

\usepackage{placeins}
\usepackage{wrapfig}
\usepackage{algorithm}
\usepackage{algorithmic}
\usepackage{amssymb}
\usepackage{amsmath}
\usepackage{multirow}
\usepackage{caption}
\usepackage{subcaption}
\usepackage{pifont}
\usepackage{makecell}
\usepackage{graphicx}

\definecolor{cvprblue}{rgb}{0.21,0.49,0.74}
\usepackage[pagebackref,breaklinks,colorlinks,citecolor=cvprblue]{hyperref}

\newcommand{\eg}{\textit{e.g.}}
\newcommand{\ie}{\textit{i.e.}}

\title{OpenCIL: Benchmarking Out-of-Distribution Detection in Class-Incremental Learning}

\author{%
  Wenjun Miao\textsuperscript{1}, 
  Guansong Pang\textsuperscript{2, *}, 
  Trong-Tung Nguyen\textsuperscript{2},  
  Ruohang Fang\textsuperscript{1}, \\
  \textbf{Jin Zheng\textsuperscript{1,}\thanks{Corresponding authors: G. Pang and J. Zheng\\\indent\indent T.T. Nguyen's participation was done when visiting Singapore Management University.} , }
  \textbf{Xiao Bai\textsuperscript{1,3}} \\
  \textsuperscript{1}School of Computer Science and Engineering, Beihang University \\
  \textsuperscript{2}School of Computing and Information Systems, Singapore Management University \\
  \textsuperscript{3}Jiangxi Research Institute, Beihang University \\
  \texttt{$\{$miaowenjun, jinzheng, baixiao$\}$@buaa.edu.cn, } \\
  \texttt{gspang@smu.edu.sg, v.tungnt132@vinai.io} \\
}

\begin{document}
\begin{sloppypar}

\maketitle

\begin{abstract}
  Class incremental learning (CIL) aims to learn a model that can not only incrementally accommodate new classes, but also maintain the learned knowledge of old classes. 
  Out-of-distribution (OOD) detection in CIL is to retain this incremental learning ability, while being able to reject unknown samples that are drawn from different distributions of the learned classes. 
  This capability is crucial to the safety of deploying CIL models in open worlds.
  However, despite remarkable advancements in the respective CIL and OOD detection, there lacks a systematic and large-scale benchmark to assess the capability of advanced CIL models in detecting OOD samples.
  To fill this gap, in this study we design a comprehensive empirical study to establish such a benchmark, named \textbf{OpenCIL}. 
  To this end, we propose two principled frameworks for enabling four representative CIL models with 15 diverse OOD detection methods, resulting in 60 baseline models for OOD detection in CIL. 
  The empirical evaluation is performed on two popular CIL datasets with six commonly-used OOD datasets.
  One key observation we find through our comprehensive evaluation is that the CIL models can be severely biased towards the OOD samples and newly added classes when they are exposed to open environments. 
  Motivated by this, we further propose a new baseline for OOD detection in CIL, namely Bi-directional Energy Regularization (\textbf{BER}), which is specially designed to mitigate these two biases in different CIL models by having energy regularization on both old and new classes. Its superior performance is justified in our experiments.
  All codes and datasets are open-source at \renewcommand\UrlFont{\color{blue}}\url{https://github.com/mala-lab/OpenCIL}.

\end{abstract}

\section{Introduction}
Training of deep neural networks (DNNs) heavily rely on large-scale collected data of a fixed set of classes \cite{russakovsky2015imagenet, krizhevsky2017imagenet}, but data in real-world applications is constantly changing, leading to continuous new classes in training data.
Continual learning (CL) enables DNNs to continuously learn a sequence of tasks, with each task consisting of a set of unique classes. 
The samples for the learned/old tasks are assumed to be not accessible in such dynamic environments. Class-incremental learning (CIL) is one type of CL where task identifiers are not known at testing time.
Compared to another type of CL, task-incremental learning, CIL is often considered a more practical setting in real applications \cite{rebuffi2017icarl, li2017learning}. 
Thus, we focus on CIL in this study.

\begin{figure}[t!]
  \centering
  \subfloat[]
  {\includegraphics[width=0.30\textwidth]{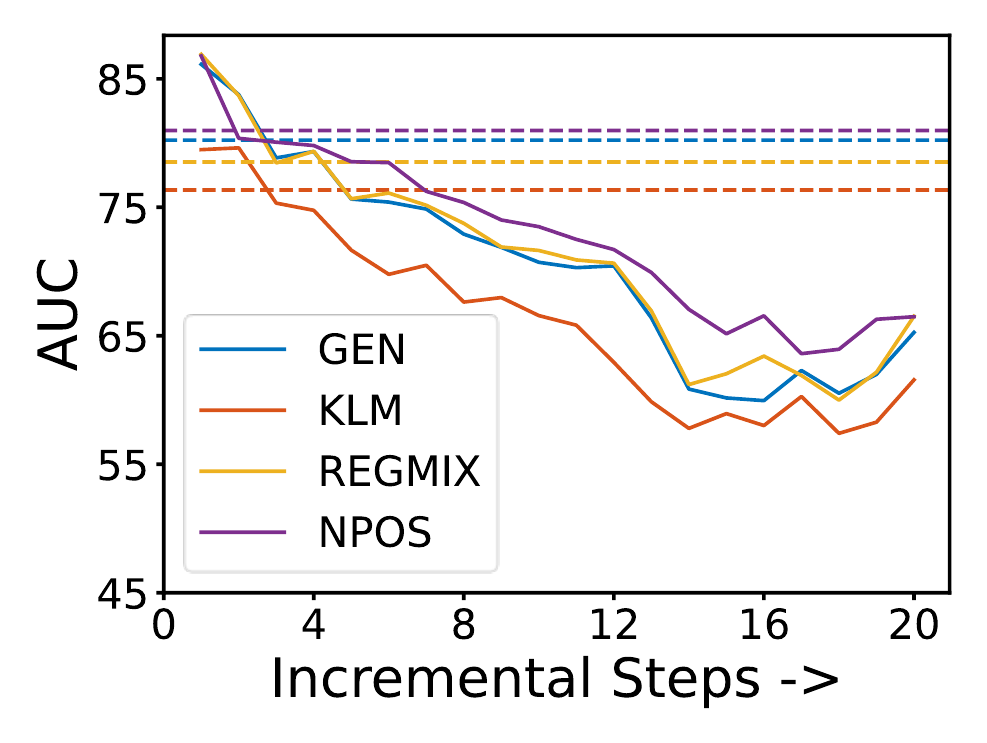} \label{Incremental}}
  \subfloat[]
  {\includegraphics[width=0.30\textwidth]{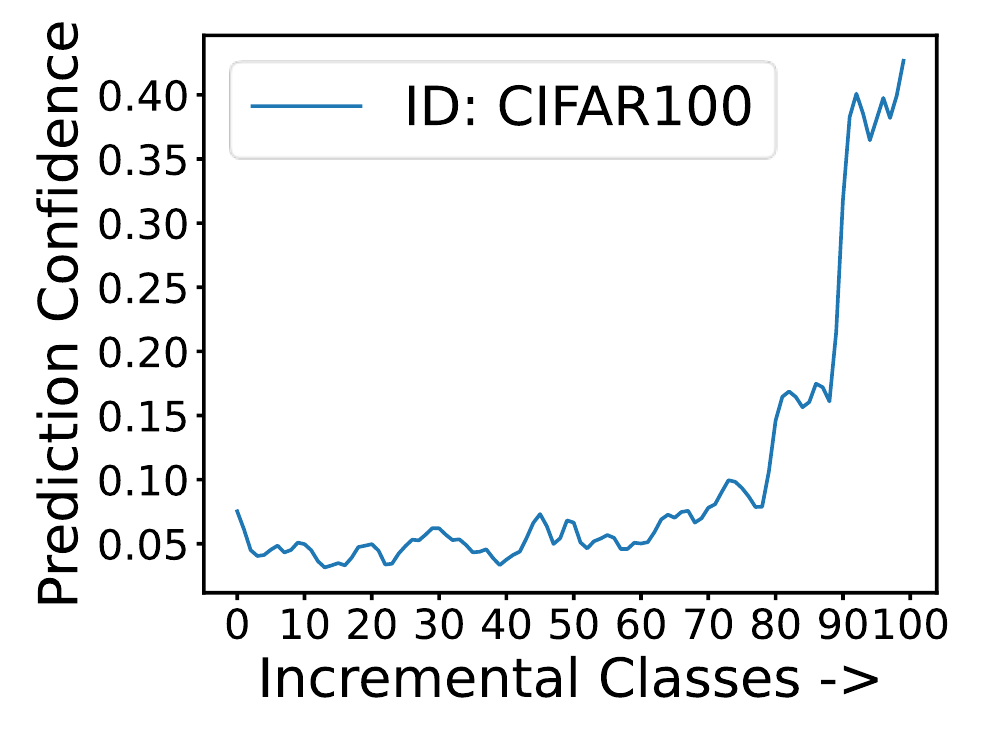}  \label{ID}}
  \subfloat[]
  {\includegraphics[width=0.30\textwidth]{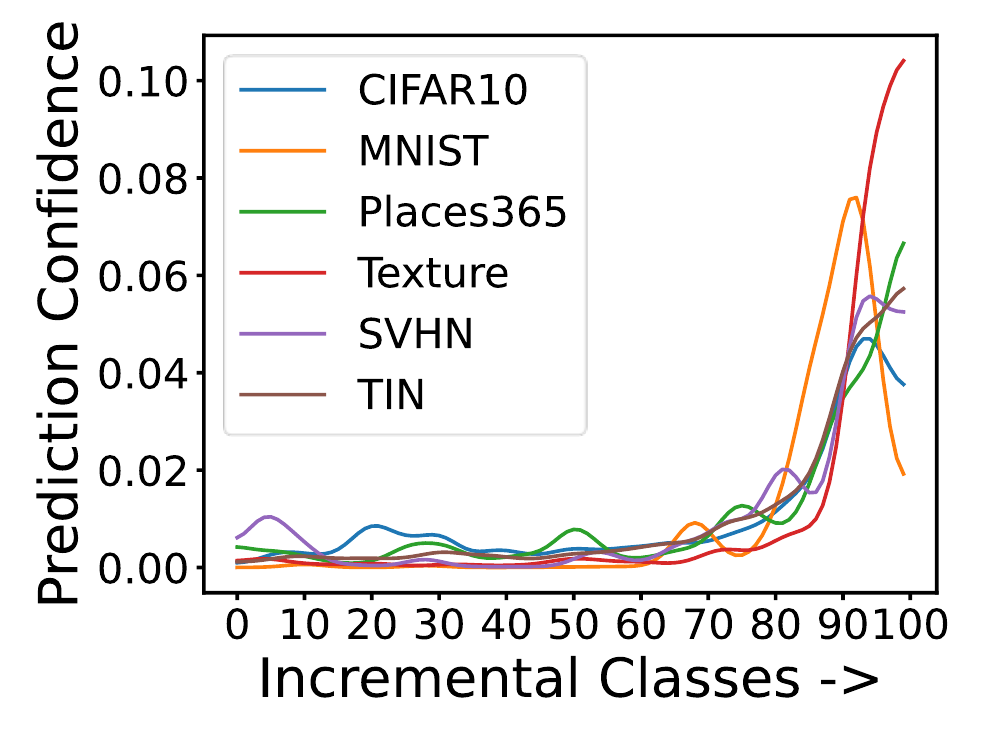}  \label{OOD}}
  \caption{Qualitative results of the CIL model iCaRL \cite{rebuffi2017icarl} with CIFAR100 \cite{krizhevsky2009learning}.
  \textbf{(a)} All four representative OOD detection methods experience a decreased AUC performance with increasing incremental steps, compared to themselves working on the full training data of all steps.
  \textbf{(b)} Mean prediction confidence of iCaRL on test samples from all incremental classes.
  \textbf{(c)} Mean prediction confidence of iCaRL classifying six OOD datasets into one of the ID classes.  
  The results for the other CIL models are provided in
  in the Appendix \ref{morequantity}.
  }
\end{figure}

Many CIL methods have been introduced over the years to overcome \textit{Catastrophic Forgetting} (CF) of the learned knowledge on old classes, \ie, degraded classification accuracy on old classes due to model updating on new classes \cite{mccloskey1989catastrophic}. 
These include regularization-based methods \cite{liu2021rmm, rebuffi2017icarl, xiao2023endpoints}, parameter-isolation-based methods \cite{xu2018reinforced, yan2021dynamically, wang2022foster}, and
replay-based methods \cite{rebuffi2017icarl, wu2019large, luo2023class,niu2024graph}. 
They have shown remarkable performance on the in-distribution (ID) classes in incremental tasks, but lack the capability to recognize and reject out-of-distribution (OOD) samples that are unseen during incremental learning \cite{huang2021mos, wang2020long} (\ie, no class overlapping between ID and OOD samples). 
Such a capability is crucial to the safety of deploying CIL models in open environments in real-world applications such as autonomous driving \cite{kendall2017uncertainties} and medical diagnosis \cite{leibig2017leveraging}.
On the other hand, numerous OOD detection methods have been proposed, which can be generally categorized into post hoc-based methods \cite{sun2021react, wang2023detecting, zhang2023decoupling} and fine-tuning-based methods \cite{liu2020energy, wei2022mitigating, tian2022pixel,yu2023block, li2023rethinking,liu2023residual,miao2023out,li2024learning,li2024negprompt}. 
However, these OOD methods are focused on static environments, where the samples of all tasks are accessible during training, making them ineffective in dynamic CIL environments, as shown in Fig. \ref{Incremental}.

There are benchmarking studies on the respective CIL and OOD detection areas, \eg, OpenOOD \cite{yang2022openood,zhang2023openood} for OOD detection and FACIL \cite{masana2022class} for CIL.
However, there lacks a systematic and large-scale benchmarking study to assess the synergy of existing state-of-the-art (SotA) CIL models and OOD detection methods.
To bridge this gap, we design a comprehensive performance benchmark for OOD detection in CIL, called \textbf{OpenCIL}, offering a unified evaluation protocol for different CIL models with diverse OOD detectors. 
In particular, OpenCIL accommodates four representative CIL models with 15 diverse OOD detection methods, resulting in 60 baseline models in the comparison study. 
To achieve this, we introduce two principled frameworks for incorporating post-hoc and fine-tuning-based OOD detection methods into three types of the aforementioned CIL models. 
It provides a module-based evaluation pipeline that enables easy examination of not only the OOD detection capability for different CIL models but also the ability of different OOD detectors in the presence of catastrophic forgetting. 
In terms of datasets, OpenCIL uses two popular CIL datasets CIFAR100 \cite{krizhevsky2009learning} and large-scale ImageNet1K \cite{russakovsky2015imagenet}, together with six commonly-used OOD datasets containing both near OOD datasets \cite{yang2022openood} and far OOD datasets \cite{yang2022openood}, thereby building a systematic benchmark from diverse CIL and OOD detection perspectives. 

One key observation we find in our experiments is that the CIL models have increasing biases towards OOD samples and newly added classes with the growth of incremental learning steps. 
The underlying reasons are two-folds. One main reason is that due to the CF problem, the CIL models often have lower prediction confidence for samples of old ID classes (\ie, classes seen in the old tasks), compared to new ID classes, as illustrated in Fig. \ref{ID}. 
This leads to a difficulty in differentiate between old ID class samples and OOD samples.
Furthermore, as illustrated in Fig. \ref{OOD}, the DNN-based CIL models typically exhibit over-confident prediction on OOD samples, causing the misclassification of the OOD samples into not only the old ID classes but also the new ID classes. 
Increasing the incremental learning steps results in larger classification semantic space and more severe CF, which continually amplifies the biases towards the OOD samples and newly added classes. 
To mitigate this issue, we propose a new baseline for OOD detection in CIL, namely Bi-directional Energy Regularization (\textbf{BER}), which jointly optimizes two energy regularization terms to modulate the energy prediction of OOD samples w.r.t. samples of old and new class samples respectively. 
This helps effectively reduce these two biases.
In summary, our main contributions are as follows:
\begin{itemize}
    \item \textbf{Benchmark. } We introduce the OpenCIL benchmark, the first benchmark for comprehensively evaluating the capability of CIL models in rejecting diverse OOD samples. In OpenCIL, we design two principled baseline frameworks and an evaluation protocol for evaluating 60 baselines on two popular CIL datasets and six near/far OOD datasets. This enables a systematic assessment of the synergy of SotA CIL and OOD detection methods.
    \item \textbf{Important observations.} We design and analyze a series of experiments in OpenCIL that leads to a number of important observations, offering crucial insights into the design of CIL models for open-world applications.
    \item \textbf{New baseline}. Beyond the 60 baselines that are synthesized from existing CIL and OOD detection methods, 
    we also introduce a new baseline, namely Bi-directional Energy Regularization (BER), which provides a new framework for mitigating increasing biases towards OOD samples and newly added classes when incorporating fine-tuning OOD detection methods into CIL models.
    \item \textbf{Code and datasets.} The code of all our baselines and the pre-processed datasets under our OpenCIL evaluation protocol are made publicly available at an open-source repository \renewcommand\UrlFont{\color{blue}}\url{https://github.com/mala-lab/OpenCIL}, from which users can easily reproduce our results and evaluate their own datasets/models with minimal effort.
\end{itemize}

\section{Related Work}
\label{related_work}

\noindent \textbf{Out-of-distribution (OOD) Detection. }
The objective of this task is to determine whether a given input sample belongs to the learned classes (in-distribution) or unknown classes (out-of-distribution). 
In recent years, OOD detection has been extensively developed, including post hoc-based methods \cite{sun2021react, wang2023detecting, zhang2023decoupling} and fine-tuning-based methods \cite{liu2020energy, wei2022mitigating, tian2022pixel,yu2023block, li2023rethinking,liu2023residual,miao2023out,li2024learning,li2024negprompt}. 
The post hoc methods focus on devising new OOD scoring functions in the inference stage.
The fine-tuning-based methods focus on separating OOD samples from ID samples by training a strong classifier as OOD detector. 
However, all these methods are applied to non-CIL models. There lacks exploration of their capability on CIL models, resulting in poor performance when there is catastrophic forgetting.
Our BER is a fine-tuning method that can be applied to different CIL models for improving their OOD detection performance.

\noindent \textbf{Class Incremental Learning (CIL). }
CIL performs the learning procedure in an incremental manner with growing data samples. It focuses on alleviating the catastrophic forgetting problem, in which the CIL models are required to remember the knowledge of the learned classes from old tasks while learning the discriminative information for the newly coming classes.
There are three main lines of work in this area \cite{luo2023class, xiao2023endpoints}. 
Regularization-based methods focus on applying discrepancy (between old and new models) as penalization terms in their objective functions \cite{liu2021rmm, rebuffi2017icarl, xiao2023endpoints}. 
Parameter-isolation-based methods aim to increase the model parameters in each new incremental step to prevent knowledge forgetting caused by parameter overwritten when learning new tasks \cite{xu2018reinforced, yan2021dynamically, wang2022foster}.
Replay-based algorithms assume there is a memory budget allowing a handful of old class examples in the memory. These memory examples can be used to re-train/fine-tune the CIL model in each new incremental step \cite{rebuffi2017icarl, wu2019large, luo2023class,niu2024graph}.
However, all these methods focus on tackling the CIL problem in a closed world, failing to take into account of distinguishing ID data from unknown samples (\eg, OOD data),
Our BER baselines can be applied to different pre-trained CIL models, in which all new class data and old class memory are used for fine-tuned a new OOD detector for the CIL models. It effectively equips the CIL models with significantly improved capability for OOD detection.

\section{Baselines and Evaluation Protocol}

\subsection{Problem Statement}
Our goal is to equip CIL models with a capability of rejecting OOD samples. Under this setting, a model is required to learn a sequence of tasks during training. 
At testing time, the model is used to classify samples into old/new classes, or recognize the samples as unknowns at each incremental step, without having access to the task identifiers. 
The model is evaluated based on its effectiveness in preventing the CF problem and distinguishing samples of old/new classes from OOD data.

Formally, CIL models are learned from a sequence of $c$ tasks ID data $T = \{T_1, T_2,..., T_c\}$. 
For each $t$-th $(1 \le t \le c)$ task, we have $T_t = (X_t^{train}, X_t^{test}, Y_t)$, where $X_t^{train}$ denotes the training ID data, $X_t^{test}$ denotes the testing ID data, and $Y_t$ denotes the classification semantic space consisting of a set of unique classes, \ie, the label spaces between any two incremental tasks have no class overlapping: when $i \neq j$, $Y_i \cap Y_j = \emptyset$. 
Thus, the set of all seen classes at task $t$ can be denoted as ${Q}_t = \cup_{i=1}^{t}Y_i$.
CIL assumes that data samples of the old classes are not accessible. 
For many CIL methods, such as the replay-based methods, they assume the availability of a memory buffer, in which a memory block is assigned to each task to store a very small set of samples for the task, \ie, $B_i \subseteq (X_i^{train}, Y_i)$ is a small subset of training ID data sampled from task $T_i$. Thus, we have the memory $M_{t} = \cup_{i=1}^{t-1}B_i$ that includes replay data from all $t$ tasks. Let $\theta_t{(\cdot)}$ be the CIL model at the incremental learning step $t$,
then the memory and the training data of task $t$ form the training ID data: $T_t^{train+} = (X_t^{train}, Y_t) \cup M_{t}$ for training $\theta_t{(\cdot)}$ at the learning step $t$. 
For replay-free CIL methods, $\theta_t{(\cdot)}$ is trained with $(X_t^{train}, Y_t)$ only.
During testing time at task $t$, let $T_t^{test} = X_1^{test} \cup X_2^{test} \cup ... \cup X_t^{test}$ be the testing ID data from all $t$ tasks, $\theta_t{(\cdot)}$ is used to classify samples in $T_t^{test}$ into one of the classes in ${Q}_t$. This is the setting for the conventional CIL problem. 

For OOD detection in CIL, in addition to $T_t^{test}$, the CIL model at task $t$, $\theta_t{(\cdot)}$, is also presented with an OOD dataset $X_t^{ood}$, which is a set of samples of unknown classes drawn from a different distribution as the ID data in $T_t$, \ie, the test data is $T_t^{test} \cup X_t^{ood}$.
Then given $x \in T_t^{test} \cup X_t^{ood}$, the goal of the CIL model $\theta_t{(\cdot)}$ is to either classify $x$ into the correct ID class from old/new tasks, or detect it as OOD data.

\begin{figure}[t!]
  \includegraphics[width=1.0\textwidth]{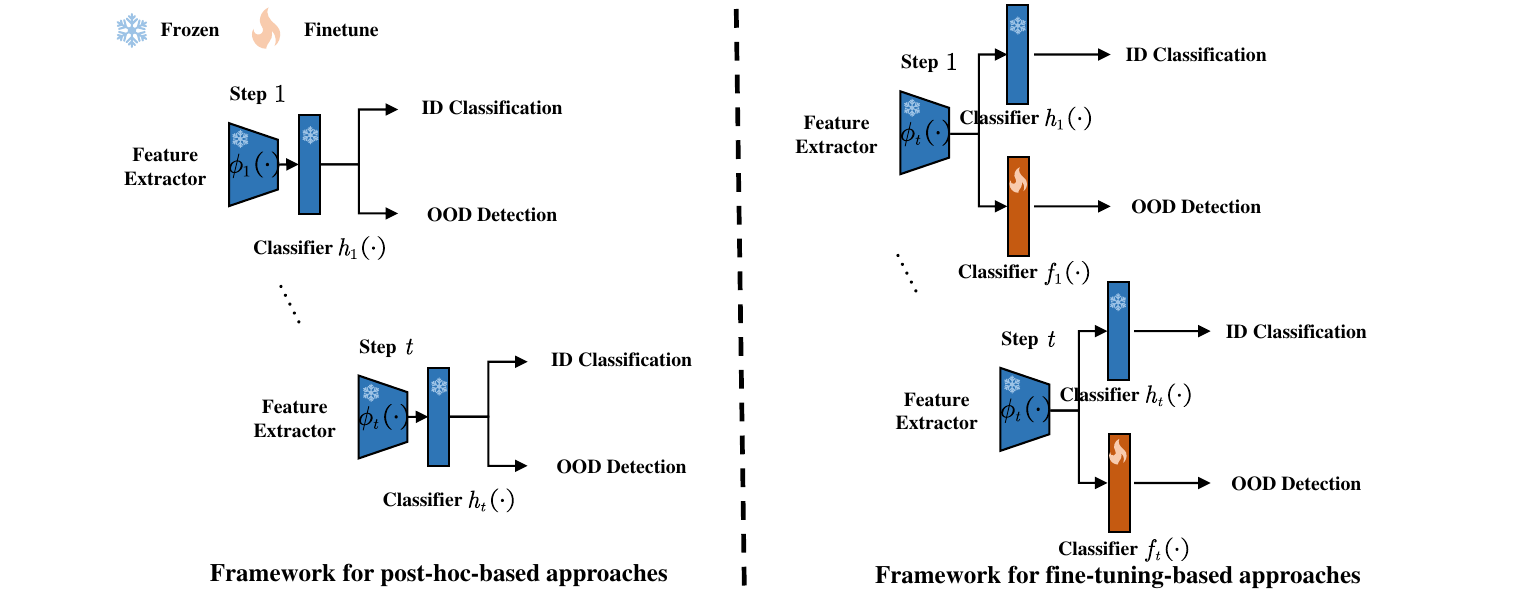}
  \caption{Two principled frameworks used in OpenCIL to incorporate OOD detection methods into the CIL models. Both frameworks are performed on pre-trained CIL models which are kept frozen when incorporating the OOD detection methods, and thus, their CIL performance is not affected.}
  \label{framework}
\end{figure}

\subsection{Baselines for OOD Detection in CIL} 
\label{baseline}
\paragraph{Adapting State-of-the-art OOD Detectors to CIL.}
There are two types of OOD detection methods: post-hoc and fine-tuning-based approaches, which exerts in the inference and training stages respectively.
Post-hoc methods use a statistic based on features or prediction logits derived from the pre-trained model as OOD score, which can be used for different pre-trained classification models. To adapt them to CIL models, we directly apply this kind of method to each incremental step of the CIL models for performing OOD detection, without having any effect on the performance of the CIL models.
Fine-tuning methods require the fine-tuning of part or all layers of the pre-trained CIL classifiers in each incremental step, which can intensify the catastrophic forgetting problem.
To avoid this issue, we freeze the feature extractor in the CIL models and utilize the fine-tuning OOD detection methods to fine-tune an extra classifier for OOD detection while maintaining the incremental learning accuracy.
These two types of methods are illustrated in Fig. \ref{framework} and summarized as follows:

\noindent
\hangafter=1
\setlength{\hangindent}{1.5em}
\textit{(1) Framework for post-hoc-based methods. }
A standard CIL model is first applied to learn the stream of ID data in an incremental manner to obtain the CIL model $\theta_t{(\cdot)}$, and then a post-hoc OOD scoring method is performed on the features/logits from the well-trained CIL model $\theta_t{(\cdot)}$ at each incremental step, as shown in Fig. \ref{framework} \textbf{Left}.

\noindent
\hangafter=1
\setlength{\hangindent}{1.5em}
\textit{(2) Framework for fine-tuning-based methods. }
As shown in Fig. \ref{framework} \textbf{Right}, the fine-tuning-based methods require the training of a standard CIL model $\theta_t{(\cdot)}$ in the first place, which contains a feature extractor $\phi_t{(\cdot)}$ and a classifier $h_t(\cdot)$, \ie, $\theta_t{(\cdot)}$ is a composition of $\phi_t{(\cdot)}$ and $h_t(\cdot)$.
Then we freeze both $\phi_t{(\cdot)}$ and $h_t(\cdot)$, and fine-tune an additional classifier $f_t(\cdot)$ only on top of $\phi_t{(\cdot)}$. 
This means that the fine-tuned CIL model contains the same feature extractor $\phi_t{(\cdot)}$ and classifier $h_t(\cdot)$ as the standard pre-trained CIL model for incremental ID classification, but it also has the additional fine-tuned classifier $f_t(\cdot)$.
Lastly, we use a post-hoc OOD scoring function on the classifier $f_t(\cdot)$ to calculate the OOD score.
Notably, the fine-tuning of $f_t(\cdot)$ is applied to each incremental step using only the training ID data $T_t^{train}$ at task $t$. In doing so, we perform these fine-tuning-based OOD detection methods to obtain better OOD scoring in each incremental step, which does not affect any of the learning procedure of the CIL model $\theta_t{(\cdot)}$ (\ie, $\phi_t{(\cdot)}$ and $h_t(\cdot)$).
Therefore, the CIL model $\theta_{t+1}{(\cdot)}$ is trained based on the $t$-th CIL classifier $h_t{(\cdot)}$, rather than the fine-tuned classifier $f_t{(\cdot)}$. 

With these two principled frameworks, different OOD detection methods can be easily incorporated into three different types of CIL models without impairing the incremental learning accuracy at all.

\paragraph{The Proposed Baseline: BER.} 
When establishing our OpenCIL benchmark, we found the increasing biases of the CIL models towards OOD samples and and newly added classes with the growth of incremental steps.
Therefore, we introduce the novel BER approach, a fine-tuning-based OOD detection approach, to tackle these bias issues, \ie, avoiding the classification of the old and new class samples as OOD samples.
BER consists of two components, namely New Task Energy Regularization (NTER) and Old Task Energy Regularization (OTER).
NTER is designed to distinguish OOD data from samples of new task classes, while OTER is designed to distinguish OOD data from samples of old task classes.
Below we introduce each component in detail.

\noindent\textit{New Task Energy Regularization (NTER). }
Due to the overwhelming presence of new class samples, CIL models typically demonstrate a strong biased prediction on the OOD samples towards the new classes, \ie, they have high prediction confidence on classifying OOD samples into the classes in the new task, as illustrated in Fig. \ref{OOD}.
To address this issue, NTER synthesizes the pseudo OOD samples that are distributed on the decision boundary of different new classes, and further utilizes them to enlarge the decision boundary margin.
Specifically, NTER first randomly mixes up samples of different new classes as the pseudo-OOD samples. Formally, let $\mathbf{x}_t = \{ x^1_t, x^2_t, ..., x^b_t, \}, \mathbf{x}_t \in X^{train}_t$ be one training batch of ID data from classes at task $t$, where task $t$ is the new task and $b$ is the batch size, $\mathbf{y}_t = \{ y^1_t, y^2_t, ..., y^b_t, \}$ and $ \mathbf{y}_t \in Y_t$ be the corresponding class label of $\mathbf{x}_t$,
then a pseudo-OOD sample $\bar{x}_t$ is synthesized as follows:
\begin{equation}\label{eq:mixup1}
    \bar{x}_t = \beta  x^i_t + (1 - \beta) x^j_t, y^i_t \neq y^j_t,
\end{equation}
where $\beta \in [0, 1]$ is sampled from Beta distribution.
Motivated by the success of energy-based methods \cite{liu2020energy}, we further utilize these pseudo-OOD samples to regularize the classification of new class samples via the follwoing energy loss function:
\begin{equation}
    \mathcal {L}_n = \mathbb E_{x_t \sim X^{train}_t} [(max(0,p_{in}-E(x_t)))^2] + \mathbb E_{(x_t^i,x_t^j) \sim X^{train}_t} [(max(0,E(\bar{x}_t)-p_{out}))^2], \label{inner}
\end{equation}
where $E(x;f) = -\tau \cdot log(\sum_{j=1}^{Q_t} e^{f_j(x)/\tau})$,
with $\tau$ be a temperature scaling hyperparameter.
Note that we do not combine all possible pairs of $(x_t^i,x_t^j)$ in the full training dataset $X^{train}_t$ to form $\bar{x}_t$, which are produced within the mini-batches for efficiency consideration \cite{zhou2021learning, verma2019manifold}. 
As a result, the time complexity is of the same magnitude as vanilla training, having minimal computational overhead.

\noindent\textit{Old Task Energy Regularization (OTER). }
Due to the catastrophic forgetting problem, as illustrated in Fig. \ref{ID}, the CIL models exhibit significantly lower prediction confidence on samples of old classes than new classes.
As a result, the CIL models often misclassify the old class samples as the OOD samples.
To address this issue, OTER performs a different mixup operation from NTER to generate more samples of old classes. In particular, it randomly mixes up new class samples with old class samples to synthesize old class samples.
This mixup operation not only increases the diversity of the small-sized old class samples but also transfers information from new class samples to old class samples. Then using these mixup samples in the fine-tuning stage largely enhances the prediction confidence of the old class samples and expand their decision boundary as well.
Formally, let $M_t = \{ m^1_t, m^2_t, ..., m^s_t \}$ be the samples of old classes in the memory bank for task $t$, and $s$ is the memory size,
then the synthesized old class samples $\bar{m_t}$ are generated via:
\begin{equation}
    \bar{m_t} = \lambda  x_t + (1 - \lambda) m_t,
\end{equation}
where $\lambda$ is a mixup hyperparameter, similar to $\beta$ in Eq. \ref{eq:mixup1}.
We further leverage these augmented ID samples to boost the prediction confidence of old class samples via the following energy loss function:
\begin{equation}
    \mathcal {L}_o = \mathbb E_{(x_t,m_t) \sim (X^{train}_t, M_t)} [(max(0,E(\bar{m}_t)-p_{in}))^2], \label{inter}
\end{equation}
where $E(x;f) = -\tau \cdot log(\sum_{j=1}^{Q_t} e^{f_j(x)/\tau})$.
Note that we only apply energy regularization to the ID data in OTER, without the energy regularization term on the pseudo-OOD samples in Eq. \ref{inner}.
The old class samples cannot be used for synthesizing the pseudo-OOD samples via, \eg, mixup between old and new class samples, or mixup between samples of different old classes.
This is because, as the samples stored in the data replay memory, these old task samples are the most representative samples of the old classes, which are typically located near the class center in their respective belonging classes in the feature space.
Therefore, generating pseudo-OOD samples using these old class samples will severely compress the decision boundary of old task classes, resulting in a significant adverse impact on the OOD detection performance.

Lastly, we utilize the cross-entropy loss, together with the two energy regularization losses, to fine-tune the extra classifier $f_t(\cdot)$ for in the CIL models following the fine-tuning OOD detection framework.
Thus, the overall optimization objective of our BER approach at each task $t$ is as follows:
\begin{equation}
    \mathcal {L} = \mathbb E_{(x,y) \sim (T^{train}_t, Q_t)} [\ell(f(x),y] + \alpha (\mathcal {L}_n + \mathcal {L}_o), \label{total}
\end{equation}
where $\ell$ is a cross-entropy loss, $\alpha$ is the hyperparameter, $\mathcal {L}_n$ is as defined in Eq. \ref{inner}, and $\mathcal {L}_o$ is as defined in Eq. \ref{inter}.
During inference, BER utilizes the Energy Score defined in \cite{liu2020energy} as OOD score.

\subsection{Evaluation Protocol}

\paragraph{CIL Datasets.}
Following \cite{rebuffi2017icarl, wang2022foster, wu2019large, wang2022beef, luo2023class}, we use two popular CIL datasets as the ID data in our benchmark: 1) \textbf{CIFAR100 \cite{krizhevsky2009learning}}, which consists of $50$K training images and $10$K testing images with $100$ classes, and 2) \textbf{large-scale ImageNet1k \cite{russakovsky2015imagenet}}, which consists of $1.28$M training images and $50$K testing images with $1,000$ classes. Besides, these two datasets are also widely used as ID datasets in the area of OOD detection.

Following \cite{rebuffi2017icarl, wang2022foster, wu2019large}, the splits of the two ID datasets are as follows.
1) \textbf{For CIFAR100}, we train the CIL model gradually with $k$ classes per incremental step with a fixed memory size of 2,000 exemplars. We respectively evaluate the performance with the step size $k\in\{5, 10, 20\}$.
2) \textbf{For ImageNet1K}, since it is a much larger dataset, we train the CIL model gradually with a larger number of $k$ classes per step with a fixed memory size of 20,000 exemplars. $k\in\{50, 100, 200\}$ is used for ImageNet1K.

\paragraph{OOD Datasets.} 
Following the recent large-scale solely OOD detection benchmark OpenOOD \cite{yang2022openood}, we select six datasets as OOD data for each ID dataset respectively. 1) For the ID dataset \textbf{CIFAR100}, the OOD data includes two near OOD datasets -- CIFAR10 \cite{krizhevsky2009learning} and Tiny-ImageNet (TIN) \cite{le2015tiny} -- and four far OOD datasets: MNIST  \cite{lecun2010mnist}, Texture \cite{cimpoi2014describing}, SVHN \cite{netzer2011reading} and Places365 \cite{zhou2017places}.
2) For \textbf{ImageNet1k}, the OOD data includes four near OOD datasets -- ImageNet\_O \cite{hendrycks2021natural}, iNaturalist \cite{van2018inaturalist}, OpenImage\_O \cite{wang2022vim} and Species \cite{basart2022scaling} -- and two far OOD datasets: MNIST \cite{lecun2010mnist} and Texture \cite{cimpoi2014describing}. 
\textbf{Near OOD datasets} means that their OOD samples have small semantic shift compared with the ID samples, while \textbf{far OOD datasets} means that their OOD samples are very different from the ID samples, typically containing obvious covariate (domain) shift \cite{yang2022openood}.

It is notable that, although these datasets have been widely used in the CIL and OOD detection community respectively, there is still no readily accessible and unified protocol for combining them into one experimental setting. 
Our OpenCIL benchmark offers one way to unify them into a systematic evaluation setting, facilitating the application of diverse OOD detectors under different CIL models.

\paragraph{Performance Metrics.}
To fairly compare the OOD detection performance among different incremental steps, we keep the ratio of testing OOD data $X_t^{ood}$ to testing ID data $T_t^{test}$ fixed at each incremental step $t$.
Specifically, we control the test OOD data fed into the CIL model in an incremental manner during the inference stage, increasing the number of testing OOD data $X_t^{ood}$ with increasing number of tasks proportionally.
Formally, the number of OOD samples for each OOD dataset at step $t$ is defined as:
\begin{equation}
    |X_t^{ood}| =  |X^{ood}| \times \frac{t}{|T|},
\end{equation}
where $ |X^{ood}|$ and $|X_t^{ood}|$ denote the number of OOD samples in the full OOD dataset and in the testing OOD data used at step $t$ respectively, and $T$ is the total number of CIL tasks/steps.
In this way, the number of testing ID and OOD data grows at the same speed, keeping the ratio of them identical across all incremental steps.

Following standard CIL evaluation \cite{rebuffi2017icarl}, the measure of \textit{average incremental accuracy} that reports the average classification accuracy over all incremental steps is used to evaluate the performance of CIL. At the same time, we also report the average results of three widely-used OOD detection performance metrics (AUC, FPR, AP) over all incremental steps.

More details about the datasets used and the performance metrics are provided in Appendix \ref{Benchmark_detail}.

\section{Results and Analysis}

We perform large-scale experiments that include the composition of $15$ OOD detection methods and four CIL models using three different incremental step sizes. Due to space limitation, we report the averaged results across all incremental steps at the step size of $k=10$ for CIFAR 100 and at the step size of $k=100$ for ImangeNet1K here. More fine-grained results at this step size and other two step sizes are presented in Appendix \ref{moreresults}.

\subsection{Implementation Details}
The 15 OOD detection methods include nine post hoc methods -- MSP \cite{hendrycks2016baseline}, ODIN \cite{liang2018enhancing}, Energy \cite{liu2020energy}, MaxLogit \cite{basart2022scaling}, GEN \cite{liu2023gen}, ReAct \cite{sun2021react}, KLM \cite{basart2022scaling}, Relation \cite{kim2023neural}, and NNGuide \cite{park2023nearest}  and six fine-tuning-based methods -- LogitNorm \cite{wei2022mitigating}, T2FNorm \cite{regmi2023t2fnorm}, AUGMIX\cite{hendrycks2020augmix}, REGMIX\cite{pinto2023RegMixup}, VOS\cite{du2022vos} and NPOS\cite{tao2023nonparametric}. The four CIL models include two regularization-based methods -- iCaRL \cite{rebuffi2017icarl} and WA \cite{zhao2020maintaining}, one replay-based method BiC\cite{wu2019large}, and one parameter-isolation-based method FOSTER\cite{wang2022foster}.

To ensure a fair comparison across methods originating from different areas, we use unified settings with common hyperparameters and architecture choices. 
For example, the backbone ResNet32 is used when CIFAR100 is used the ID dataset, and ResNet18 is used whenever ImageNet1K is the ID dataset. These two backbones are the most commonly used architecture for the respective ID dataset in the CIL community.
For the CIL pre-training, we employ default hyperparameters of these CIL models as stated in their original papers.
For OOD fine-tuning, following \cite{yang2022openood, zhang2023openood}, we use the common setting with SGD optimizer, using a learning rate of $0.1$, a momentum of $0.9$, a weight decay of $0.0005$, and adjusting the learning rate using a cosine annealing learning rate schedule.
The batch size is fixed at $128$ for all experiments.
We freeze the feature extractor, and fine-tune the extra classifier for $10$ epochs.
For our proposed BER baseline, following \cite{liu2020energy} , we set $\tau=1$, $\alpha=0.1$, $p_{in} = -5$ and $p_{out} = -27$ by default. $\lambda =0.002$ is used throughout the experiments.
All results are averaged over three independent runs using different random seeds.
All experiments are performed using $8$ NVIDIA RTX $3090$. More implementation details can be found in Appendix \ref{app:library}.

\subsection{Main Results}

\begin{table*}[t!]
\centering
\caption{Main results on OpenCIL benchmark. We report the average performance over six OOD datasets and all incremental steps. 
The \underline{\textbf{best}}, \textbf{second-best}, and \underline{third-best} performance per dataset are highlighted. Either the post hoc-based or fine-tuning-based OOD methods do not affect the original CIL performance, so their average CIL accuracy on the ID data remains the same.
}
\scalebox{0.6}{
\begin{tabular}{c|cc|cc|cc|cc|cc|cc|cc|cc}
\hline\hline
 & \multicolumn{8}{c|}{ID Dataset: CIFAR100} & \multicolumn{8}{c}{ID Dataset: ImageNet1K} \\
 \cline{2-17}
 & \multicolumn{2}{c|}{iCaRL \cite{rebuffi2017icarl}} & \multicolumn{2}{c|}{BiC\cite{wu2019large}}  & \multicolumn{2}{c|}{WA \cite{zhao2020maintaining}} & \multicolumn{2}{c|}{FOSTER\cite{wang2022foster}} & \multicolumn{2}{c|}{iCaRL \cite{rebuffi2017icarl}} & \multicolumn{2}{c|}{BiC\cite{wu2019large}}  & \multicolumn{2}{c|}{WA \cite{zhao2020maintaining}} & \multicolumn{2}{c}{FOSTER\cite{wang2022foster}} \\
 \cline{2-17}
 & AUC$\uparrow$ & FPR$\downarrow$ & AUC$\uparrow$ & FPR$\downarrow$ & AUC$\uparrow$ & FPR$\downarrow$ & AUC$\uparrow$ & FPR$\downarrow$ & AUC$\uparrow$ & FPR$\downarrow$ & AUC$\uparrow$ & FPR$\downarrow$ & AUC$\uparrow$ & FPR$\downarrow$ & AUC$\uparrow$ & FPR$\downarrow$ \\
 \hline
\multicolumn{11}{l}{\textbf{\textit{Average CIL accuracy}}} \\
 & \multicolumn{2}{c|}{60.08} & \multicolumn{2}{c|}{61.68} & \multicolumn{2}{c|}{65.88} & \multicolumn{2}{c|}{66.01} & \multicolumn{2}{c|}{44.44} & \multicolumn{2}{c|}{49.63} & \multicolumn{2}{c|}{52.22} & \multicolumn{2}{c}{52.29} \\   
\hline\hline
\multicolumn{11}{l}{\textbf{\textit{Post hoc-based OOD methods (w/o Fine-tuning)}}} \\
MSP \cite{hendrycks2016baseline} & 67.77 & 88.33 & 66.67 & 86.98 & 70.49 & 86.04 & 71.69 & 84.94 & 63.84 & 89.81 & 67.45 & 91.80 & 65.97 & 89.17 & 67.64 & 89.32 \\
ODIN \cite{liang2018enhancing} & 70.05 & 81.93 & 69.76 & 80.52 & 71.86 & 80.18 & 74.70 & 75.68 & \underline{68.41} & \textbf{87.91} & \textbf{71.09} & 87.45 & 67.86 & 89.08 & \underline{70.24} & \textbf{88.15} \\
Energy \cite{liu2020energy} & 70.35 & 83.74 & 69.76 & 82.01 & 73.03 & 81.22 & 75.76 & 77.24 & 65.64 & 91.88 & 70.16 & 89.57 & 66.57 & 92.22 & 69.39 & 90.49 \\
MaxLogit \cite{basart2022scaling} & 70.34 & 84.45 & 69.54 & 82.78 & 72.98 & 82.33 & 75.71 & 78.11 & 65.83 & 91.38 & 69.93 & 89.42 & 66.75 & 92.03 & 69.56 & 90.02 \\
GEN \cite{liu2023gen} & 70.86 & 84.31 & 69.87 & 82.51 & 73.30 & 81.42 & \underline{76.11} & 77.64 & 66.89 & 90.78 & 69.51 & 89.77 & \underline{68.55} & \textbf{88.54} & 69.74 & 89.91 \\
ReAct \cite{sun2021react} & 68.79 & 84.43 & 70.62 & 82.60 & 74.87 & 81.82 & \textbf{76.15} & 78.03 & 55.66 & 94.47 & 66.57 & 89.86 & 63.85 & 90.94 & 65.15 & 91.16 \\
KLM \cite{basart2022scaling}  & 67.51 & 88.00 & 66.27 & 87.66 & 69.89 & 87.65 & 71.34 & 85.90 & 66.48 & 88.28 & 67.32 & \textbf{87.17} & 65.24 & 89.59 & 70.23 & 88.45\\
Relation \cite{kim2023neural} & 64.40 & 89.57 & 65.48 & 80.85 & \textbf{76.12} & 88.37 & 72.17 & \underline{75.29} & 66.38 & 89.92 & 69.10 & 89.63 & 67.51 & 94.59 & 67.93 & 88.99 \\
NNGuide \cite{park2023nearest} & 72.14 & \underline{79.65} & 68.74 & \textbf{78.79} & \underline{75.60} & \textbf{76.00} & 75.90 & \textbf{74.70} & 66.53 & 89.88 & 70.64 & 87.33 & \textbf{68.69} & 88.82 & 67.63 & \underline{88.32} \\
\hline
Average & 69.13 & 84.93 & 68.52 & 82.74 & 73.13 & 82.78 & 74.39 & 78.61 & 65.07 & 90.48 & 69.09 & 89.11 & 66.78 & 90.55 & 68.61 & 89.42  \\ 
\hline\hline
\multicolumn{11}{l}{\textbf{\textit{Fine-tuning-based OOD methods (w/ Fine-tuning)}}}\\
LogitNorm \cite{wei2022mitigating} & 70.44 & 83.78 & 70.64 & 82.46 & 72.04 & 81.33 & 74.74 & 78.28 & 65.15 & 92.51 & 69.54 & \underline{87.26} & 66.30 & 93.16 & 67.84 & 92.39 \\
T2FNorm \cite{regmi2023t2fnorm} & 70.65 & 84.14 & \textbf{71.75} & 82.72 & 71.98 & 82.33 & 74.65 & 78.83 & 66.09 & 91.58 & 69.53 & 89.09 & 67.39 & 91.83 & 68.19 & 90.95 \\
AUGMIX \cite{hendrycks2020augmix}  & 70.72 & 83.91 & 70.78 & 82.94 & 72.11 & 81.53 & 74.75 & 78.62 & \textbf{68.60} & \underline{87.98} & \underline{70.83} & 87.85 & 68.10 & 89.37 & 69.62 & 88.53 \\
REGMIX \cite{pinto2023RegMixup} & 71.47 & 84.53 & \underline{71.43} & 82.23 & 72.01 & 82.95 & 75.26 & 79.35 & 66.41 & 90.69 & 69.83 & 89.68 & 67.45 & 90.69 & 69.43 & 90.78 \\
VOS \cite{du2022vos} & \underline{72.31} & 82.53 & 66.66 & 79.03 & 68.49 & 78.15 & 74.58 & 77.71 & 66.28 & 91.67 & 66.67 & 89.47 & 65.41 & 89.92 & 68.98 & 88.74 \\
NPOS \cite{tao2023nonparametric} & \textbf{73.23} & \textbf{79.50} & 67.03 & \underline{78.83} & 74.25 & \underline{76.78} & 75.91 & 76.00 & 68.12 & 88.74 & 70.42 & 88.36 & 68.00 & \underline{88.60} & \textbf{70.32} & 88.40 \\
\textbf{BER (Ours)} & \underline{\textbf{74.17}} & \underline{\textbf{78.87}} & \underline{\textbf{71.83}} & \underline{\textbf{77.76}} & \underline{\textbf{76.77}} & \underline{\textbf{75.58}} & \underline{\textbf{76.40}} & \underline{\textbf{74.58}} & \underline{\textbf{68.80}} & \underline{\textbf{87.51}} & \underline{\textbf{71.94}} & \underline{\textbf{86.80}} & \underline{\textbf{69.06}} & \underline{\textbf{87.59}} & \underline{\textbf{70.54}} & \underline{\textbf{87.44}} \\\hline
Average & 71.86 & 82.47 & 70.02 & 80.85 & 72.52 & 79.81 & 75.18 & 77.62 & 67.06 & 90.10 & 69.82 & 88.36 & 67.39 & 90.17 & 69.27 & 89.60 \\
\hline\hline
Average (All) & 70.33 & 83.85 & 69.18 & 81.92 & 72.86 & 81.48 & 74.74 & 78.18 & 65.94 & 90.31 & 69.41 & 88.78 & 67.04 & 90.38 & 68.90 & 89.50 \\
\hline\hline
\end{tabular}}
\label{Main}
\end{table*}

\begin{figure}[t!]
  \centering
  \subfloat[]
  {\includegraphics[width=0.4\textwidth]{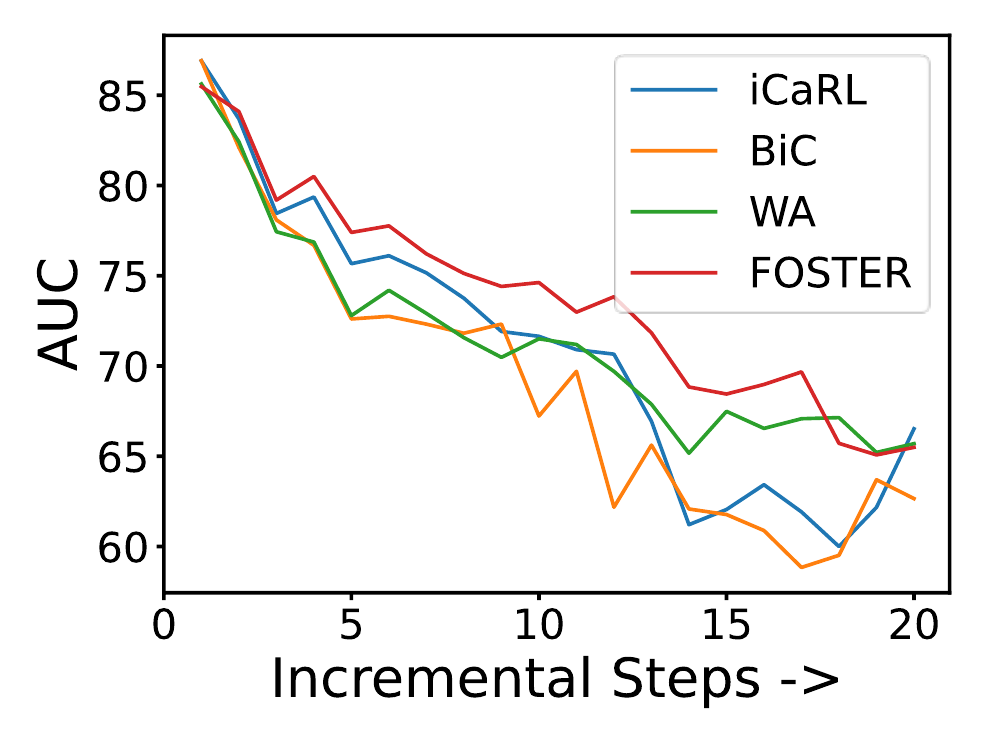}   \label{low_bound}}
  \subfloat[]
  {\includegraphics[width=0.4\textwidth]{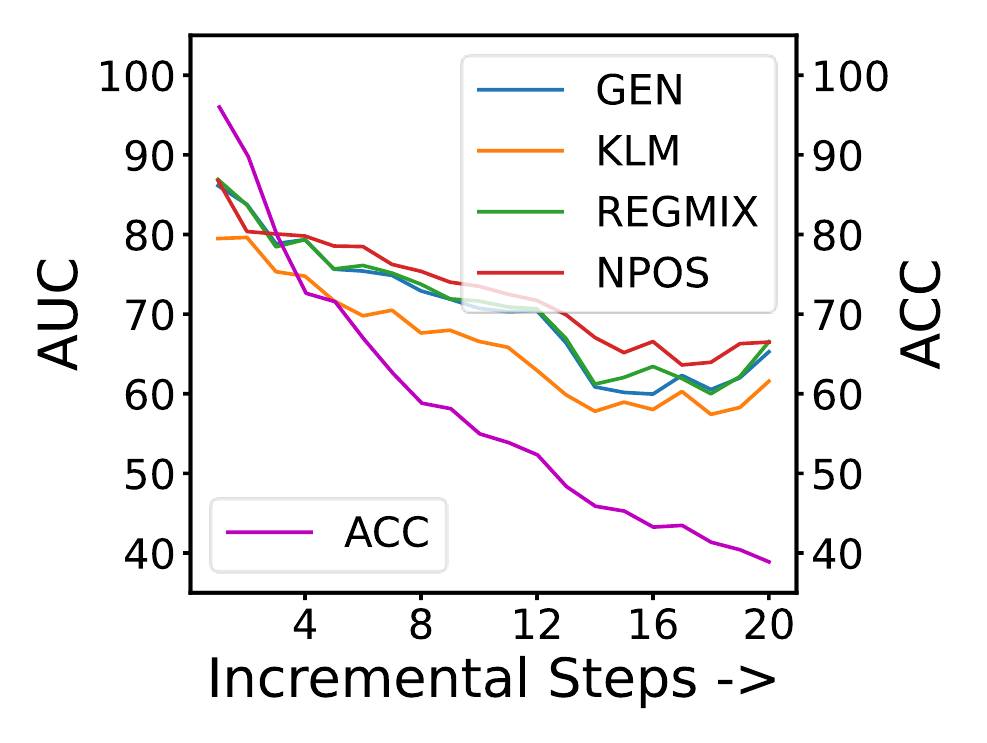}   \label{ACC}}
  \caption{
  \textbf{(a)} Average performance of CIL models with the OOD detector REGMIX \cite{pinto2023RegMixup} on six OOD datasets at each incremental step on CIFAR100. The results for the other OOD methods are provided in the Appendix \ref{morequantity}.
  \textbf{(b)} Average performance of four representative OOD methods on six OOD datasets at each incremental step, where the CIL model iCaRL \cite{rebuffi2017icarl} is used. ACC is the accuracy of iCaRL on CIFAR100 at each step. The results for the other three CIL models are provided in the Appendix \ref{morequantity}.
  }
\end{figure}

\paragraph{CIL Models with Higher CIL accuracy Not Necessarily Have Better OOD Detection Performance.} 
As shown in Table \ref{Main}, averaged over all 15 OOD methods, the CIL models with higher CIL accuracy, \eg, WA and FOSTER, often achieve better AUC and FPR performance than the other two CIL models on CIFAR100 (this is generally consistent with the performance at each incremental step, as shown in Fig. \ref{low_bound}), but this does not hold on the large ID dataset ImageNet1K.
This may be because better CIL algorithms can keep more essential information about ID data to improve ID classification, which can also prevent OOD data from being misclassified into ID classes; the latter case is mainly due to the poor accuracy for all CIL models.

\paragraph{CIL Models with Fine-tuning-based OOD Methods often Outperform than that with Post Hoc-based Methods.} This can be observed by looking at the average AUC and FPR results of each CIL model for the post hoc-based and the fine-tuning methods in Table \ref{Main}. The observation holds for both ID datasets. This demonstrates the advantage of tuning an additional classifier for OOD detection, but it is at the expense of some computational overhead.

\paragraph{CIL Models Exhibit Stronger Prediction Confidence on OOD Samples than Old Class Samples. } 
As shown in Figs. \ref{ID} and \ref{OOD}, the ID samples from old classes/tasks often have lower prediction confidence than different OOD samples, thereby being misclassified as the OOD samples. This phenomenon become more severe with increasing number of incremental steps. 
This is because the CIL models tend to be less confident when predicting the samples of old classes due to the catastrophic forgetting problem, making their OOD scores lower than samples from the OOD data. 

\paragraph{CIL Models often Misclassify OOD Samples into New Classes. } 
As also shown in Fig. \ref{OOD}, the CIL models often yield significantly higher prediction confidence on misclassifying the OOD samples into the new classes than the old classes, due to the presence of large samples from these new classes at each incremental step.
The reason is similar to the case that the CIL models tend to predict the ID samples of the new classes with higher confidence than the old classes.

\paragraph{Energy Regularization on old and new classes helps mitigate the biases towards OOD samples and new classes. }
This observation is justified by the superiority OOD detection performance of our proposed baseline BER in Table \ref{Main}, where BER consistently achieves the best performance in the averaged results in all cases on both CIFAR100 and ImageNet1K. More specifically, as shown in 
Table \ref{Ablation}, either New Task Energy Regularization (NTER) or Old Task Energy Regularization (OTER) helps boost the AUC and reduce the FPR on both near and far OOD data datasets. Since NTER is designed to alleviate the misclassificaiton of OOD samples into new classes while OTER is designed to reduce the misclassification of old class samples as OOD samples, their combination results in a detection model that largely reduces the overall detection errors.

\begin{table*}[t!]
\centering
\caption{Ablation study of our BER baseline on CIFAR100. Energy \cite{liu2020energy} is used as the baseline that does not use both NTER and OTER.} 
\scalebox{0.70}{
\begin{tabular}{cc|cc|cc|cc|cc|cc|cc}
\hline
\multicolumn{2}{c|}{\multirow{2}{*}{Loss Component}} & \multicolumn{4}{c|}{Near OOD Datasets} & \multicolumn{8}{c}{Far OOD Datasets}  \\
\cline{3-14}
& & \multicolumn{2}{c|}{CIFAR10 \cite{krizhevsky2009learning}} & \multicolumn{2}{c|}{TIN \cite{le2015tiny}} & \multicolumn{2}{c|}{MNIST  \cite{lecun2010mnist}} & \multicolumn{2}{c|}{SVHN \cite{netzer2011reading}} & \multicolumn{2}{c|}{Texture \cite{cimpoi2014describing}} & \multicolumn{2}{c}{Places365 \cite{zhou2017places}}  \\
\hline
NTER & OTER & AUC$\uparrow$ & FPR$\downarrow$ & AUC$\uparrow$ & FPR$\downarrow$ & AUC$\uparrow$ & FPR$\downarrow$ & AUC$\uparrow$ & FPR$\downarrow$ & AUC$\uparrow$ & FPR$\downarrow$ & AUC$\uparrow$ & FPR$\downarrow$ \\
\hline 
\ding{55} & \ding{55} & 70.70 & 84.86 & 72.74 & 83.12 & 86.63 & 62.37 & 62.56 & 96.81 & 56.18 & 92.92 & 73.31 & 82.33  \\
\ding{51} & \ding{55} & 71.31 & 84.16 & 73.64 & 82.37 & 89.93 & 52.73 & 65.63 & 95.93 & 59.66 & 91.15 & 75.33 & 79.72  \\
\ding{55} & \ding{51} & 70.77 & 83.84 & 73.80 & 82.10 & 88.65 & 55.36 & 64.67 & 95.80 & 59.26 & 91.94 & 74.35 & 81.34 \\
\ding{51} & \ding{51} & \textbf{71.60} & \textbf{83.17} & \textbf{75.84} & \textbf{81.06} & \textbf{91.56} & \textbf{46.02} & \textbf{67.52} & \textbf{93.08} & \textbf{62.59} & \textbf{90.52} & \textbf{75.90} & \textbf{79.35}  \\
\hline
\end{tabular}}
\label{Ablation}
\end{table*}

\paragraph{The Catastrophic Forgetting Problem Applies to OOD Detection Methods too.}  
This is illustrated in Fig. \ref{Incremental}, where the  performance of different OOD detection methods decreases significantly with increasing incremental steps. This is because, with more incremental steps, more old class samples are wrongly detected as OOD samples, while at the same time, more OOD samples are misclassified into new classes, leading to fast downgraded OOD detection performance.

\paragraph{Catastrophic Forgetting is More Persistent in CIL models than OOD Detection Methods.} 
As shown in Fig. \ref{ACC}, the AUC performance for OOD detection drops fast in early incremental steps, and then it slows down and levers off, but the ACC performance for CIL has a continuous, fast decrease throughout all incremental steps.
This is because remembering the forgotten ID samples is more difficult than distinguishing them from OOD samples.
Also, after the CIL models reach a certain level of forgetting for some ID classes, the performance of distinguishing them from OOD data tends to be stable, resulting in relatively stable OOD detection performance at the end of incremental learning.

\section{Conclusion and Discussion}
\label{con}
In this paper, we introduce OpenCIL, the first benchmark for evaluating the capability of OOD detectors in different types of CIL models.
In OpenCIL, we establish a new evaluation protocol to fairly compare diverse OOD detection methods among different incremental steps, and comprehensively evaluate 60 baselines that are composed by 15 OOD detectors and four CIL models. We further propose a new baseline called BER that can effectively mitigate the common issues in the 60 baselines.
Several insights from our large-scale experiments are summarized as follows:
1) OOD detection performance does not necessarily correlate with the incremental CIL accuracy, 
2) fine-tuning-based OOD methods often achieve better performance than post hoc-based methods, 
3) catastrophic forgetting applies to both CIL classification and OOD detection, but its effect on OOD detection tends to less persistent than on CIL classification,
4) CIL Models are biased towards OOD samples and newly added classes, in which old class samples exhibit high OOD scores and OOD samples exhibit high confidence to be classified into new classes,
and 5) the energy regularization in our proposed BER baseline helps effectively mitigate these biases.

\noindent\textbf{Limitation and Future Work.} OpenCIL focuses on assessing the capability of advanced CIL models in detecting OOD samples in the first place.
As a result, the number of the OOD detectors are much more than that of the CIL models.
Therefore, we plan to extend the benchmark with more CIL methods.
We hope this benchmark will push the development of reliable and safe OOD detection methods tailored for CIL.
In addition, it is also important to study the performance of the CIL models in more open environments, where distribution shifts may exist in the ID datasets.

\bibliographystyle{splncs04}
\bibliography{egbib}


\newpage
\appendix

\section{More Evaluation Setting Details}
\label{Benchmark_detail}

\subsection{Datasets}
The in-distribution (ID) datasets for class incremental learning (CIL) include two datasets. 1) CIFAR100 \cite{krizhevsky2009learning} contains $50,000$ training images and $10,000$ test images of size $32 \times 32$ with $100$ classes, and 2) ImageNet \cite{russakovsky2015imagenet} is a large-scale classification dataset, which contains $1.28$M training images and $50$K testing images with $1,000$ classes sampled from nature images.

We select six commonly used out-of-distribution (OOD) datasets for OOD detection on CIFRA100. Following the recent large-scale OOD detection benchmark OpenOOD \cite{yang2022openood}, our OOD datasets include two near OOD datasets: CIFAR10 \cite{krizhevsky2009learning} and Tiny-ImageNet (TIN) \cite{le2015tiny}, and four far OOD datasets: MNIST  \cite{lecun2010mnist}, Texture \cite{cimpoi2014describing}, Places365 \cite{zhou2017places}, and SVHN \cite{netzer2011reading}.
CIFAR10 \cite{krizhevsky2009learning} contains $10,000$ images with $10$ classes.
TIN \cite{le2015tiny} contains $7,498$ images, which removes the $2502$ images that have overlapping semantics \cite{yang2021semantically} with CIFAR100 classes.
MNIST \cite{lecun2010mnist} contains $10,000$ images with $10$ classes.
Texture \cite{cimpoi2014describing} contains $5,640$ images with $47$ classes.
Places365 \cite{zhou2017places} contains $33,773$ images, which removes the $2727$ images that have overlapping semantics \cite{yang2021semantically} with CIFAR100 classes.
SVHN \cite{netzer2011reading} contains $26,032$ images with $10$ class.

We also select six commonly used OOD datasets for OOD detection on ImageNet1k. Following OpenOOD \cite{yang2022openood}, these includes four near OOD datasets: ImageNet\_O \cite{hendrycks2021natural}, iNaturalist \cite{van2018inaturalist}, OpenImage\_O \cite{wang2022vim} and Species \cite{basart2022scaling}, and two far OOD datasets: MNIST \cite{lecun2010mnist} and Texture \cite{cimpoi2014describing}.
ImageNet\_O \cite{hendrycks2021natural} contains $2,000$ images from categories not found in the ImageNet1k dataset.
We use a $10,000$ image subset of iNaturalist \cite{van2018inaturalist}, which is based on  $110$ mannually selected plant classes not present in ImageNet1k. The OOD samples are randomly sampled images from these $110$ classes \cite{huang2021mos}. All images are resized to have a max dimension of $800$ pixels.
OpenImage\_O \cite{wang2022vim} contains $15, 869$ images with the support of a manual filter.
We use a $10,000$ subset of $713$K images Species \cite{basart2022scaling} with $10$ classes.
Two far OOD datasets are the same as CIFAR100.
A summary of the CIL ID and OOD datasets is presented in Table \ref{data_stats}.

\begin{table*}[ht]
\centering
\caption{Key statistics of used CIL ID datasets and OOD datasets.}
\scalebox{0.8}{
\begin{tabular}{c|ccc|ccc}
\hline
\multirow{2}{*}{Benchmark} & \multicolumn{3}{c|}{CIFAR100} & \multicolumn{3}{c}{ImageNet}  \\
 & Dataset & Images & Class & Dataset & Images & Class\\
\hline
ID data (Training) & CIFAR100 & 50,000 & 100 & ImageNet & 1.28M & 1000 \\
ID data (Testing) & CIFAR100 & 10,000 & 100 & ImageNet & 50,000 & 1000 \\
\hline
\multirow{6}{*}{OOD data} & CIFAR10 & 10,000 & 10 & ImageNet\_O & 2,000 & /  \\
 & TinyImageNet & 7,498 & / & iNaturalist & 10,000 & 110 \\
 & MNIST & 10,000 & 10 & OpenImage\_O & 15,869 & / \\
 & Texture & 5,640 & 47 & Species & 10,000 & 10  \\
 & Places365 & 33,773 & / & MNIST & 10,000 & 10  \\
 & SVHN & 26,032 & 10 & Texture & 5,640 & 47 \\
\hline
\end{tabular}
}
\label{data_stats}
\end{table*}

\subsection{Performance Metrics}
This section provides more introduction to the performance measures used in our experiments.
(1) FPR is the false positive rate of OOD examples when the true positive rate of ID examples is at 95\%. It measures the portion of falsely recognized OOD when the most of ID samples are recalled.
(2) AUC computes the area under the receiver operating characteristic curve of detecting OOD samples, evaluating the OOD detection performance.
(3) AP measures the area under the precision-recall curve, in which the OOD samples are treated as positive samples.
(4) ACC calculates the classification accuracy of the ID data for the CIL models. 
Among all these metrics, only FPR95 is expected to have a lower value for a better model. Higher values indicate better performance for the other three metrics.

\subsection{Codabase}
Existing benchmarks are often incorporated into their proposed methods as a data processing procedure, which does not have a module-based framework that can easily alter CIL or OOD detection methods.
Therefore, we have built an extensible modular-based codebase as the basis of OpenCIL, which consists of four core modules in the framework, including \textit{Data Processing Module}, \textit{CIL Pre-train Module}, \textit{OOD Fine-tune Module} (optional), and \textit{Evaluation Module}.
Users can execute a rearranged script to create a unified \textit{YML} file for each dataset, CIL algorithm, OOD fine-tuning method, and OOD detectors.
These can provide unified data/model management to guarantee consistency across all incremental inputs, such as data preprocessing, incremental step division, memory data collection, and OOD data growth, facilitating direct comparisons among different combinations of CIL models and OOD detection methods, enhancing the transparency and reproducibility of our performance evaluation.

\section{Baseline Library}\label{app:library}
We include four different popular CIL models. iCaRL \cite{rebuffi2017icarl} and WA \cite{zhao2020maintaining} are the regularization-based algorithms, BiC\cite{wu2019large} is a replay-based algorithm, and FOSTER\cite{wang2022foster} is a parameter-isolation-based method but it also uses data replay.

For OOD methods, there are two main categories, post hoc-based and fine-tuning-based methods. 
For post hoc-based methods, we include $9$ OOD detection methods: MSP \cite{hendrycks2016baseline}, ODIN \cite{liang2018enhancing}, Energy \cite{liu2020energy}, MaxLogit \cite{basart2022scaling}, and GEN \cite{liu2023gen} use the statistic based on prediction output of each test samples without using any training ID information, while the other four methods:
ReAct \cite{sun2021react}, KLM \cite{basart2022scaling}, Relation \cite{kim2023neural}, and NNGuide \cite{park2023nearest} need the posterior information of training ID samples to obtain the OOD score. 
Therefore, if they need a whole label space of training ID data, we feed $T^{train+}_t$ to them, which is the combination of memory data and current task training data. 
For fine-tuning-based methods, we include $6$ methods: LogitNorm \cite{wei2022mitigating} and T2FNorm \cite{regmi2023t2fnorm} are the regularization-based methods that apply the normalization to calibrate the loss function,
AUGMIX\cite{hendrycks2020augmix} and REGMIX\cite{pinto2023RegMixup} are the augmentation-based methods that apply the data augmentation to ID samples for enhancing the ID data training, while
VOS\cite{du2022vos} and NPOS\cite{tao2023nonparametric} are synthesis-based methods that generate pseudo-OOD samples to assist the detector training.
Notably, VOS and NPOS focus on outlier synthesis, but applying their original training method to fine-tune the extra classifier is difficult since they are originally design without the final linear classifier. Thus, to have fair comparison, we apply the same energy regularization as our BER for them, which replaces our synthesized pseudo-OOD samples $\bar{x}_t$ with their synthesized outlier samples.

\section{Experimental Results for Other Step Sizes and Near/Far-OOD Detection}
\label{moreresults}
In this section, we provide more detailed averaged results of the OOD detection performance across all incremental steps in our OpenCIL benchmark.

\subsection{Results for Two Other Incremental Step Sizes}

Here we report the averaged results across all incremental steps at the step size of $k = 5$ for CIFAR 100 and at the step size of $k = 50$ for ImangeNet1K in Table \ref{Main_5}, and at the step size of $k = 20$ for CIFAR 100 and at the step size of $k = 200$ for ImangeNet1K in Table \ref{Main_20}. The results for the step size of $k = 10$ for CIFAR 100 and the step size of $k = 100$ for ImangeNet1K are presented in the main text.

\begin{table*}[ht]
\centering
\caption{Main results on OpenCIL benchmark. We report the average performance over six OOD datasets and all incremental steps at the \textbf{step size of $k = 5$ for CIFAR 100} and at the \textbf{step size of $k = 50$ for ImangeNet1K}. 
The \underline{\textbf{best}}, \textbf{second-best}, and \underline{third-best} performance per dataset are highlighted. Either the post hoc-based or fine-tuning-based OOD methods do not affect the original CIL performance, so their average CIL accuracy on the ID data remains the same.
}
\scalebox{0.6}{
\begin{tabular}{c|cc|cc|cc|cc|cc|cc|cc|cc}
\hline\hline
 & \multicolumn{8}{c|}{ID Dataset: CIFAR100} & \multicolumn{8}{c}{ID Dataset: ImageNet1K} \\
 \cline{2-17}
 & \multicolumn{2}{c|}{iCaRL \cite{rebuffi2017icarl}} & \multicolumn{2}{c|}{BiC\cite{wu2019large}}  & \multicolumn{2}{c|}{WA \cite{zhao2020maintaining}} & \multicolumn{2}{c|}{FOSTER\cite{wang2022foster}} & \multicolumn{2}{c|}{iCaRL \cite{rebuffi2017icarl}} & \multicolumn{2}{c|}{BiC\cite{wu2019large}}  & \multicolumn{2}{c|}{WA \cite{zhao2020maintaining}} & \multicolumn{2}{c}{FOSTER\cite{wang2022foster}} \\
 \cline{2-17}
 & AUC$\uparrow$ & FPR$\downarrow$ & AUC$\uparrow$ & FPR$\downarrow$ & AUC$\uparrow$ & FPR$\downarrow$ & AUC$\uparrow$ & FPR$\downarrow$ & AUC$\uparrow$ & FPR$\downarrow$ & AUC$\uparrow$ & FPR$\downarrow$ & AUC$\uparrow$ & FPR$\downarrow$ & AUC$\uparrow$ & FPR$\downarrow$ \\
 \hline
\multicolumn{11}{l}{\textbf{\textit{Average CIL accuracy}}} \\
 & \multicolumn{2}{c|}{58.20} & \multicolumn{2}{c|}{55.87} & \multicolumn{2}{c|}{61.44} & \multicolumn{2}{c|}{63.51} & \multicolumn{2}{c|}{40.86} & \multicolumn{2}{c|}{42.26} & \multicolumn{2}{c|}{45.99} & \multicolumn{2}{c}{45.96} \\   
\hline\hline
\multicolumn{11}{l}{\textbf{\textit{Post hoc-based OOD methods (w/o Fine-tuning)}}} \\
MSP \cite{hendrycks2016baseline} & 66.89 & 87.86 & 68.66 & 87.22 & 69.19 & 86.11 & 70.01 & 85.68 & 61.87 & 91.03 & 63.06 & 92.46 & 61.69 & 92.85 & 65.45 & 90.89  \\
ODIN \cite{liang2018enhancing} & 70.26 & 79.60 & \underline{70.78} & \underline{78.54} & 71.70 & 80.09 & 72.89 & \textbf{75.48} & \underline{\textbf{66.00}} & \underline{\textbf{88.90}} & \underline{67.62} & 90.54 & \underline{63.47} & 91.07 & \textbf{69.16} & \underline{\textbf{87.58}} \\
Energy \cite{liu2020energy} & 70.23 & 81.17 & 69.55 & 82.78 & 71.99 & 82.53 & 73.89 & 76.34 & 62.97 & 91.95 & 65.61 & 92.43 & 62.43 & 92.94 & 67.46 & 90.94 \\
MaxLogit \cite{basart2022scaling} & 70.16 & 81.87 & 69.87 & 83.31 & 71.89 & 82.96 & 73.82 & 77.34 & \textbf{63.94} & 91.58 & 64.67 & 93.32 & \textbf{63.51} & 91.44 & 67.31 & 91.30 \\
GEN \cite{liu2023gen} & 70.39 & 82.07 & \textbf{70.89} & 79.54 & 72.22 & 82.14 & \textbf{74.21} & 76.95 & 55.67 & 93.51 & 60.65 & 92.82 & 61.27 & 93.38 & 59.94 & 94.26 \\
ReAct \cite{sun2021react} & 70.21 & 81.26 & 69.86 & 84.89 & \underline{\textbf{73.54}} & 82.15 & \underline{\textbf{74.29}} & 76.97 & 63.34 & 89.99 & 62.93 & \underline{89.79} & 63.18 & \textbf{89.52} & 66.33 & \textbf{87.95} \\
KLM \cite{basart2022scaling} & 66.21 & 88.89 & 67.53 & 86.71 & 68.19 & 87.70 & 69.38 & 86.36  & 63.34 & 89.99 & 62.93 & 89.79 & 63.18 & \textbf{89.52} & 66.33 & \textbf{87.95} \\
Relation \cite{kim2023neural} & 66.33 & \underline{78.06} & 70.64 & 81.42 & 71.89 & \underline{\textbf{77.19}} & 72.49 & 75.74  & 63.13 & 89.86 & 66.39 & 93.63 & 63.11 & 95.41 & 63.48 & 90.83\\
NNGuide \cite{park2023nearest} & 70.27 & 78.83 & 70.70 & 79.64 & 71.60 & \underline{79.00} & 73.68 & 75.96 & 63.00 & \textbf{89.47} & \underline{\textbf{69.87}} & \underline{\textbf{85.81}} & 62.61 & 90.30 & 68.63 & 88.54\\
\hline
Average & 68.99 & 82.18 & 69.83 & 82.67 & 71.36 & 82.21 & 72.74 & 78.54 & 62.58 & 90.70 & 64.86 & 91.18 & 62.72 & 91.83 & 66.01 & 90.03 \\
\hline\hline
\multicolumn{11}{l}{\textbf{\textit{Fine-tuning-based OOD methods (w/ Fine-tuning)}}}\\
LogitNorm \cite{wei2022mitigating} & 70.21 & 81.18 & 69.22 & 83.39 & 71.13 & 82.65 & 73.31 & 76.78 & 62.02 & 92.73 & 66.30 & 91.20 & 61.58 & 93.94 & 65.87 & 93.26  \\
T2FNorm \cite{regmi2023t2fnorm} & 70.45 & 81.50 & 69.59 & 83.29 & 70.90 & 83.26 & 73.26 & 77.35 & 62.82 & 92.12 & 65.80 & 92.30 & 62.40 & 92.84 & 66.55 & 92.01 \\
AUGMIX \cite{hendrycks2020augmix} & 70.27 & 81.22 & 68.65 & 83.16 & 71.11 & 82.93 & 73.12 & 76.98 & 62.19 & 90.86 & 65.97 & 89.82 & 62.88 & 90.12 & 67.53 & 89.45 \\
REGMIX \cite{pinto2023RegMixup} & 70.92 & 81.33 & 68.99 & 83.65 & 71.44 & 84.30 & 73.78 & 77.56  & \underline{63.93} & 91.20 & 66.24 & 90.29 & 62.96 & 91.64 & 67.84 & 91.56\\
VOS \cite{du2022vos}  & \underline{71.54} & 79.83 & 67.63 &  \underline{\textbf{77.80}} & 66.73 & 81.28 & 72.70 & 77.08 & 61.33 & 89.74 & 65.74 & 90.22 & 62.08 & 89.98 & 66.73 & 90.14  \\
NPOS \cite{tao2023nonparametric} & \textbf{71.82} & \textbf{78.03} & 69.47 & 78.68 & \underline{72.46} & 79.31 & 74.17 & \underline{75.55} & 63.06 & 89.86 & 67.22 & 89.83 & 63.21 & \underline{89.72} & \underline{68.93} & 89.63 \\
\textbf{BER (Ours)} & \underline{\textbf{72.75}} & \underline{\textbf{77.59}} & \underline{\textbf{71.47}} & \textbf{77.82} & \textbf{72.47} & \textbf{78.69} & \underline{74.20} & \underline{\textbf{74.93}}  & 63.45 & \underline{89.53} & \textbf{67.72} & \textbf{88.61} & \underline{\textbf{64.09}} & \underline{\textbf{89.15}} & \underline{\textbf{69.34}} & \underline{88.39} \\
\hline
Average & 71.14 & 80.10 & 69.29 & 81.11 & 70.89 & 81.77 & 73.51 & 76.60 & 62.69 & 90.86 & 66.43 & 90.32 & 62.74 & 91.06 & 67.54 & 90.63 \\
\hline\hline
Average (All) & 69.93 & 81.27 & 69.59 & 81.99 & 71.15 & 82.02 & 73.08 & 77.69 & 62.63 & 90.77 & 65.55 & 90.80 & 62.73 & 91.49 & 66.68 & 90.29 \\
\hline\hline
\end{tabular}}
\label{Main_5}
\end{table*}

\begin{table*}[ht]
\centering
\caption{Main results on OpenCIL benchmark. We report the average performance over six OOD datasets and all incremental steps at the \textbf{step size of $k = 20$ for CIFAR 100} and at the \textbf{step size of $k = 200$ for ImangeNet1K}. 
The \underline{\textbf{best}}, \textbf{second-best}, and \underline{third-best} performance per dataset are highlighted. Either the post hoc-based or fine-tuning-based OOD methods do not affect the original CIL performance, so their average CIL accuracy on the ID data remains the same.
}
\scalebox{0.6}{
\begin{tabular}{c|cc|cc|cc|cc|cc|cc|cc|cc}
\hline\hline
 & \multicolumn{8}{c|}{ID Dataset: CIFAR100} & \multicolumn{8}{c}{ID Dataset: ImageNet1K} \\
 \cline{2-17}
 & \multicolumn{2}{c|}{iCaRL \cite{rebuffi2017icarl}} & \multicolumn{2}{c|}{BiC\cite{wu2019large}}  & \multicolumn{2}{c|}{WA \cite{zhao2020maintaining}} & \multicolumn{2}{c|}{FOSTER\cite{wang2022foster}} & \multicolumn{2}{c|}{iCaRL \cite{rebuffi2017icarl}} & \multicolumn{2}{c|}{BiC\cite{wu2019large}}  & \multicolumn{2}{c|}{WA \cite{zhao2020maintaining}} & \multicolumn{2}{c}{FOSTER\cite{wang2022foster}} \\
 \cline{2-17}
 & AUC$\uparrow$ & FPR$\downarrow$ & AUC$\uparrow$ & FPR$\downarrow$ & AUC$\uparrow$ & FPR$\downarrow$ & AUC$\uparrow$ & FPR$\downarrow$ & AUC$\uparrow$ & FPR$\downarrow$ & AUC$\uparrow$ & FPR$\downarrow$ & AUC$\uparrow$ & FPR$\downarrow$ & AUC$\uparrow$ & FPR$\downarrow$ \\
 \hline
\multicolumn{11}{l}{\textbf{\textit{Average CIL accuracy}}} \\
 & \multicolumn{2}{c|}{62.65} & \multicolumn{2}{c|}{64.14} & \multicolumn{2}{c|}{68.05} & \multicolumn{2}{c|}{68.75} & \multicolumn{2}{c|}{48.85} & \multicolumn{2}{c|}{53.30} & \multicolumn{2}{c|}{58.42} & \multicolumn{2}{c}{57.84} \\   
\hline\hline
\multicolumn{11}{l}{\textbf{\textit{Post hoc-based OOD methods (w/o Fine-tuning)}}} \\
MSP \cite{hendrycks2016baseline} & 68.38 & 87.19 & 69.72 & 87.75 & 72.46 & 84.74 & 72.10 & 85.02 & 66.83 & 87.47 & 71.39 & 86.68 & 70.73 & 87.12 & 71.57 & 84.44 \\
ODIN \cite{liang2018enhancing} & 71.34 & 81.20 & 71.46 & 80.62 & 74.27 & 78.08 & 74.45 & 76.73 & \underline{71.36} & \underline{\textbf{83.61}} & 75.27 & \textbf{82.55} & 71.69 & \underline{\textbf{84.60}} & 73.56 & 82.93 \\
Energy \cite{liu2020energy} & 72.38 & 81.99 & 71.48 & 82.45 & 75.22 & 79.00 & 76.14 & 78.54 & 69.52 & 88.30 & 74.36 & 84.92 & 70.56 & 87.73 & 72.67 & 87.09 \\
MaxLogit \cite{basart2022scaling} & 72.32 & 82.20 & 71.60 & 83.73 & 75.22 & 80.02 & 76.08 & 78.84 & 69.62 & 87.90 & 74.17 & 85.99 & 71.08 & 87.26 & 73.15 & 85.57 \\
GEN \cite{liu2023gen} & 72.73 & 82.44 & \textbf{71.92} & 83.00 & 75.36 & 79.38 &\textbf{76.55} & 78.39 & 70.16 & 87.60 & 74.19 & 85.84 & 71.68 & \textbf{85.67} & \textbf{73.84} & 84.30 \\
ReAct \cite{sun2021react} & 71.59 & 82.99 & \underline{\textbf{73.69}} & 82.87 & \underline{\textbf{76.82}}  & 78.49 & \underline{\textbf{77.33}} & 78.85 & 59.48 & 92.57 & 69.67 & 85.93 & 63.12 & 88.28 & 64.82 & 89.13 \\
KLM \cite{basart2022scaling} & 69.10 & 85.53 & 70.19 & 86.76 & 72.47 & 86.06 & 72.47 & 84.39 & 69.51 & \textbf{84.99} & 70.55 & 85.72 & 69.61 & 87.49 & 73.55 & \underline{\textbf{82.10}} \\
Relation \cite{kim2023neural} & 65.37 & 80.08 & 69.10 & 79.46 & 76.23 & 77.46 & 71.27 & \underline{76.53} & 71.24 & 87.09 & 75.12 & 83.34 & 71.45 & 92.75 & 73.27 & 83.31 \\
NNGuide \cite{park2023nearest} & \underline{74.14} & \textbf{77.05} & 71.24 & 78.89 & 76.28 & \underline{73.78} & 76.20 & \textbf{76.13}  & \textbf{71.41} & 88.03 & \underline{\textbf{77.95}}  & 86.53 & \textbf{71.90} & 86.90 & 73.27 & \underline{82.74} \\
\hline
Average & 70.82 & 82.30 & 71.16 & 82.84 & 74.93 & 79.67 & 74.73 & 79.27 & 68.79 & 87.51 & 73.63 & 85.28 & 70.20 & 87.53 & 72.19 & 84.62 \\
\hline\hline
\multicolumn{11}{l}{\textbf{\textit{Fine-tuning-based OOD methods (w/ Fine-tuning)}}}\\
LogitNorm \cite{wei2022mitigating} & 72.41 & 82.32 & 71.27 & 83.80 & 74.29 & 79.49 & 75.38 & 79.28 & 69.40 & 89.17 & 73.18 & 84.05 & 69.14 & 89.57 & 71.62 & 89.30 \\
T2FNorm \cite{regmi2023t2fnorm} & 72.67 & 82.39 & 71.51 & 83.79 & 74.38 & 80.66 & 75.70 & 79.60 & 70.45 & 87.47 & 74.21 & \underline{\textbf{82.17}} & 70.95 & 87.36 & 71.64 & 87.72 \\
AUGMIX \cite{hendrycks2020augmix} & 72.50 & 82.62 & 70.70 & 84.33 & 74.35 & 79.97 & 75.54 & 79.61 & 71.32 & 87.91 & \textbf{76.72} & 84.63 & 70.84 & 87.20 & 72.81 & 83.66\\
REGMIX \cite{pinto2023RegMixup} & 72.74 & 82.81 & 71.62 & 84.65 & 74.78 & 80.45 & 75.69 & 80.09 & 69.56 & 89.61 & 72.78 & 87.57 & 70.68 & 88.13 & 71.84 & 88.85 \\
VOS \cite{du2022vos} & 72.93 & 81.39 & 68.77 & \textbf{78.28} & 72.21 & 76.57 & 75.07 & 78.52 & 62.94 & 89.63 & 64.19 & 88.20 & 62.86 & 89.02 & 65.61 & 88.09 \\
NPOS \cite{tao2023nonparametric} & \textbf{74.22} & \underline{78.89} & 70.86 & \underline{78.43} & \underline{76.61} & \textbf{71.86} & 76.44 & 76.92  & 71.00 & 87.21 & 75.59 & 84.11 & \underline{71.80} & 87.86 & \underline{73.57} & 83.13 \\
\textbf{BER (Ours)} & \underline{\textbf{74.55}} & \underline{\textbf{76.75}} & \underline{71.90} & \underline{\textbf{77.97}} & \textbf{76.64} & \underline{\textbf{71.72}} & \underline{76.52}  & \underline{\textbf{75.04}} & \underline{\textbf{71.88}} & \underline{87.00} & \underline{76.46} & \underline{83.01} & \underline{\textbf{71.98}} & \underline{86.24} & \underline{\textbf{73.95}} & \textbf{82.66} \\
\hline
Average & 73.15 & 81.02 & 70.95 & 81.61 & 74.75 & 77.25 & 75.76 & 78.44 & 69.51 & 88.29 & 73.30 & 84.82 & 69.75 & 87.91 & 71.58 & 86.20 \\
\hline\hline
Average (All) & 71.84 & 81.74 & 71.06 & 82.30 & 74.85 & 78.61 & 75.18 & 78.91 & 69.10 & 87.85 & 73.49 & 85.08 & 70.00 & 87.70 & 71.92 & 85.31 \\
\hline\hline
\end{tabular}}
\label{Main_20}
\end{table*}

\subsection{Fine-grained Results on Near- and Far-OOD Detection Datasets}

Following the experiment results in our paper, we provide more fine-grained results (on near-OOD datasets and far-OOD datasets, respectively) at different step sizes in Tables 
\ref{Main_fine_5}, \ref{Main_fine}, and \ref{Main_fine_20}, in which we also add the AP results of the OOD detection performance.

\begin{table*}[!ht]
\centering
\caption{Detailed results for those in Table \ref{Main_5} on \textbf{near- and far-OOD detection datasets} at the \textbf{step size of $k = 5$ for CIFAR 100} and at the \textbf{step size of $k = 50$ for ImangeNet1K}.
The \underline{\textbf{best}}, \textbf{second-best}, and \underline{third-best} performance per dataset are highlighted. The upper, middle, lower parts of the table are for AUC, AP, FPR performance, respectively.
} 
\scalebox{0.6}{
\begin{tabular}{c|cc|cc|cc|cc|cc|cc|cc|cc}
\hline

\hline
 \multirow{3}{*}{\textcolor{blue}{\textbf{AUC$\uparrow$}}} & \multicolumn{8}{c|}{ID Dataset: CIFAR100} & \multicolumn{8}{c}{ID Dataset: ImageNet1K} \\
 \cline{2-17}
 & \multicolumn{2}{c|}{iCaRL \cite{rebuffi2017icarl}} & \multicolumn{2}{c|}{BiC\cite{wu2019large}}  & \multicolumn{2}{c|}{WA \cite{zhao2020maintaining}} & \multicolumn{2}{c|}{FOSTER\cite{wang2022foster}} & \multicolumn{2}{c|}{iCaRL \cite{rebuffi2017icarl}} & \multicolumn{2}{c|}{BiC\cite{wu2019large}}  & \multicolumn{2}{c|}{WA \cite{zhao2020maintaining}} & \multicolumn{2}{c}{FOSTER\cite{wang2022foster}} \\
 \cline{2-17}
 & Near & Far & Near & Far & Near & Far & Near & Far & Near & Far & Near & Far & Near & Far & Near & Far \\
 \hline
\multicolumn{11}{l}{\textbf{\textit{Average CIL accuracy}}} \\
 & \multicolumn{2}{c|}{58.20} & \multicolumn{2}{c|}{55.87} & \multicolumn{2}{c|}{61.44} & \multicolumn{2}{c|}{63.51} & \multicolumn{2}{c|}{40.86} & \multicolumn{2}{c|}{42.26} & \multicolumn{2}{c|}{45.99} & \multicolumn{2}{c}{45.96} \\ 
\hline

\hline
\multicolumn{11}{l}{\textbf{-\textit{Post hoc-based OOD methods (w/o Fine-tuning)}}} \\
MSP \cite{hendrycks2016baseline}  & 67.21 & 66.73 & 67.69 & 69.15 & 70.22 & 68.68 & 70.14 & 69.94 & 61.82 & 61.98 & 65.12 & 58.92 & 62.50 & 60.07 & 64.12 & 68.12 \\
ODIN \cite{liang2018enhancing}  & 70.97 & 69.91 & \underline{69.22} & 71.56 & 72.98 & 71.06 & 72.25 & 73.21 & \textbf{64.80} & \textbf{68.40} & 68.11 & \underline{66.65}  & 63.29 & 63.83 & 66.19 & \textbf{75.10} \\
Energy \cite{liu2020energy}  & 71.23 & 69.73 & 68.56 & 70.05 & \textbf{73.19} & 71.38 & 72.98 & 74.34 & 62.62 & 63.07 & 67.58 & 63.37 & 62.00 & 63.50 & 64.59 & 73.29 \\
MaxLogit \cite{basart2022scaling} & 71.16 & 69.66 & 68.85 & 70.38 & \underline{73.16} & 71.26 & 72.98 & 74.25 & 62.87 & 63.17 & 67.11 & 62.62 & 62.65 & 62.01 & 65.02 & 72.34 \\
GEN \cite{liu2023gen} & 71.18 & 69.99 & \textbf{69.44} & 71.61 & \underline{\textbf{73.30}} & 71.68 & 73.34 & \underline{74.65} & 63.55 & 64.72 & 65.69 & 62.61 & \underline{\textbf{64.69}} & 61.15 & 65.23 & 71.47 \\
ReAct \cite{sun2021react} & 68.67 & 70.98 & 66.29 & 71.65 & 72.30 & \textbf{74.16} & 70.81 & \textbf{76.03} & 57.92 & 51.17 & 63.42 & 55.09 & 62.11 & 59.59 & 59.26 & 61.31 \\
KLM \cite{basart2022scaling} & 66.78 & 65.92 & 66.44 & 68.09 & 68.55 & 68.01 & 69.47 & 69.35 & 63.29 & 63.46 & 63.62 & 61.55 & 63.40 & 62.74 & 65.93 & 67.15 \\
Relation \cite{kim2023neural} & 61.30 & 68.85 & 65.38 &  \underline{\textbf{73.27}} & 67.53 & \underline{74.07} & 68.42 & 74.52 & 58.73 & \underline{\textbf{69.95}} & 66.19 & \textbf{66.80} & 63.22 & 62.87 & 61.05 & 68.37 \\
NNGuide \cite{park2023nearest} & 67.33 &  \underline{71.73} & 66.81 & \underline{72.65} & 68.53 & 73.13 & 68.97 &  \underline{\textbf{76.03}} & 61.20 & \underline{66.61} & \underline{\textbf{70.63}} & \underline{\textbf{68.36}} & 58.74 & \underline{\textbf{70.33}} & 63.94 & \underline{\textbf{78.02}} \\
\hline
Average  & 68.43 & 69.28 & 67.63 & 70.93 & 71.08 & 71.49 & 71.04 & 73.59 & 61.87 & 63.61 & 66.39 & 62.89 & 62.51 & 62.90 & 63.93 & 70.57  \\
\hline

\hline
\multicolumn{11}{l}{\textbf{-\textit{Fine-tuning-based OOD methods (w/ Fine-tuning)}}} \\
LogitNorm \cite{wei2022mitigating} & 71.22 & 69.70 & 68.28 & 69.70 & 72.11 & 70.64 & 72.59 & 73.66 & 61.49 & 63.09 & 66.52 & 65.86 & 61.25 & 62.25 & 63.53 & 70.53 \\
T2FNorm \cite{regmi2023t2fnorm}  & 71.37 & 69.99 & 68.03 & 70.37 & 71.45 & 70.62 & 72.50 & 73.64 & 62.65 & 63.16 & 66.47 & 64.45 & 62.00 & 63.21 & 64.16 & 71.34 \\
AUGMIX \cite{hendrycks2020augmix}  & 71.47 & 69.67 & 68.12 & 68.92 & 72.15 & 70.58 & 72.74 & 73.32 & 62.79 & 60.99 & 67.43 & 63.05 & 61.96 & \textbf{64.72} & 65.97 & 71.09  \\
REGMIX \cite{pinto2023RegMixup} & \textbf{71.99} & 70.38 & 68.08 & 69.44 & 72.94 & 70.69 & \textbf{73.56} & 73.90 & \underline{\textbf{66.05}} & 59.70 & 68.13 & 62.45 & \textbf{64.26} & 60.36 & \textbf{67.13} & 69.27\\
VOS \cite{du2022vos} & \underline{71.96} & 71.33 & 61.91 & 70.48 & 62.23 & 68.98 & 72.18 & 72.96 & 62.17 & 59.65 & 67.03 & 63.16 & 61.48 & 63.28 & 65.24 & 69.71 \\
NPOS \cite{tao2023nonparametric} & 71.54 & \textbf{71.96} & 62.03 &  \textbf{73.20} & 70.37 & 73.50 & \underline{73.40} & 74.56 & 62.22 & 64.74 & \underline{68.45} & 64.76 & 63.56 & 62.51 & \underline{66.97} & 72.85 \\
\textbf{BER (Ours)} & \underline{\textbf{72.54}} & \underline{\textbf{72.85}} & \underline{\textbf{69.87}} & 72.28 & 68.21 & \underline{\textbf{74.60}} & \underline{\textbf{73.60}} & 74.49 & \underline{64.04}  & 62.29 & \textbf{68.96} & 65.25 & \underline{64.00} & \underline{64.30}  & \underline{\textbf{67.16}} & \underline{73.70}  \\
\hline
Average  & 71.73 & 70.84 & 66.62 & 70.63 & 69.92 & 71.37 & 72.94 & 73.79 & 63.06 & 61.95 & 67.57 & 64.14 & 62.64 & 62.95 & 65.74 & 71.21
\\
\hline 

\hline
Average (All)  & 69.87 & 69.96 & 67.19 & 70.80 & 70.58 & 71.44 & 71.87 & 73.68 & 62.39 & 62.88 & 66.90 & 63.43 & 62.57 & 62.92 & 64.72 & 70.85 \\
\hline

\hline
\hline

\hline
\multirow{2}{*}{\textcolor{blue}{\textbf{AP$\uparrow$}}}  & \multicolumn{2}{c|}{iCaRL \cite{rebuffi2017icarl}} & \multicolumn{2}{c|}{BiC\cite{wu2019large}}  & \multicolumn{2}{c|}{WA \cite{zhao2020maintaining}} & \multicolumn{2}{c|}{FOSTER\cite{wang2022foster}} & \multicolumn{2}{c|}{iCaRL \cite{rebuffi2017icarl}} & \multicolumn{2}{c|}{BiC\cite{wu2019large}}  & \multicolumn{2}{c|}{WA \cite{zhao2020maintaining}} & \multicolumn{2}{c}{FOSTER\cite{wang2022foster}} \\
 \cline{2-17}
 & Near & Far & Near & Far & Near & Far & Near & Far & Near & Far & Near & Far & Near & Far & Near & Far \\
 \hline
\multicolumn{11}{l}{\textbf{\textit{Average CIL accuracy}}} \\
 & \multicolumn{2}{c|}{58.20} & \multicolumn{2}{c|}{55.87} & \multicolumn{2}{c|}{61.44} & \multicolumn{2}{c|}{63.51} & \multicolumn{2}{c|}{40.86} & \multicolumn{2}{c|}{42.26} & \multicolumn{2}{c|}{45.99} & \multicolumn{2}{c}{45.96} \\ 
\hline

\hline
\multicolumn{11}{l}{\textbf{-\textit{Post hoc-based OOD methods (w/o Fine-tuning)}}} \\
MSP \cite{hendrycks2016baseline} & 62.73 & 77.05 & 62.57 & 76.91 & 65.46 & 77.77 & 65.08 & 78.70 & 22.90 & 20.01 & 24.27 & 16.79 & 22.30 & 18.64 & 23.44 & 23.64 \\
ODIN \cite{liang2018enhancing} & 66.72 & 76.69 & \textbf{64.51} & 77.01 & 68.34 & 77.22 & 67.73 & 78.80 & 24.11 & \underline{\textbf{26.47}} & 26.44 & 20.77 & 22.84 & \underline{21.73} & 24.24 & \underline{\textbf{31.43}} \\
Energy \cite{liu2020energy}  & 66.91 & 76.88 & 63.27 & 76.01 & \underline{68.42} & 77.48 & 68.44 & 79.37 & 22.40 & 21.21 & 25.73 & 17.93 & 21.63 & 19.61 & 22.91 & \underline{27.15} \\
MaxLogit \cite{basart2022scaling} & 66.79 & 76.94 & 63.73 & 76.48 & 68.34 & 77.69 & 68.38 & 79.46 & 22.70 & 21.16 & 25.32 & 17.72 & 22.10 & 19.25 & 23.25 & 26.18 \\
GEN \cite{liu2023gen} & 66.82 & 77.17 & \underline{\textbf{64.69}} & 77.14 & \textbf{68.47} & 77.76 & \underline{68.63} & \underline{79.70} & 23.10 & 21.71 & 23.77 & 18.57 & \underline{24.54} & 18.68 & 23.54 & 24.96 \\
ReAct \cite{sun2021react} & 65.35 & \textbf{78.84} & 61.33 & 77.20 & 67.66 & 79.19 & 66.75 & \textbf{80.90} & 21.14 & 15.51 & 23.59 & 15.14 & 23.24 & 16.84 & 21.55 & 17.84 \\
KLM \cite{basart2022scaling} & 62.17 & 76.80 & 62.02 & 78.65 & 62.92 & 78.44 & 64.09 & 79.07 & 24.13 & 20.25 & 23.96 & 19.58 & 23.91 & 21.54 & \textbf{26.15} & 23.51 \\
Relation \cite{kim2023neural} & 61.13 & 78.48 & 61.63 & \underline{\textbf{80.21}} & 64.64 & \underline{\textbf{82.66}} & 65.50 & \underline{\textbf{84.27}} & \underline{25.75} & 20.03 & \underline{26.93} & \underline{21.43} & 22.16 & 19.27 & 22.46 & 25.88 \\
NNGuide \cite{park2023nearest} & 66.45 & \underline{\textbf{80.63}} & 62.47 & 78.56 & 67.86 & \textbf{81.45} & 66.92 & 78.62 & 24.47 & 18.74 & 21.35 & \underline{\textbf{23.34}} & 23.98 & \underline{\textbf{25.34}} & 25.62 & \textbf{30.94} \\
\hline 
Average  & 65.01 & 77.72 & 62.91 & 77.57 & 66.90 & 78.85 & 66.84 & 79.88 & 23.41 & 20.57 & 24.60 & 19.03 & 22.97 & 20.10 & 23.68 & 25.73 \\
\hline

\hline
\multicolumn{11}{l}{\textbf{\textit{Fine-tuning-based OOD methods (w/ Fine-tuning)}}}\\
LogitNorm \cite{wei2022mitigating} & 66.92 & 76.86 & 63.39 & 76.09 & 67.54 & 77.16 & 68.07 & 79.06 & 21.68 & 20.97 & 25.67 & 19.59 & 21.05 & 19.16 & 22.21 & 23.25 \\
T2FNorm \cite{regmi2023t2fnorm} & 67.03 & 77.10 & 63.24 & 76.85 & 66.88 & 77.44 & 67.93 & 79.19 & 22.44 & 21.16 & 24.99 & 18.61 & 21.71 & 19.77 & 22.82 & 24.41 \\
AUGMIX \cite{hendrycks2020augmix} & 67.12 & 76.84 & 63.17 & 75.63 & 67.52 & 77.13 & 68.22 & 78.88 & 24.63 & 19.31 & 23.96 & 20.14 & 23.79 & 18.54 & 24.83 & 22.17 \\
REGMIX \cite{pinto2023RegMixup} & \underline{67.65} & 77.26 & 63.17 & 75.92 & 68.31 & 77.23 & \underline{\textbf{69.22}} & 79.32 & \textbf{25.98} & 17.61 & \underline{\textbf{27.59}} & 17.62 & 24.36 & 17.61 & \underline{25.92} & 20.54 \\
VOS \cite{du2022vos} & 67.50 & 77.56 & 58.66 & 77.45 & 59.23 & 77.19 & 67.59 & 78.69 & 23.73 & 19.60 & 26.89 & 20.23 & 23.48 & 19.63 & 24.86 & 24.53 \\
NPOS \cite{tao2023nonparametric} & \textbf{67.78} & 78.56 & 59.06 & \underline{79.51} & 66.09 & 79.16 & 68.62 & 79.60 & 24.96 & \underline{22.63} & 26.89 & 20.78 & \textbf{24.64} & 21.03 & 24.02 & 25.70 \\
\textbf{BER (Ours)} & \underline{\textbf{67.81}} & \underline{78.63} & \underline{64.43}  & \textbf{79.57} & \underline{\textbf{69.56}} & \underline{81.00} & \textbf{68.89} & 79.69 & \underline{\textbf{26.86}} & \textbf{22.65} & \textbf{27.04} & \textbf{22.05} & \underline{\textbf{26.54}} & \textbf{22.23} & \underline{\textbf{27.46}} & 24.02 \\
\hline
Average  & 67.40 & 77.54 & 62.16 & 77.29 & 66.45 & 78.04 & 68.36 & 79.20 & 24.33 & 20.56 & 26.15 & 19.86 & 23.65 & 19.71 & 24.59 & 23.52 \\
\hline 

\hline
Average (All)  & 66.05 & 77.64 & 62.58 & 77.45 & 66.70 & 78.50 & 67.50 & 79.58 & 23.81 & 20.56 & 25.27 & 19.39 & 23.27 & 19.93 & 24.08 & 24.76 \\
\hline

\hline
\hline

\hline
\multirow{2}{*}{\textcolor{blue}{\textbf{FPR$\downarrow$}}}  & \multicolumn{2}{c|}{iCaRL \cite{rebuffi2017icarl}} & \multicolumn{2}{c|}{BiC\cite{wu2019large}}  & \multicolumn{2}{c|}{WA \cite{zhao2020maintaining}} & \multicolumn{2}{c|}{FOSTER\cite{wang2022foster}} & \multicolumn{2}{c|}{iCaRL \cite{rebuffi2017icarl}} & \multicolumn{2}{c|}{BiC\cite{wu2019large}}  & \multicolumn{2}{c|}{WA \cite{zhao2020maintaining}} & \multicolumn{2}{c}{FOSTER\cite{wang2022foster}} \\
 \cline{2-17}
 & Near & Far & Near & Far & Near & Far & Near & Far & Near & Far & Near & Far & Near & Far & Near & Far \\
 \hline
\multicolumn{11}{l}{\textbf{\textit{Average CIL accuracy}}} \\
 & \multicolumn{2}{c|}{58.20} & \multicolumn{2}{c|}{55.87} & \multicolumn{2}{c|}{61.44} & \multicolumn{2}{c|}{63.51} & \multicolumn{2}{c|}{40.86} & \multicolumn{2}{c|}{42.26} & \multicolumn{2}{c|}{45.99} & \multicolumn{2}{c}{45.96} \\   
\hline

\hline
\multicolumn{11}{l}{\textbf{-\textit{Post hoc-based OOD methods (w/o Fine-tuning)}}} \\
MSP \cite{hendrycks2016baseline} & 87.34 & 88.12 & 88.23 & 86.72 & 85.16 & 86.59 & 86.10 & 85.47 & 91.12 & 90.83 & 91.25 & 94.89 & 92.70 & 93.14 & 91.83 & 89.02 \\
ODIN \cite{liang2018enhancing}  & \underline{83.67} & 77.57 & \underline{\textbf{85.70}} & 74.96 & \textbf{82.91} & 78.69 & 83.14 & 71.65 & 90.89 & \underline{\textbf{84.91}} & 89.34 & 92.93 & 91.66 & 89.91 & 90.92 &  \underline{\textbf{80.90}} \\
Energy \cite{liu2020energy} & 83.78 & 79.86 & 87.19 & 80.57 & 83.40 & 82.10 & 82.72 & 73.14 & 92.65 & 90.98 & 90.53 & 95.36 & 92.95 & 93.25 & 92.54 & 87.43 \\
MaxLogit \cite{basart2022scaling}  & 83.91 & 80.85 & 86.80 & 81.57 & 83.07 & 82.90 & 82.94 & 74.54 & 92.46 & 90.94 & 91.00 & 95.30 & 92.92 & 92.97 & 92.41 & 88.00 \\
GEN \cite{liu2023gen} & 83.84 & 81.19 & \textbf{85.88} & 76.37 & \underline{\textbf{82.77}} & 81.82 & \underline{82.56} & 74.14 & 92.12 & 90.50 & 92.30 & 95.37 & 90.40 & 93.52 & 92.00 & 89.90 \\
ReAct \cite{sun2021react} & 83.83 & 79.97 & 88.95 & 82.85 & 83.64 & 81.41 & 83.63 & 73.64 & 92.98 & 94.57 & 91.42 & 95.62 & 92.18 & 95.77 & 93.90 & 94.98 \\
KLM \cite{basart2022scaling} & 87.94 & 89.37 & 88.13 & 86.00 & 87.78 & 87.66 & 87.06 & 86.00 & 89.88 & 90.20 & 89.77 & \textbf{89.84} & 90.59 & \underline{\textbf{87.38}} & \textbf{88.30} & \underline{87.25} \\
Relation \cite{kim2023neural} & 85.39 & \underline{\textbf{74.39}} & 88.48 & 77.88 & 84.44 & \underline{\textbf{73.56}} & 84.10 & 71.57 & 91.78 & \textbf{86.00} & 92.85 & 95.19 & 95.52 & 95.19 & 92.56 & 87.38 \\
NNGuide \cite{park2023nearest}  & 85.66 & 75.41 & 87.97 & 75.47 & 86.10 & \textbf{75.44} & 87.47 & \underline{\textbf{70.20}} & 90.00 & \underline{88.40} & \underline{\textbf{84.14}} & \underline{\textbf{89.16}} & \underline{90.30} & 90.30 & 89.94 & \textbf{85.73} \\
\hline
Average  & 85.04 & 80.75 & 87.48 & 80.27 & 84.36 & 81.13 & 84.41 & 75.59 & 91.54 & 89.70 & 90.29 & 93.74 & 92.14 & 92.38 & 91.60 & 87.84 \\
\hline

\hline
\multicolumn{11}{l}{\textbf{\textit{Fine-tuning-based OOD methods (w/ Fine-tuning)}}}\\
LogitNorm \cite{wei2022mitigating} & 83.77 & 79.88 & 86.75 & 81.70 & 83.71 & 82.11 & 83.01 & 73.66 & 93.23 & 91.73 & 89.61 & 94.38 & 93.92 & 93.98 & 93.39 & 93.03 \\
T2FNorm \cite{regmi2023t2fnorm} & 83.69 & 80.39 & 86.75 & 81.55 & 83.97 & 82.91 & 82.95 & 74.56 & 92.62 & 91.12 & 90.77 & 95.36 & 92.97 & 92.58 & 92.67 & 90.70 \\
AUGMIX \cite{hendrycks2020augmix} & 83.69 & 79.98 & 86.92 & 81.28 & 83.71 & 82.55 & \textbf{82.53} & 74.20 & 90.16 & 92.26 & 89.24 & 90.98 & 90.73 & \underline{88.90} & 90.02 & 88.31  \\
REGMIX \cite{pinto2023RegMixup} & \textbf{83.47} & 80.26 & 87.06 & 81.94 & 83.31 & 84.80 & \underline{\textbf{81.93}} & 75.38 & 89.56 & 94.48 & \underline{88.16} & 94.54 & 90.30 & 94.34 & 89.97 & 94.73\\
VOS \cite{du2022vos}  & 83.67 & 77.91 & 89.30 & \underline{\textbf{72.05}} & 88.44 & 77.70 & 83.55 & 73.84 & \underline{89.43} & 90.36 & 89.65 & 91.36 & 90.74 & \textbf{88.46} & 89.66 & 91.10 \\
NPOS \cite{tao2023nonparametric} & 84.72 & \textbf{74.68} & 88.84 & \underline{73.61} & 83.47 & 77.22 & 84.62 & \underline{71.00} & \textbf{88.68} & 92.22 & 89.04 & 91.41 & \textbf{89.86} & 89.44 & \underline{88.92} & 91.05 \\
\textbf{BER (Ours)} & \underline{\textbf{82.78}} & \underline{75.00} & \underline{86.44} & \textbf{73.51} & \underline{83.02} & \underline{76.52} & 82.81 & \textbf{70.98} & \underline{\textbf{88.42}} & 91.78 & \textbf{87.48} & \underline{90.87} & \underline{\textbf{88.28}} & 90.88 & \underline{\textbf{87.29}} & 90.58  \\
\hline
Average  & 83.68 & 78.30 & 87.44 & 77.95 & 84.23 & 80.54 & 83.06 & 73.37 & 90.30 & 91.99 & 89.14 & 92.70 & 90.97 & 91.23 & 90.27 & 91.36 \\

\hline

\hline
Average (All)  & 84.45 & 79.68 & 87.46 & 79.25 & 84.31 & 80.87 & 83.82 & 74.62 & 91.00 & 90.70 & 89.78 & 93.28 & 91.63 & 91.88 & 91.02 & 89.38 \\
\hline

\hline
\end{tabular}}
\label{Main_fine_5}
\end{table*}

\begin{table*}[ht]
\centering
\caption{Fine-grained results on \textbf{near- and far-OOD detection datasets} at the \textbf{step size of $k = 10$ for CIFAR 100} and at the \textbf{step size of $k = 100$ for ImangeNet1K}.
The \underline{\textbf{best}}, \textbf{second-best}, and \underline{third-best} performance per dataset are highlighted. The upper, middle, lower parts of the table are for AUC, AP, FPR performance, respectively.
} 
\scalebox{0.6}{
\begin{tabular}{c|cc|cc|cc|cc|cc|cc|cc|cc}
\hline

\hline
 \multirow{3}{*}{\textcolor{blue}{\textbf{AUC$\uparrow$}}} & \multicolumn{8}{c|}{ID Dataset: CIFAR100} & \multicolumn{8}{c}{ID Dataset: ImageNet1K} \\
 \cline{2-17}
 & \multicolumn{2}{c|}{iCaRL \cite{rebuffi2017icarl}} & \multicolumn{2}{c|}{BiC\cite{wu2019large}}  & \multicolumn{2}{c|}{WA \cite{zhao2020maintaining}} & \multicolumn{2}{c|}{FOSTER\cite{wang2022foster}} & \multicolumn{2}{c|}{iCaRL \cite{rebuffi2017icarl}} & \multicolumn{2}{c|}{BiC\cite{wu2019large}}  & \multicolumn{2}{c|}{WA \cite{zhao2020maintaining}} & \multicolumn{2}{c}{FOSTER\cite{wang2022foster}} \\
 \cline{2-17}
 & Near & Far & Near & Far & Near & Far & Near & Far & Near & Far & Near & Far & Near & Far & Near & Far \\
 \hline
\multicolumn{11}{l}{\textbf{-\textit{Average CIL accuracy}}} \\
 & \multicolumn{2}{c|}{60.08} & \multicolumn{2}{c|}{61.68} & \multicolumn{2}{c|}{65.88} & \multicolumn{2}{c|}{66.01} & \multicolumn{2}{c|}{44.44} & \multicolumn{2}{c|}{49.63} & \multicolumn{2}{c|}{52.22} & \multicolumn{2}{c}{52.29} \\   
\hline

\hline
\multicolumn{11}{l}{\textbf{-\textit{Post hoc-based OOD methods (w/o Fine-tuning)}}} \\
MSP \cite{hendrycks2016baseline} & 67.99 & 67.66 & 71.12 & 64.45 & 71.82 & 69.83 & 71.24 & 71.92 & 63.57 & 64.36 & 67.76 & 66.81 & 66.64 & 64.66 & 66.59 & 69.75 \\
ODIN \cite{liang2018enhancing} & 71.10 & 69.53 & 73.95 & 67.67 & 74.18 & 70.69 & 73.37 & 75.37 & \underline{67.06} & 71.13 & \textbf{70.54} & 72.19 & 67.62 & 68.34 & 67.61 & \textbf{75.49} \\
Energy \cite{liu2020energy} & 71.72 & 69.67 & 74.01 & 67.63 & \underline{74.62} & 72.23 & 74.12 & 76.54 & 64.87 & 67.17 & 69.35 & 71.79 & 65.96 & 67.78 & 66.80 & 74.59 \\
MaxLogit \cite{basart2022scaling} & 71.65 & 69.69 & \underline{74.04} & 67.29 & \underline{74.62} & 72.16 & 74.23 & 76.45 & 65.13 & 67.22 & \underline{69.50} & 70.81 & 66.80 & 66.64 & 67.36 & 73.94 \\
GEN \cite{liu2023gen} & 71.82 & 70.39 & \textbf{74.12} & 67.75 & \textbf{74.73} & 72.58 & 74.51 & \textbf{76.92} & 65.95 & 68.78 & 69.43 & 69.66 & \underline{\textbf{69.07}} & 67.52 & 67.92 & 73.40  \\
ReAct \cite{sun2021react} & 67.63 & 69.38 & 73.24 & 69.31 & 73.54 & 75.54 & \textbf{75.16} & 76.64 & 58.88 & 49.21 & 66.53 & 66.63 & 65.11 & 61.31 & 65.10 & 65.25 \\
KLM \cite{basart2022scaling} & 67.82 & 67.36 & 69.91 & 64.45 & 70.46 & 69.61 & 70.98 & 71.52 & 65.77 & 67.91 & 68.62 & 64.72 & 65.99 & 63.72 & \underline{\textbf{69.18}} & 72.31 \\
Relation \cite{kim2023neural} & 58.11 & 67.55 & 67.34 & 64.55 & 72.13 & \underline{\textbf{78.12}} & 67.40 & 74.55 & 63.33 & \textbf{72.47} & 68.51 & 70.30 & 67.61 & 67.31 & 65.67 & 72.42 \\
NNGuide \cite{park2023nearest} & 68.34 & \textbf{74.04} & 67.37 & 69.43 & 72.53 & \underline{77.13} & 74.45 & 76.63 & 64.88 & 69.85 & 69.43 & \underline{73.05} & 66.06 & \underline{\textbf{73.97}} & 63.58 & \underline{\textbf{75.72}} \\
\hline
Average  & 68.46 & 69.47 & 71.68 & 66.95 & 73.18 & 73.10 & 72.83 & 75.17 & 64.38 & 66.46 & 68.85 & 69.55 & 66.76 & 66.81 & 66.65 & 72.54 \\
\hline

\hline
\multicolumn{11}{l}{\textbf{-\textit{Fine-tuning-based OOD methods (w/ Fine-tuning)}}} \\
LogitNorm \cite{wei2022mitigating} & 71.88 & 69.72 & 73.52 & 69.20 & 73.56 & 71.28 & 73.86 & 75.18 & 64.30 & 66.85 & 68.25 & 72.13 & 65.31 & 68.28 & 66.36 & 70.78 \\
T2FNorm \cite{regmi2023t2fnorm} & 71.90 & 70.03 & 72.63 & \underline{\textbf{71.30}} & 72.93 & 71.50 & 73.75 & 75.09 & 65.19 & 67.91 & 68.18 & 72.23 & 66.08 & 70.00 & 66.10 & 72.36 \\
AUGMIX \cite{hendrycks2020augmix} & 72.50 & 69.82 & 73.19 & 69.58 & 73.63 & 71.35 & 74.33 & 74.96 & 64.94 & \underline{\textbf{75.91}} & 67.75 & \underline{\textbf{76.97}} & 65.71 & \textbf{72.87} & 67.23 & 69.06 \\
REGMIX \cite{pinto2023RegMixup} & \underline{73.12} & 70.64 & 72.97 & \underline{69.91} & 73.81 & 71.11 & \underline{74.74} & 75.51 & \underline{\textbf{67.86}} & 63.51 & 69.41 & 70.68 & \underline{67.70} & 66.97 & \textbf{68.96} & 70.36 \\
VOS \cite{du2022vos} & \textbf{73.41} & 71.76 & 67.55 & 66.23 & 65.31 & 70.08 & 73.80 & 74.97 & 63.69 & 71.47 & 64.94 & 70.13 & 64.14 & 67.94 & 67.19 & 72.57 \\
NPOS \cite{tao2023nonparametric} & 72.74 & \underline{73.48} & 66.17 & 67.46 & 74.00 & 74.37 & 74.28 & \underline{76.73} & 66.79 & 70.78 & 69.12 & 73.02 & 67.19 & 69.62 & 68.07 & 74.81 \\
\textbf{BER (Ours)} & \underline{\textbf{73.72}} & \underline{\textbf{74.39}} & \underline{\textbf{75.08}} & \textbf{70.20} & \underline{\textbf{75.43}} & \textbf{77.44}  & \underline{\textbf{75.20}} & \underline{\textbf{77.00}} &  \textbf{67.40} & \underline{71.60} & \underline{\textbf{70.94}} & \textbf{73.94} & \textbf{68.51} & \underline{70.16} & \underline{68.10} & \underline{75.41} \\
\hline
Average & 72.75 & 71.41 & 71.59 & 69.13 & 72.67 & 72.45 & 74.28 & 75.63 & 65.74 & 69.72 & 68.37 & 72.73 & 66.38 & 69.41 & 67.43 & 72.19\\
\hline 

\hline
Average (All)  & 70.34 & 70.32 & 71.64 & 67.90 & 72.96 & 72.81 & 73.46 & 75.37 & 64.98 & 67.88 & 68.64 & 70.94 & 66.59 & 67.94 & 66.99 & 72.39 \\
\hline

\hline
\hline

\hline
\multirow{2}{*}{\textcolor{blue}{\textbf{AP$\uparrow$}}}  & \multicolumn{2}{c|}{iCaRL \cite{rebuffi2017icarl}} & \multicolumn{2}{c|}{BiC\cite{wu2019large}}  & \multicolumn{2}{c|}{WA \cite{zhao2020maintaining}} & \multicolumn{2}{c|}{FOSTER\cite{wang2022foster}} & \multicolumn{2}{c|}{iCaRL \cite{rebuffi2017icarl}} & \multicolumn{2}{c|}{BiC\cite{wu2019large}}  & \multicolumn{2}{c|}{WA \cite{zhao2020maintaining}} & \multicolumn{2}{c}{FOSTER\cite{wang2022foster}} \\
 \cline{2-17}
 & Near & Far & Near & Far & Near & Far & Near & Far & Near & Far & Near & Far & Near & Far & Near & Far \\
 \hline
\multicolumn{11}{l}{\textbf{-\textit{Average CIL accuracy}}} \\
 & \multicolumn{2}{c|}{60.08} & \multicolumn{2}{c|}{61.68} & \multicolumn{2}{c|}{65.88} & \multicolumn{2}{c|}{66.01} & \multicolumn{2}{c|}{44.44} & \multicolumn{2}{c|}{49.63} & \multicolumn{2}{c|}{52.22} & \multicolumn{2}{c}{52.29} \\   
\hline

\hline
\multicolumn{11}{l}{\textbf{-\textit{Post hoc-based OOD methods (w/o Fine-tuning)}}} \\
MSP \cite{hendrycks2016baseline} & 62.92 & 77.47 & 66.42 & 75.73 & 66.93 & 78.47 & 66.22 & 79.97 & 24.27 & 21.29 & 26.87 & 22.73 & 25.14 & 19.96 & 25.38 & 25.00\\
ODIN \cite{liang2018enhancing} & 66.69 & 76.57 & \underline{69.50} & 75.62 & 69.80 & 77.20 & 68.69 & 80.05 & 26.30 & 27.44 & \underline{\textbf{28.84}} & 26.58 & 26.27 & 23.63 & 26.08 & \underline{\textbf{30.71}}  \\
Energy \cite{liu2020energy} & 67.16 & 76.79 & 69.45 & 75.66 & \underline{70.12} & 78.12 & 69.40 & 80.78 & 23.79 & 22.11 & 26.69 & 25.28 & 24.12 & 21.18 & 24.37 & 27.24 \\
MaxLogit \cite{basart2022scaling} & 67.02 & 76.91 & 69.47 & 75.73 & 70.03 & 78.37 & 69.41 & 80.90 & 24.18 & 22.41 & 27.17 & 24.60 & 24.90 & 20.69 & 25.00 & 26.82  \\
GEN \cite{liu2023gen} & 67.18 & 77.32 & \textbf{69.52} & 75.81 & \textbf{70.16} & 78.40 & \textbf{69.61} & \underline{81.13} & 24.77 & 23.06 & 27.41 & 23.32 & \underline{\textbf{28.65}} & 21.59 & 25.54 & 26.09 \\
ReAct \cite{sun2021react} & 64.32 & 78.22 & 68.84 & 77.00 & 69.24 & \underline{80.52} & 67.72 & \underline{\textbf{82.49}} & 21.25 & 13.04 & 25.82 & 20.62 & 25.94 & 18.00 & 25.85 & 20.11 \\
KLM \cite{basart2022scaling} & 62.71 & 78.06 & 64.73 & 77.12 & 64.53 & 79.28 & 65.79 & 80.35 & 26.60 & 24.14 & 28.25 & 21.19 & 25.73 & 20.66 & \textbf{30.06} & 27.00 \\
Relation \cite{kim2023neural} & 58.70 & \underline{79.28} & 65.41 & \textbf{78.51} & 68.32 & \underline{\textbf{80.80}} & 65.20 & 80.88 & 27.17 & \underline{30.24} & 28.16 & 25.52 & 28.12 & 24.41 & 24.89 & 27.99 \\
NNGuide \cite{park2023nearest} & 66.33 & 78.44 & 69.36 & \underline{78.26} & \underline{\textbf{70.19}} & 80.44 & 68.32 & 80.14 & 27.25 & 28.60 & \underline{28.61} & \underline{28.19} & \underline{28.16} & \underline{24.66} & 27.96 & \underline{30.08} \\
\hline 
Average  & 64.78 & 77.67 & 68.08 & 76.60 & 68.81 & 79.07 & 67.82 & 80.74 & 25.06 & 23.59 & 27.54 & 24.23 & 26.34 & 21.64 & 26.13 & 26.78 \\
\hline

\hline
\multicolumn{11}{l}{\textbf{\textit{Fine-tuning-based OOD methods (w/ Fine-tuning)}}}\\
LogitNorm \cite{wei2022mitigating} & 67.34 & 76.81 & 69.06 & 76.32 & 69.11 & 77.66 & 69.05 & 80.11 & 23.42 & 21.69 & 26.54 & \textbf{29.55} & 23.43 & 21.51 & 24.01 & 23.52 \\
T2FNorm \cite{regmi2023t2fnorm} & 67.28 & 77.07 & 68.19 & 77.89 & 68.33 & 78.02 & 68.92 & 80.24 & 24.12 & 22.60 & 25.87 & 27.36 & 24.15 & 22.77 & 24.25 & 24.90 \\
AUGMIX \cite{hendrycks2020augmix} & \underline{67.89} & 76.84 & 68.75 & 76.51 & 69.13 & 77.69 & \underline{69.52} & 79.90 & 26.98 & \underline{\textbf{33.75}} & 26.75 & \underline{\textbf{34.07}} & 25.41 & \underline{\textbf{26.99}} & 25.46 & 27.22 \\
REGMIX \cite{pinto2023RegMixup} & \textbf{68.56} & 77.26 & 68.57 & 76.82 & 69.36 & 77.65 & 69.12 & 80.13 & \underline{27.42} & 18.63 & 27.65 & 22.81 & 26.48 & 19.82 & 27.03 & 21.45  \\
VOS \cite{du2022vos} & 67.64 & 77.73 & 67.63 & 75.52 & 61.74 & 77.84 & 69.02 & 80.09 & 21.51 & 22.55 & 25.32 & 27.85 & 24.27 & 23.24 & 24.12 & 26.19  \\
NPOS \cite{tao2023nonparametric} & 67.88 & \textbf{79.63} & 66.17 & 76.09 & 69.46 & 79.98 & 69.49 & 81.11 & \textbf{27.55} & 25.66 & 27.12 & 26.85 & 27.95 & 23.07 & \underline{\textbf{30.18}} & 26.05 \\
\textbf{BER (Ours)} & \underline{\textbf{68.85}} & \underline{\textbf{79.73}} & \underline{\textbf{70.81}} & \underline{\textbf{78.81}} & 69.84 & \textbf{80.53}  & \underline{\textbf{69.64}} & \textbf{81.39} & \underline{\textbf{28.78}} & \textbf{30.85} & \textbf{28.62} & 27.81 & \textbf{28.38} & \textbf{25.27} & \underline{28.45} & \textbf{30.63} \\
\hline
Average  & 67.92 & 77.87 & 68.45 & 76.85 & 68.00 & 78.48 & 69.25 & 80.42 & 25.68 & 25.10 & 26.84 & 28.04 & 25.72 & 23.24 & 26.21 & 25.71 \\
\hline 

\hline
Average (All)  & 66.15 & 77.76 & 68.24 & 76.71 & 68.46 & 78.81 & 68.45 & 80.60 & 25.34 & 24.25 & 27.23 & 25.90 & 26.07 & 22.34 & 26.16 & 26.31\\
\hline

\hline
\hline

\hline
\multirow{2}{*}{\textcolor{blue}{\textbf{FPR$\downarrow$}} } & \multicolumn{2}{c|}{iCaRL \cite{rebuffi2017icarl}} & \multicolumn{2}{c|}{BiC\cite{wu2019large}}  & \multicolumn{2}{c|}{WA \cite{zhao2020maintaining}} & \multicolumn{2}{c|}{FOSTER\cite{wang2022foster}} & \multicolumn{2}{c|}{iCaRL \cite{rebuffi2017icarl}} & \multicolumn{2}{c|}{BiC\cite{wu2019large}}  & \multicolumn{2}{c|}{WA \cite{zhao2020maintaining}} & \multicolumn{2}{c}{FOSTER\cite{wang2022foster}} \\
 \cline{2-17}
 & Near & Far & Near & Far & Near & Far & Near & Far & Near & Far & Near & Far & Near & Far & Near & Far \\
 \hline
\multicolumn{11}{l}{\textbf{-\textit{Average CIL accuracy}}} \\
 & \multicolumn{2}{c|}{60.08} & \multicolumn{2}{c|}{61.68} & \multicolumn{2}{c|}{65.88} & \multicolumn{2}{c|}{66.01} & \multicolumn{2}{c|}{44.44} & \multicolumn{2}{c|}{49.63} & \multicolumn{2}{c|}{52.22} & \multicolumn{2}{c}{52.29} \\   
\hline

\hline
\multicolumn{11}{l}{\textbf{-\textit{Post hoc-based OOD methods (w/o Fine-tuning)}}} \\
MSP \cite{hendrycks2016baseline} & 88.31 & 88.34 & 85.34 & 87.80 & 85.38 & 86.37 & 85.97 & 84.43 & 90.01 & 89.42 & 89.26 & 88.99 & 91.20 & 93.00 & 90.37 & 87.20 \\
ODIN \cite{liang2018enhancing} & 84.48 & 80.66 & \underline{\textbf{82.06}} & 79.75 & \underline{81.78} & 79.37 & 82.86 & 72.09 & \underline{88.87} & \underline{\textbf{84.20}} & 87.74 & 86.88 & 89.16 & \underline{88.91} & 89.57 & \underline{\textbf{82.31}} \\
Energy \cite{liu2020energy} & 83.99 & 83.61 & 82.41 & 81.82 & \textbf{81.75} & 80.95 & 82.58 & 74.57 & 92.13 & 91.38 & 90.36 & 87.98 & 91.81 & 93.03 & 91.78 & 87.91 \\
MaxLogit \cite{basart2022scaling} & 84.57 & 84.39 & \underline{82.26} & 83.04 & 82.12 & 82.44 & 82.72 & 75.80 & 91.72 & 90.72 & 89.80 & 88.68 & 91.53 & 93.03 & 91.26 & 87.53 \\
GEN \cite{liu2023gen} & 84.26 & 84.33 & 82.57 & 82.48 & 81.81 & 81.22 & 82.52 & 75.20 & 91.18 & 89.97 & 89.21 & 90.88 & \underline{\textbf{87.00}} & 91.62 & 90.64 & 88.47 \\
ReAct \cite{sun2021react} & 85.04 & 84.12 & 82.62 & 82.59 & 82.22 & 81.62 & 83.73 & 75.17 & 93.09 & 97.23 & 89.67 & 90.24 & 90.02 & 92.80 & 90.45 & 92.61 \\
KLM \cite{basart2022scaling} & 88.24 & 87.89 & 86.78 & 88.09 & 88.03 & 87.46 & 86.18 & 85.76 & \textbf{88.79} & 87.27 & \textbf{86.30} & 88.93 & 89.25 & 90.26 & \underline{\textbf{88.33}} & 88.69 \\
Relation \cite{kim2023neural} & 90.91 & 88.90 & 84.06 & 79.24 & 90.86 & 87.12 & 84.22 & \underline{70.82} & 91.68 & 86.41 & 89.85 & 89.19 & 94.14 & 95.47 & 89.79 & 87.38 \\
NNGuide \cite{park2023nearest} & 84.23 & \underline{77.32} & \underline{82.26} & 77.06 & 82.06 & \textbf{72.97} & 83.57 & \underline{\textbf{70.27}} & 91.40 & 86.84 & \underline{87.44} & 87.11 & 89.02 & \textbf{88.42} & \underline{89.17} & 86.63 \\
\hline
Average & 86.00 & 84.40 & 83.37 & 82.43 & 84.00 & 82.17 & 83.82 & 76.01 & 90.99 & 89.27 & 88.85 & 88.76 & 90.35 & 91.84 & 90.15 & 87.64 \\
\hline

\hline
\multicolumn{11}{l}{\textbf{\textit{Fine-tuning-based OOD methods (w/ Fine-tuning)}}}\\
LogitNorm \cite{wei2022mitigating} & 83.94 & 83.70 & \underline{82.26} & 82.56 & 82.37 & 80.81 & 82.88 & 75.98 & 92.56 & 92.39 & 89.75 & \underline{\textbf{82.28}} & 92.84 & 93.78 & 92.58 & 92.02 \\
T2FNorm \cite{regmi2023t2fnorm} & 84.16 & 84.13 & 83.15 & 82.50 & 83.16 & 81.91 & 82.75 & 76.87 & 91.88 & 91.00 & 90.87 & \underline{85.53} & 91.79 & 91.92 & 91.57 & 89.71 \\
AUGMIX \cite{hendrycks2020augmix} & \textbf{83.61} & 84.07 & 82.48 & 83.17 & 82.33 & 81.14 & \underline{82.47} & 76.70 & 89.26 & \textbf{85.42} & 89.94 & \textbf{83.66} & 89.29 & 89.55 & 89.42 & 86.75 \\
REGMIX \cite{pinto2023RegMixup} & \underline{83.62} & 84.98 & 82.70 & 83.49 & 82.20 & 83.32 & \textbf{82.25} & 77.89 & 88.88 & 94.31 & 88.86 & 91.31 & 89.25 & 93.57 & 89.35 & 93.62 \\
VOS \cite{du2022vos} & \underline{\textbf{83.34}} & 82.12 & 85.81 & \underline{75.64} & 84.71 & 74.86 & 82.75 & 75.18 & 92.36 & 90.29 & 91.35 & 85.71 & 90.14 & 89.48 & 90.24 & \underline{85.72} \\
NPOS \cite{tao2023nonparametric} & 85.45 & \textbf{76.52} & 86.09 & \textbf{75.20} & 82.76 & \underline{73.78} & 85.58 & 71.21  & 88.92 & 88.38 & 87.60 & 89.88 & \underline{87.36} & 91.10 & 89.22 & 86.75 \\
\textbf{BER (Ours)} & \textbf{83.61} & \underline{\textbf{76.49}} & 85.30 & \underline{\textbf{73.99}} & \underline{\textbf{80.94}} &  \underline{\textbf{72.89}}  & \underline{\textbf{82.17}} & \textbf{70.78} & \underline{\textbf{88.31}} & \underline{85.91} & \underline{\textbf{85.32}} & 89.75 & \textbf{87.18} & \underline{\textbf{88.40}} & \textbf{88.45} & \textbf{85.39} \\
\hline
Average  & 83.96 & 81.72 & 83.97 & 79.51 & 82.64 & 78.39 & 82.98 & 74.94 & 90.31 & 89.67 & 89.10 & 86.87 & 89.69 & 91.11 & 90.12 & 88.57 \\
\hline

\hline
Average (All)  & 85.11 & 83.22 & 83.63 & 81.15 & 83.41 & 80.51 & 83.45 & 75.55 & 90.69 & 89.45 & 88.96 & 87.94 & 90.06 & 91.52 & 90.14 & 88.04 \\
\hline

\hline
\end{tabular}}
\label{Main_fine}
\end{table*}

\begin{table*}[t!]
\centering
\caption{Detailed results for those in Table \ref{Main_20} on \textbf{near- and far-OOD detection datasets} at the \textbf{step size of $k = 20$ for CIFAR 100} and at the \textbf{step size of $k = 200$ for ImangeNet1K}.
The \underline{\textbf{best}}, \textbf{second-best}, and \underline{third-best} performance per dataset are highlighted. The upper, middle, lower parts of the table are for AUC, AP, FPR performance, respectively.
} 
\scalebox{0.6}{
\begin{tabular}{c|cc|cc|cc|cc|cc|cc|cc|cc}
\hline

\hline
 \multirow{3}{*}{\textcolor{blue}{\textbf{AUC$\uparrow$}}} & \multicolumn{8}{c|}{ID Dataset: CIFAR100} & \multicolumn{8}{c}{ID Dataset: ImageNet1K} \\
 \cline{2-17}
 & \multicolumn{2}{c|}{iCaRL \cite{rebuffi2017icarl}} & \multicolumn{2}{c|}{BiC\cite{wu2019large}}  & \multicolumn{2}{c|}{WA \cite{zhao2020maintaining}} & \multicolumn{2}{c|}{FOSTER\cite{wang2022foster}} & \multicolumn{2}{c|}{iCaRL \cite{rebuffi2017icarl}} & \multicolumn{2}{c|}{BiC\cite{wu2019large}}  & \multicolumn{2}{c|}{WA \cite{zhao2020maintaining}} & \multicolumn{2}{c}{FOSTER\cite{wang2022foster}} \\
 \cline{2-17}
 & Near & Far & Near & Far & Near & Far & Near & Far & Near & Far & Near & Far & Near & Far & Near & Far \\
 \hline
\multicolumn{11}{l}{\textbf{\textit{Average CIL accuracy}}} \\
 & \multicolumn{2}{c|}{62.65} & \multicolumn{2}{c|}{64.14} & \multicolumn{2}{c|}{68.05} & \multicolumn{2}{c|}{68.75} & \multicolumn{2}{c|}{48.85} & \multicolumn{2}{c|}{53.30} & \multicolumn{2}{c|}{58.42} & \multicolumn{2}{c}{57.84} \\   
\hline

\hline
\multicolumn{11}{l}{\textbf{-\textit{Post hoc-based OOD methods (w/o Fine-tuning)}}} \\
MSP \cite{hendrycks2016baseline} & 69.42 & 67.86 & 71.44 & 68.87 & 73.14 & 72.11 & 72.33 & 71.99 & 66.14 & 68.23 & 70.66 & 72.84 & 69.22 & 73.77 & 69.35 & 76.01\\
ODIN \cite{liang2018enhancing} & 72.67 & 70.68 & 74.14 & 70.11 & 75.06 & 73.87 & 74.02 & 74.67 & \textbf{69.56} & 74.95 & \underline{72.70} & 80.40 & 69.88 & 75.29 & 70.83 & 79.00\\
Energy \cite{liu2020energy} & 73.62 & 71.75 & \underline{74.19} & 70.13 & \underline{75.32} & 75.17 & 75.64 & 76.39 & 67.58 & 73.41 & 71.44 & 80.20 & 68.36 & 74.95 & 69.92 & 78.17\\
MaxLogit \cite{basart2022scaling} & 73.58 & 71.69 & \textbf{74.31} & 70.25 & \textbf{75.45} & 75.10 & 75.65 & 76.30 & 67.88 & 73.09 & 71.87 & 78.76 & 69.29 & 74.66 & 70.50 & 78.46  \\
GEN \cite{liu2023gen} & 73.74 & 72.22 & \underline{\textbf{74.56}} & 70.59 & \underline{\textbf{75.58}} & 75.25 & \underline{75.88} & 76.88 & 68.77 & 72.94 & 72.27 & 78.04 & \underline{\textbf{70.97}} & 73.09 & \underline{71.27} & 79.00\\
ReAct \cite{sun2021react} & 70.33 & 72.22 & 73.64 & \underline{\textbf{73.72}} & 74.16 & \underline{\textbf{78.15}} & 74.81 & \underline{\textbf{78.60}} & 60.87 & 56.69 & 68.14 & 72.73 & 63.74 & 61.87 & 66.82 & 60.83 \\
KLM \cite{basart2022scaling} & 69.38 & 68.96 & 71.08 & 69.75 & 72.76 & 72.33 & 72.62 & 72.40 & 68.67 & 71.19 & 71.37 & 68.92 & 69.94 & 68.94 & \underline{\textbf{72.37}} & 75.89\\
Relation \cite{kim2023neural} & 60.26 & 67.92 & 67.36 & 69.97 & 72.90 & \textbf{77.89} & 66.61 & 73.61 & 69.03 & 75.66 & 72.58 & 80.21 & 69.52 & \underline{75.30} & 69.76 & \textbf{80.29} \\
NNGuide \cite{park2023nearest} & 71.30 & \underline{\textbf{75.56}} & 71.33 & 71.20 & 73.40 & 77.72 & 72.02 & \textbf{78.30} & 68.64 & \textbf{76.97} & \underline{\textbf{75.00}} & \textbf{83.84} & 68.53 & \underline{\textbf{78.64}} & 70.03 & 79.75\\
\hline
Average  & 70.48 & 70.98 & 72.45 & 70.51 & 74.20 & 75.29 & 73.29 & 75.46 & 67.46 & 71.46 & 71.78 & 77.33 & 68.83 & 72.95 & 70.09 & 76.38 \\
\hline

\hline
\multicolumn{11}{l}{\textbf{-\textit{Fine-tuning-based OOD methods (w/ Fine-tuning)}}} \\
LogitNorm \cite{wei2022mitigating} & 74.05 & 71.59 & 73.34 & 70.23 & 74.81 & 74.03 & 75.63 & 75.25 & 68.55 & 71.11 & 70.12 & 79.30 & 67.75 & 71.92 & 70.00 & 74.85 \\
T2FNorm \cite{regmi2023t2fnorm}   & 74.00 & 72.01 & 72.31 & 71.12 & 74.69 & 74.23 & 75.80 & 75.65 & 68.68 & 73.98 & 70.42 & 81.78 & 69.08 & 74.69 & 69.34 & 76.24\\
AUGMIX \cite{hendrycks2020augmix}   & \underline{\textbf{74.65}} & 71.43 & 73.06 & 69.52 & 75.30 & 73.87 & \underline{\textbf{76.22}} & 75.20 & 67.42 & \underline{\textbf{79.12}} & 71.80 & \underline{\textbf{86.56}} & 67.22 & \textbf{78.09} & 68.86 & \underline{\textbf{80.72}} \\
REGMIX \cite{pinto2023RegMixup}   & \textbf{74.54} & 71.84 & 72.97 & 70.94 & 75.19 & 74.57 & \textbf{76.00} & 75.53 & \underline{\textbf{69.64}} & 69.41 & 70.05 & 78.26 & 69.44 & 73.16 & 70.85 & 73.83 \\
VOS \cite{du2022vos}   & \underline{74.16} & 72.31 & 70.53 & 67.89 & 69.55 & 73.55 & 75.17 & 75.01 & 60.12 & 68.59 & 59.60 & 73.36 & 60.57 & 67.45 & 61.50 & 73.81 \\
NPOS \cite{tao2023nonparametric}  & 73.51 & \underline{74.58} & 69.51 & \underline{71.53} & 74.85 & 77.48 & 75.12 & 77.10 & 68.86 & 75.30 & 71.50 & \underline{83.78} & \underline{70.21} & 74.98 & \textbf{71.84} & 77.03 \\
\textbf{BER (Ours)} & 74.03 & \textbf{74.81} & 70.73 & \textbf{72.49}  & 74.36 & \underline{77.78} & 75.07 & \underline{77.24} & \underline{69.39} & \underline{76.87} & \textbf{73.15} & 83.08 & \textbf{70.61} & 74.72 & 70.89 & \underline{80.07} \\
\hline
Average  & 74.13 & 72.65 & 71.78 & 70.53 & 74.11 & 75.07 & 75.57 & 75.85 & 67.52 & 73.48 & 69.52 & 80.87 & 67.84 & 73.57 & 69.04 & 76.65  \\
\hline 

\hline
Average (All)  & 72.08 & 71.71 & 72.16 & 70.52 & 74.16 & 75.19 & 74.29 & 75.63 & 67.49 & 72.34 & 70.79 & 78.88 & 68.40 & 73.22 & 69.63 & 76.50 \\
\hline

\hline
\hline

\hline
\multirow{2}{*}{\textcolor{blue}{\textbf{AP$\uparrow$}} } & \multicolumn{2}{c|}{iCaRL \cite{rebuffi2017icarl}} & \multicolumn{2}{c|}{BiC\cite{wu2019large}}  & \multicolumn{2}{c|}{WA \cite{zhao2020maintaining}} & \multicolumn{2}{c|}{FOSTER\cite{wang2022foster}} & \multicolumn{2}{c|}{iCaRL \cite{rebuffi2017icarl}} & \multicolumn{2}{c|}{BiC\cite{wu2019large}}  & \multicolumn{2}{c|}{WA \cite{zhao2020maintaining}} & \multicolumn{2}{c}{FOSTER\cite{wang2022foster}} \\
 \cline{2-17}
 & Near & Far & Near & Far & Near & Far & Near & Far & Near & Far & Near & Far & Near & Far & Near & Far \\
 \hline
\multicolumn{11}{l}{\textbf{\textit{Average CIL accuracy}}} \\
 & \multicolumn{2}{c|}{62.65} & \multicolumn{2}{c|}{64.14} & \multicolumn{2}{c|}{68.05} & \multicolumn{2}{c|}{68.75} & \multicolumn{2}{c|}{48.85} & \multicolumn{2}{c|}{53.30} & \multicolumn{2}{c|}{58.42} & \multicolumn{2}{c}{57.84} \\   
\hline

\hline
\multicolumn{11}{l}{\textbf{-\textit{Post hoc-based OOD methods (w/o Fine-tuning)}}} \\
MSP \cite{hendrycks2016baseline} & 64.69 & 77.88 & 66.09 & 77.36 & 68.06 & 79.42 & 67.19 & 79.81 & 26.39 & 25.10 & 29.95 & 27.19 & 27.46 & 28.99 & 28.24 & 33.60  \\
ODIN \cite{liang2018enhancing} & 68.22 & 76.99 & \underline{68.99} & 75.95 & 70.69 & 78.59 & 69.52 & 79.47 & 29.08 & \textbf{32.24} & \underline{\textbf{31.33}} & \underline{37.17} & 29.02 & \textbf{31.30} & 29.01 & \underline{\textbf{36.89}}  \\
Energy \cite{liu2020energy} & 69.22 & 77.93 & 68.94 & 76.22 & 70.91 & 79.51 & 71.12 & 80.44 & 25.81 & 28.64 & 28.82 & 35.39 & 26.01 & 29.60 & 27.01 & 32.20  \\
MaxLogit \cite{basart2022scaling} & 69.11 & 78.06 & 68.98 & 76.48 & 70.94 & 79.75 & 71.03 & 80.58 & 26.38 & 28.39 & 29.68 & 32.27 & 27.10 & 29.57 & 27.87 & \underline{33.98}  \\
GEN \cite{liu2023gen} & 69.27 & 78.40 & \underline{\textbf{69.25}}  & 76.53 & 71.05 & 79.60 & 71.25 & 80.91 & 27.02 & 27.81 & 30.20 & 31.68 & \underline{30.52}  & 27.10 & 28.67 & \textbf{36.42}  \\
ReAct \cite{sun2021react} & 66.69 & 79.64 & 68.56 & 78.62 & 70.19 & 81.96 & 70.37 & \underline{82.46} & 22.29 & 17.14 & 28.76 & 26.53 & 25.82 & 20.59 & 28.13 & 18.63\\
KLM \cite{basart2022scaling} & 65.01 & 79.43 & 65.97 & \underline{79.45}  & 67.14 & 80.87 & 67.58 & 81.20 & \underline{29.76} & 26.16 & \underline{30.73} & 24.36 & 29.31 & 23.31 & \underline{\textbf{34.15}} & 28.59 \\
Relation \cite{kim2023neural} & 58.89 & \underline{81.39} & 65.14 & \underline{\textbf{80.71}} & 69.00 & \underline{\textbf{85.23}} & 65.80 & 82.40 & 28.20 & \underline{\textbf{32.80}} & \textbf{31.11} & 30.76 & \underline{\textbf{33.80}} & 27.37 & 28.26 & 28.12 \\
NNGuide \cite{park2023nearest} & 69.13 & \underline{\textbf{82.22}} & \textbf{69.16} & 79.13 & \textbf{71.16}  & \underline{83.33} & \textbf{71.80} & \underline{\textbf{82.63}} & \underline{\textbf{31.55}} & 28.89 & 25.52 & 29.69 & \textbf{32.09} & 29.09 &  \textbf{33.44} & 29.75  \\
\hline 
Average  & 66.69 & 79.10 & 67.90 & 77.83 & 69.90 & 80.92 & 69.52 & 81.10 & 27.39 & 27.46 & 29.57 & 30.56 & 29.01 & 27.44 & 29.42 & 30.91 \\
\hline

\hline
\multicolumn{11}{l}{\textbf{\textit{Fine-tuning-based OOD methods (w/ Fine-tuning)}}}\\
LogitNorm \cite{wei2022mitigating} & 69.65 & 77.80 & 68.22 & 76.29 & 70.29 & 78.78 & 71.03 & 79.71 & 26.76 & 26.25 & 27.62 & \textbf{37.77} & 25.39 & 27.70 & 27.39 & 27.84\\
T2FNorm \cite{regmi2023t2fnorm} & 69.53 & 78.22 & 67.29 & 77.07 & 70.04 & 79.23 & 71.12 & 80.11 & 26.88 & 29.40 & 27.96 & \underline{\textbf{41.73}} & 26.59 & 30.40 & 27.04 & 29.86 \\
AUGMIX \cite{hendrycks2020augmix} & \underline{70.16} & 77.73 & 67.84 & 76.12 & 70.68 & 78.69 & 71.57 & 79.62 & 29.41 & 31.62 & 28.77 & 28.31 & 27.78 & 27.66 & 29.20 & 29.28 \\
REGMIX \cite{pinto2023RegMixup} & \textbf{70.27} & 78.14 & 68.12 & 76.98 & \underline{71.06} & 79.33 & \underline{71.58} & 79.97 & 28.46 & 22.69 & 28.06 & 29.67 & 27.67 & 25.48 & 29.05 & 24.29\\
VOS \cite{du2022vos} & 69.66 & 78.12 & 66.03 & 75.72 & 65.00 & 79.31 & 70.44 & 79.58 & 22.86 & 26.04 & 21.60 & 31.07 & 22.16 & 26.43 & 23.24 & 29.48\\
NPOS \cite{tao2023nonparametric} & 70.01 & 80.50 & 64.89 & 78.62 & 70.61 & 82.14 & 70.67 & 82.01 & 27.82 & 30.66 & 28.15 & 30.45 & 27.37 & \underline{31.04} & 30.18 & 27.65  \\
\textbf{BER (Ours)} & \underline{\textbf{70.63}} & \textbf{81.41} & 68.86 & \textbf{79.98} & \underline{\textbf{71.41}} & \textbf{83.73} & \underline{\textbf{72.86}} & \textbf{82.60} & \textbf{30.26} & \underline{31.87} & 30.71 & 36.50 & 29.82 & \underline{\textbf{33.39}}  & \underline{30.66} & 32.12 \\
\hline
Average & 69.99 & 78.85 & 67.32 & 77.25 & 69.87 & 80.17 & 71.32 & 80.51 & 27.49 & 28.36 & 27.55 & 33.64 & 26.68 & 28.87 & 28.11 & 28.65 \\
\hline 

\hline
Average (All) & 68.13 & 78.99 & 67.65 & 77.58 & 69.89 & 80.59 & 70.31 & 80.84 & 27.43 & 27.86 & 28.69 & 31.91 & 27.99 & 28.06 & 28.85 & 29.92 \\
\hline

\hline
\hline

\hline
\multirow{2}{*}{\textcolor{blue}{\textbf{FPR$\downarrow$}}}  & \multicolumn{2}{c|}{iCaRL \cite{rebuffi2017icarl}} & \multicolumn{2}{c|}{BiC\cite{wu2019large}}  & \multicolumn{2}{c|}{WA \cite{zhao2020maintaining}} & \multicolumn{2}{c|}{FOSTER\cite{wang2022foster}} & \multicolumn{2}{c|}{iCaRL \cite{rebuffi2017icarl}} & \multicolumn{2}{c|}{BiC\cite{wu2019large}}  & \multicolumn{2}{c|}{WA \cite{zhao2020maintaining}} & \multicolumn{2}{c}{FOSTER\cite{wang2022foster}} \\
 \cline{2-17}
 & Near & Far & Near & Far & Near & Far & Near & Far & Near & Far & Near & Far & Near & Far & Near & Far \\
 \hline
\multicolumn{11}{l}{\textbf{\textit{Average CIL accuracy}}} \\
 & \multicolumn{2}{c|}{62.65} & \multicolumn{2}{c|}{64.14} & \multicolumn{2}{c|}{68.05} & \multicolumn{2}{c|}{68.75} & \multicolumn{2}{c|}{48.85} & \multicolumn{2}{c|}{53.30} & \multicolumn{2}{c|}{58.42} & \multicolumn{2}{c}{57.84} \\   
\hline

\hline
\multicolumn{11}{l}{\textbf{-\textit{Post hoc-based OOD methods (w/o Fine-tuning)}}} \\
MSP \cite{hendrycks2016baseline}  & 86.59 & 87.49 & 86.66 & 88.30 & 84.62 & 84.80 & 85.18 & 84.94 & 88.25 & 85.89 & 86.92 & 86.19 & 89.12 & 83.14 & 87.95 & 77.41 \\
ODIN \cite{liang2018enhancing}   & 83.94 & 79.84 & 84.45 & 78.70 & 82.02 & 76.10 & 82.65 & 73.77 & \textbf{85.99} & \underline{\textbf{78.86}} & \textbf{85.81} & \underline{76.03} & 86.75 & \underline{80.31} & \textbf{87.37} & 74.06 \\
Energy \cite{liu2020energy}   & 82.74 & 81.62 & 84.56 & 81.39 & 82.06 & 77.47 & 81.72 & 76.95 & 90.62 & 83.67 & 88.72 & 77.34 & 90.36 & 82.46 & 90.03 & 81.20 \\
MaxLogit \cite{basart2022scaling}   & 82.62 & 81.99 & 84.72 & 83.23 & 81.85 & 79.10 & 81.44 & 77.54 & 89.86 & 84.00 & 87.78 & 82.40 & 89.54 & 82.69 & 89.14 & 78.44 \\
GEN \cite{liu2023gen}   & 82.86 & 82.23 & \underline{84.43} & 82.28 & \underline{81.67} & 78.23 & \underline{81.24} & 76.96 & 89.24 & 84.32 & 87.31 & 82.91 & \textbf{85.46} & 86.07 & 88.44 & 76.02 \\
ReAct \cite{sun2021react}  & 84.18 & 82.40 & \underline{\textbf{84.02}} & 82.30 & \underline{\textbf{81.51}} & 76.98 & 81.86 & 77.35 & 92.37 & 92.98 & 86.31 & 85.15 & 88.68 & 87.47 & 87.89 & 91.61\\
KLM \cite{basart2022scaling}   & 86.34 & 85.12 & 86.48 & 86.89 & 86.24 & 85.97 & 84.75 & 84.21 & \underline{\textbf{84.68}} & 85.60 & \underline{\textbf{85.24}} & 86.67 & \underline{86.42} & 89.64 & \underline{\textbf{80.57}} & 85.16 \\
Relation \cite{kim2023neural}   & 87.14 & 76.55 & 84.52 & 76.94 & 83.35 & 74.51 & 83.91 & \underline{72.84} & 89.50 & \underline{82.30} & 86.36 & 77.31 & 93.46 & 91.33 & 88.11 & 73.72 \\
NNGuide \cite{park2023nearest}   & 82.70 & \textbf{74.23} & \textbf{84.31} & 76.17 & 81.69 & \underline{69.83} & 85.19 & \underline{\textbf{71.60}} & 91.91 & \textbf{80.27} & 91.24 & 77.11 & 91.99 & \underline{\textbf{76.72}} & \underline{87.52} & \textbf{73.19} \\
\hline
Average  & 84.35 & 81.27 & 84.91 & 81.80 & 82.78 & 78.11 & 83.10 & 77.35 & 89.16 & 84.21 & 87.30 & 81.23 & 89.09 & 84.43 & 87.45 & 78.98  \\
\hline

\hline
\multicolumn{11}{l}{\textbf{\textit{Fine-tuning-based OOD methods (w/ Fine-tuning)}}}\\
LogitNorm \cite{wei2022mitigating}  & \underline{82.61} & 82.17 & 84.77 & 83.31 & 82.23 & 78.12 & 81.87 & 77.98 & 90.18 & 87.16 & 89.59 & \textbf{72.99} & 91.49 & 85.73 & 90.31 & 87.28\\
T2FNorm \cite{regmi2023t2fnorm}   & 82.63 & 82.28 & 85.28 & 83.05 & 82.63 & 79.67 & 81.59 & 78.60 & 89.78 & 82.85 & 89.25 & \underline{\textbf{68.00}} & 90.11 & 81.85 & 89.61 & 83.92 \\
AUGMIX \cite{hendrycks2020augmix}   & \textbf{82.34} & 82.77 & 84.96 & 84.02 & 81.91 & 78.99 & 81.34 & 78.74 & 89.08 & 85.59 & 88.27 & 77.35 & 92.17 & \textbf{77.26} & 88.27 & 74.44  \\
REGMIX \cite{pinto2023RegMixup}   & \underline{\textbf{81.88}} & 83.27 & 85.08 & 84.43 & 81.84 & 79.75 & \underline{\textbf{81.07}} & 79.61 &  88.40 & 92.04 & 88.75 & 85.22 & 88.44 & 87.50 & 87.88 & 90.78\\
VOS \cite{du2022vos}   & 82.83 & 80.67 & 85.24 & \underline{74.80} & 86.31 & 71.70 & 82.50 & 76.54 & 91.28 & 86.32 & 88.06 & 88.46 & 91.31 & 84.44 & 90.04 & 84.19\\
NPOS \cite{tao2023nonparametric}  & 85.94 & \underline{75.37} & 88.36 & \underline{\textbf{73.47}} & 81.89 & \textbf{66.84} & 84.47 & 73.14 & 87.81 & 86.00 & 86.91 & 78.49 & 88.48 & 86.62 & 87.92 & \underline{73.53}  \\
\textbf{BER (Ours)} & 82.70 & \underline{\textbf{73.77}} & 85.96 & \textbf{73.97} & \textbf{81.59} & \underline{\textbf{66.79}}  & \textbf{81.14} & 71.99 & \underline{87.71} & 85.57 & \underline{86.06} & 76.91 & \underline{\textbf{85.40}} & 87.91 & 87.63 & \underline{\textbf{72.72}} \\
\hline
Average  & 82.99 & 80.04 & 85.66 & 79.58 & 82.63 & 74.55 & 82.00 & 76.66 & 89.18 & 86.50 & 88.13 & 78.20 & 89.63 & 84.47 & 88.81 & 80.98  \\
\hline

\hline
Average (All)  & 83.75 & 80.74 & 85.24 & 80.83 & 82.71 & 76.55 & 82.62 & 77.05 & 89.17 & 85.21 & 87.66 & 79.91 & 89.32 & 84.45 & 88.04 & 79.85 \\
\hline

\hline
\end{tabular}}
\label{Main_fine_20}
\end{table*}
\FloatBarrier

\section{More Results w.r.t. Increasing Incremental Steps}
\label{morequantity}
\begin{figure}[ht]
  \centering
  \subfloat[]
  {\includegraphics[width=0.30\textwidth]{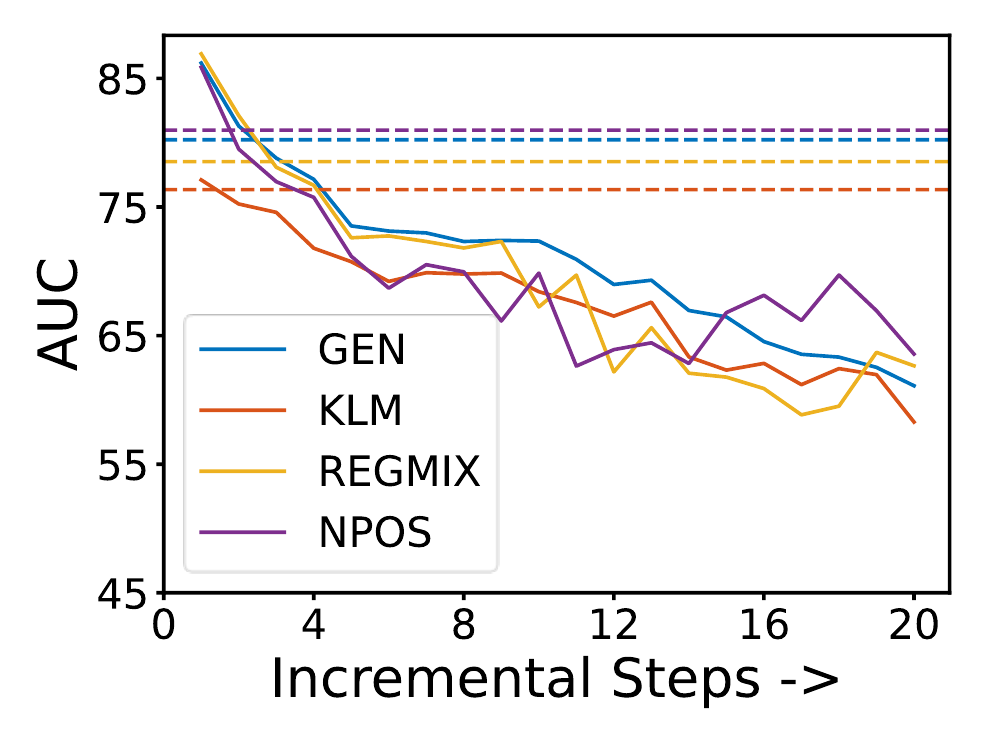} \label{Incremental_bic}}
  \subfloat[]
  {\includegraphics[width=0.30\textwidth]{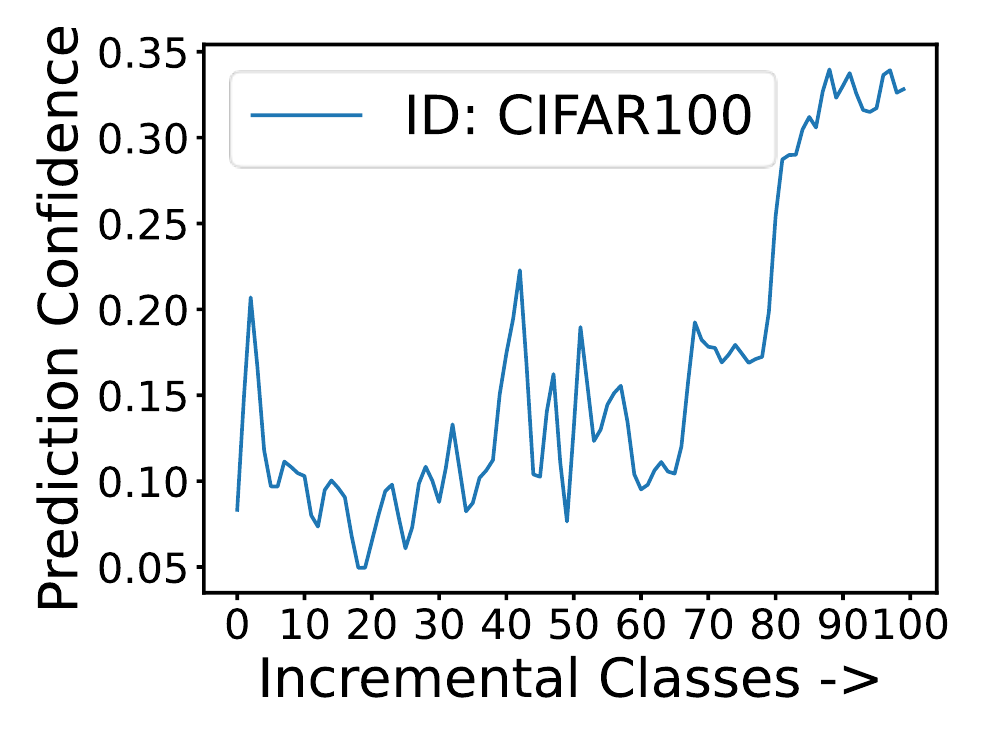}  \label{ID_bic}}
  \subfloat[]
  {\includegraphics[width=0.30\textwidth]{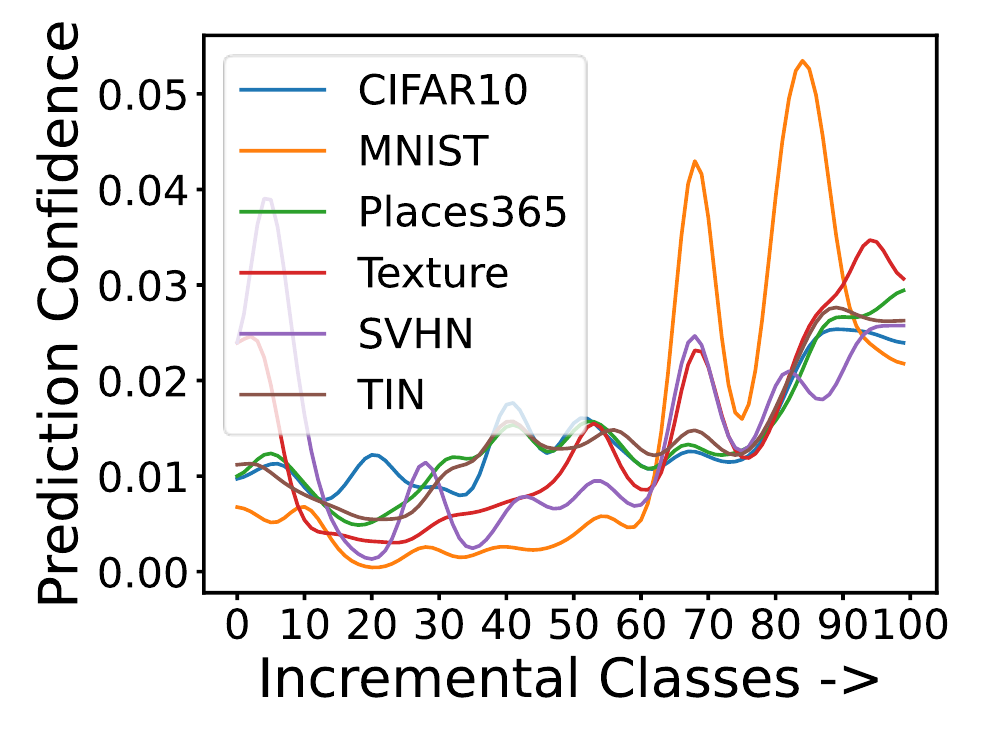}  \label{OOD_bic}}
  \caption{Qualitative results of the CIL model BiC \cite{wu2019large} with CIFAR100 \cite{krizhevsky2009learning}.
  \textbf{(a)} 
  All four representative OOD detection methods experience a decreased AUC performance with increasing incremental steps, compared to themselves working on the full training data of all steps.
  \textbf{(b)} Mean prediction confidence of BiC on test samples from all incremental classes.
  \textbf{(c)} Mean prediction confidence of BiC classifying six OOD datasets into one of the ID classes.}
  \label{qua_bic}
\end{figure}

\begin{figure}[ht]
  \centering
  \subfloat[]
  {\includegraphics[width=0.28\textwidth]{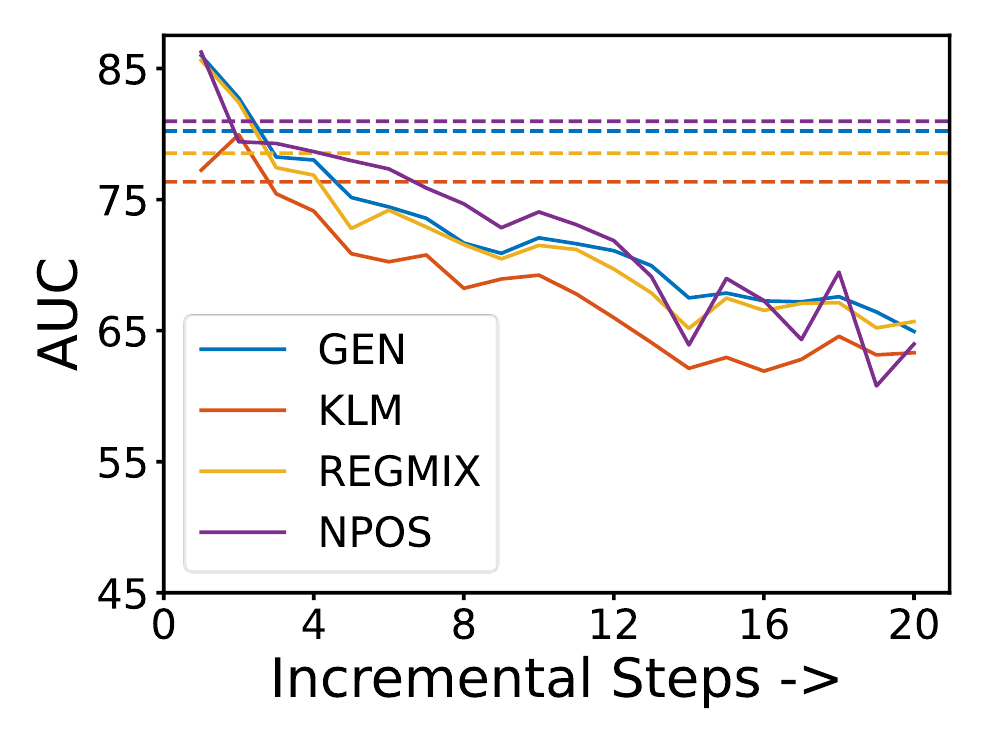} \label{Incremental_wa}}
  \subfloat[]
  {\includegraphics[width=0.28\textwidth]{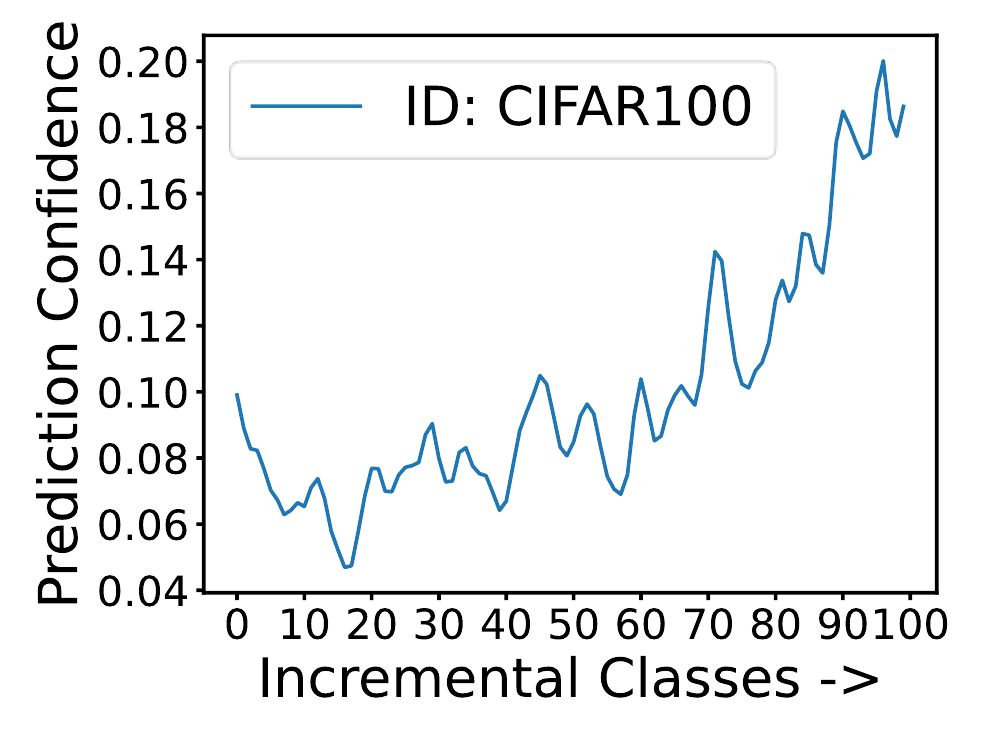}  \label{ID_wa}}
  \subfloat[]
  {\includegraphics[width=0.28\textwidth]{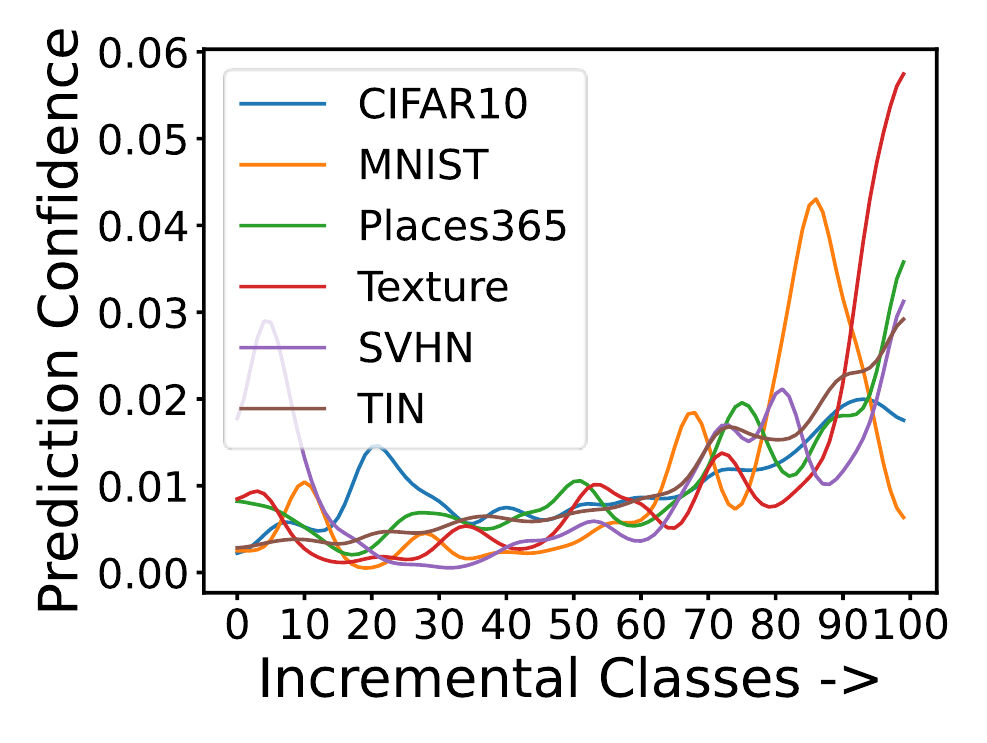}  \label{OOD_wa}}
  \caption{Qualitative results of the CIL model WA \cite{zhao2020maintaining} with CIFAR100 \cite{krizhevsky2009learning}.
  \textbf{(a)} 
  All four representative OOD detection methods experience a decreased AUC performance with increasing incremental steps, compared to themselves working on the full training data of all steps.
  \textbf{(b)} Mean prediction confidence of WA on test samples from all incremental classes.
  \textbf{(c)} Mean prediction confidence of WA classifying six OOD datasets into one of the ID classes.}
  \label{qua_wa}
\end{figure}

\begin{figure}[ht]
  \centering
  \subfloat[]
  {\includegraphics[width=0.28\textwidth]{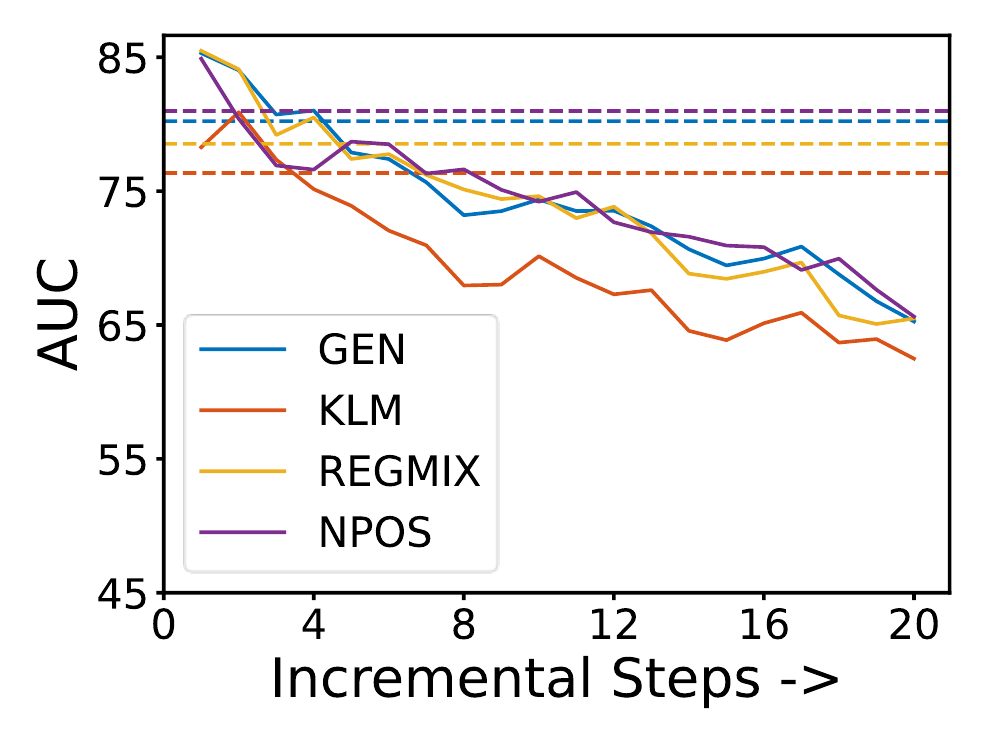}} \label{Incremental_foster}
  \subfloat[]
  {\includegraphics[width=0.28\textwidth]{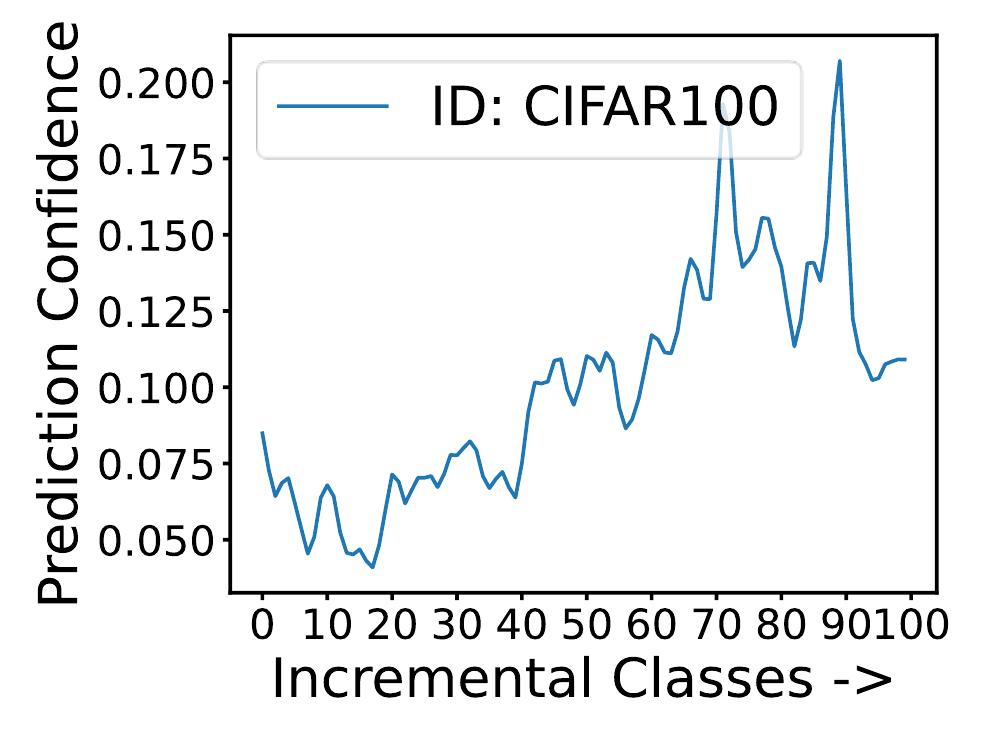}  \label{ID_foster}}
  \subfloat[]
  {\includegraphics[width=0.28\textwidth]{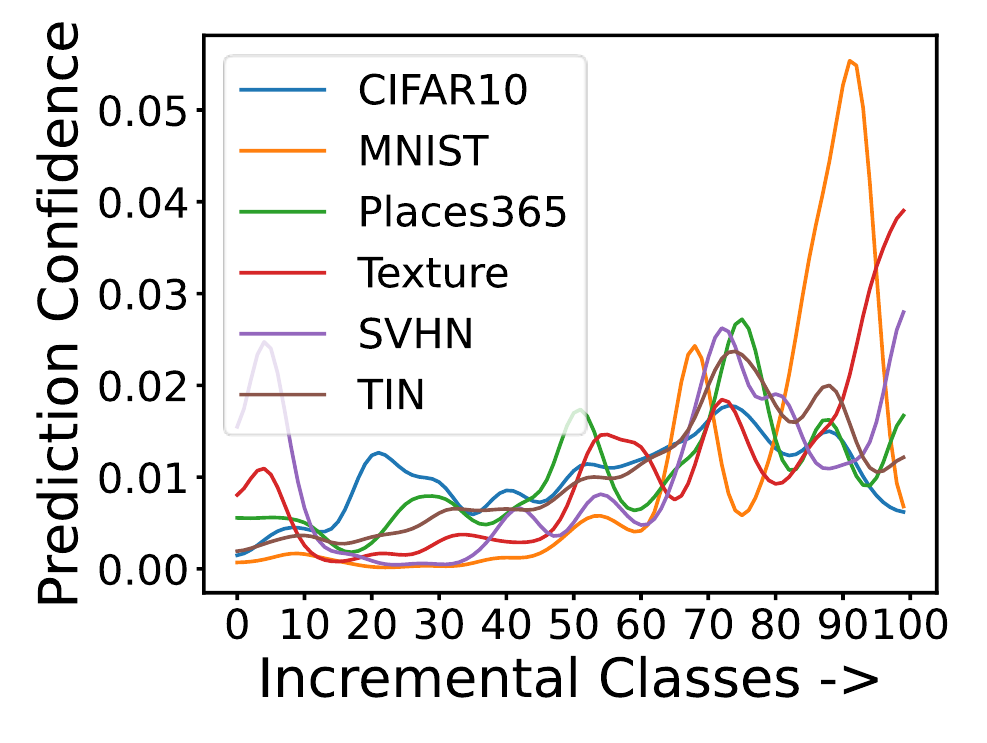}  \label{OOD_foster}}
  \caption{Qualitative results of the CIL model FOSTER\cite{wang2022foster} with CIFAR100 \cite{krizhevsky2009learning}.
  \textbf{(a)} 
  All four representative OOD detection methods experience a decreased AUC performance with increasing incremental steps, compared to themselves working on the full training data of all steps.
  \textbf{(b)} Mean prediction confidence of FOSTER on test samples from all incremental classes.
  \textbf{(c)} Mean prediction confidence of FOSTER classifying six OOD datasets into one of the ID classes.}
  \label{qua_foster}
\end{figure}

\begin{figure}[t!]
  \centering
  \subfloat[BiC \cite{wu2019large}]
  {\includegraphics[width=0.28\textwidth]{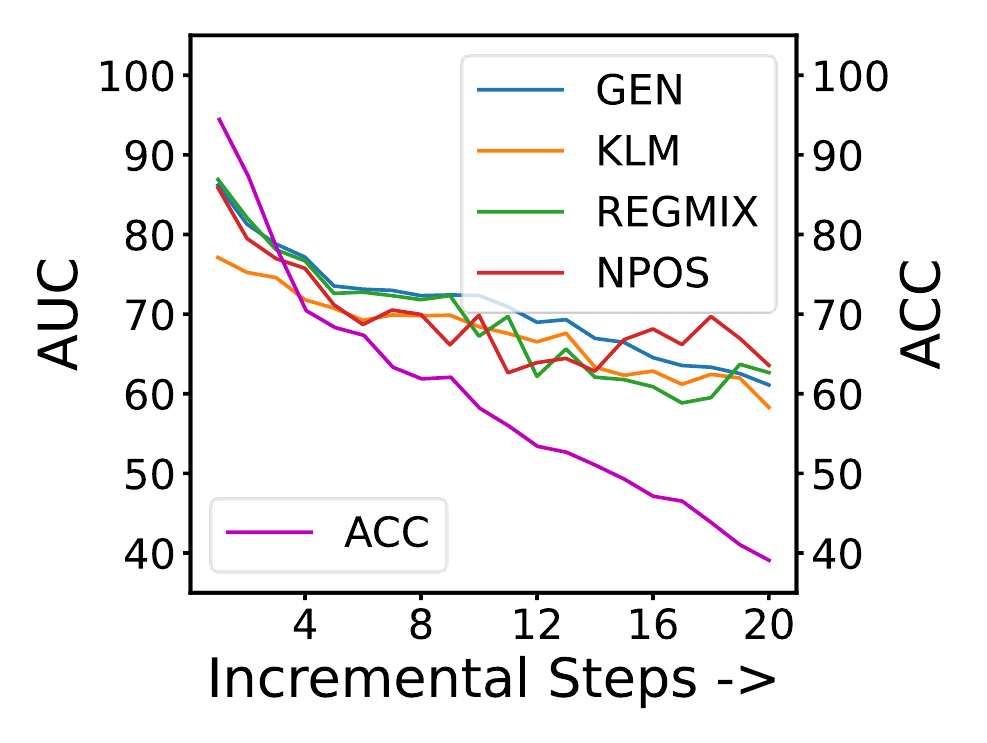}   \label{ACC_bic}}
  \subfloat[WA \cite{zhao2020maintaining}]
  {\includegraphics[width=0.28\textwidth]{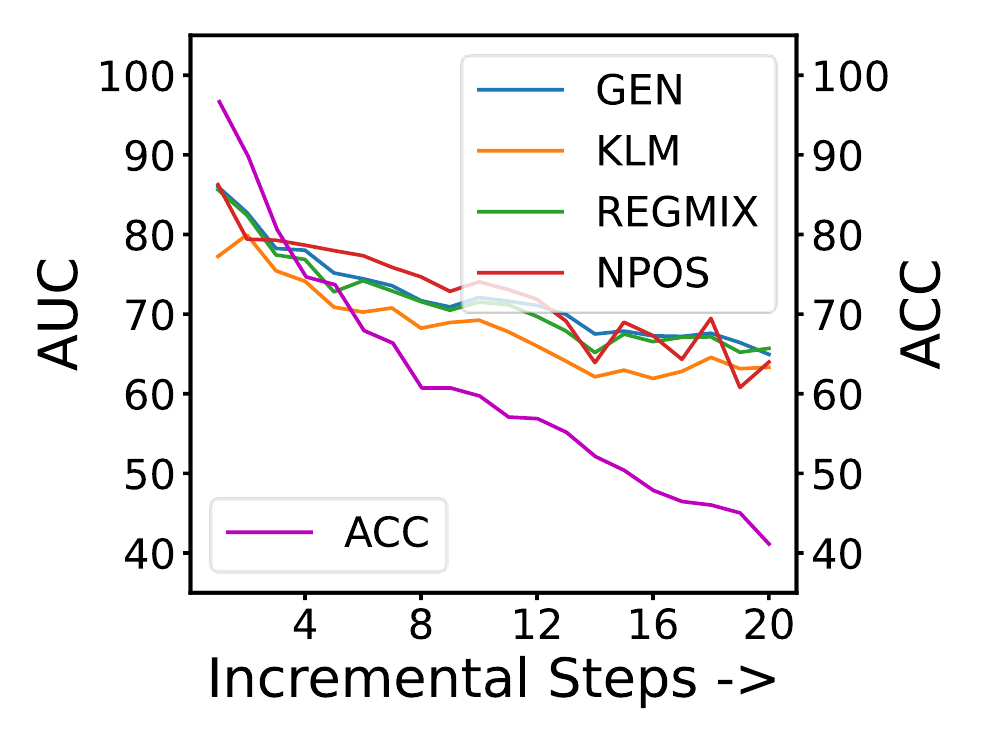}   \label{ACC_wa}}
  \subfloat[FOSTER\cite{wang2022foster}]
  {\includegraphics[width=0.28\textwidth]{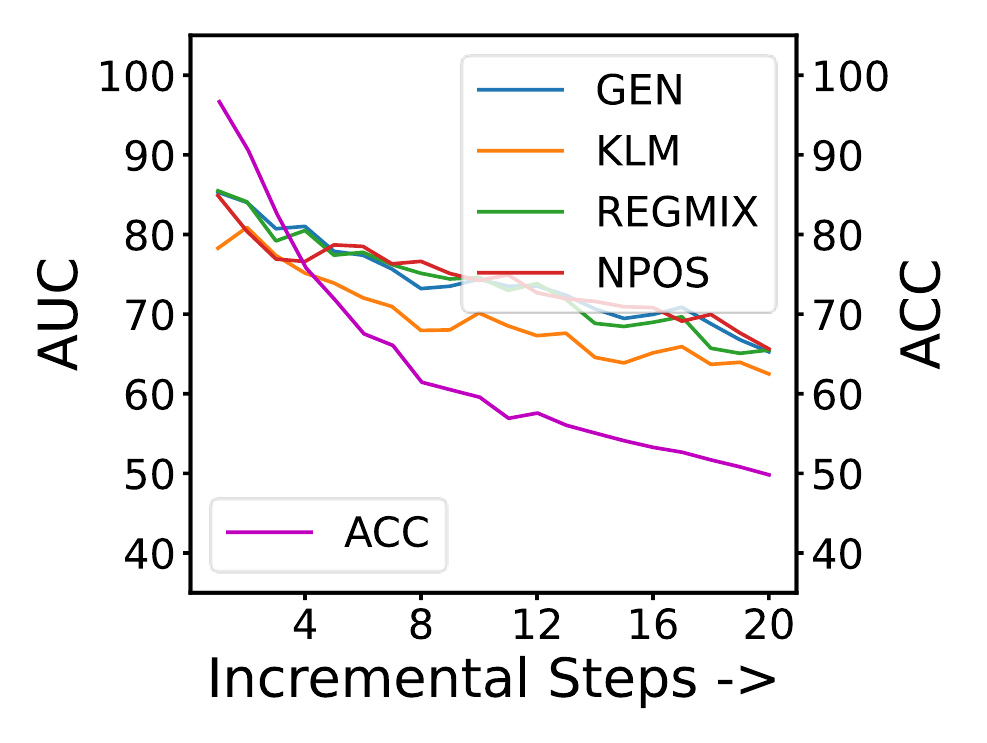}   \label{ACC_foster}}
  \caption{
  Average performance of four representative OOD methods on six OOD datasets at each incremental step, where the different CIL models are used. 
  ACC is the accuracy of CIL models on CIFAR100 at each step.
  }
  \label{ACC_more}
\end{figure}

\begin{figure}[t!]
  \centering
  \subfloat[GEN\cite{liu2023gen}]
  {\includegraphics[width=0.28\textwidth]{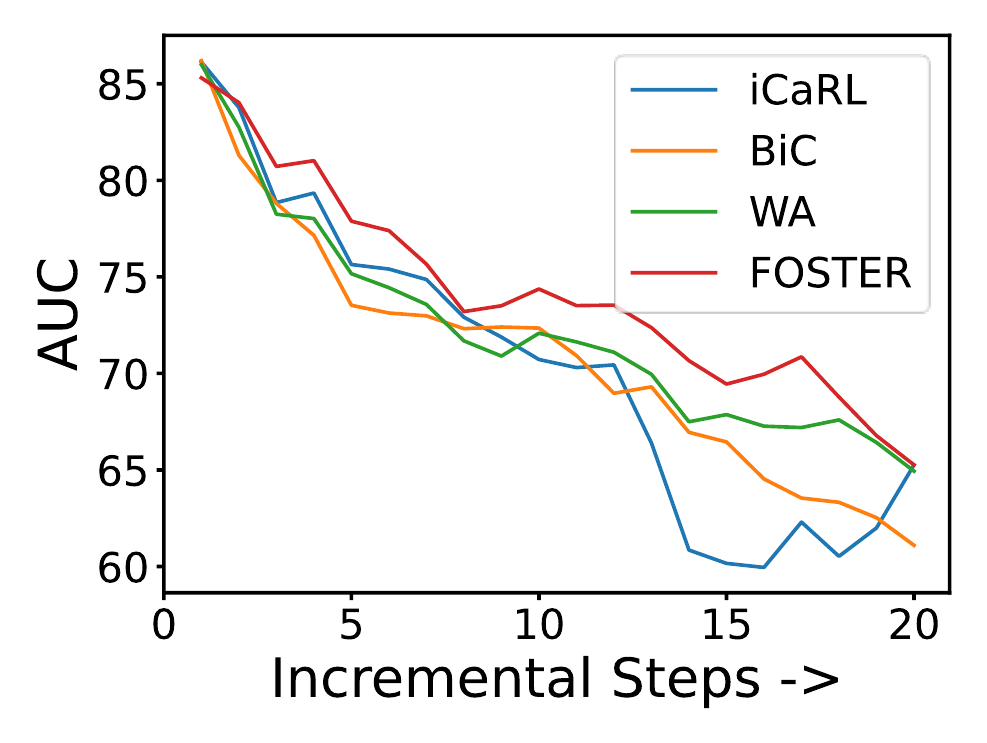}   \label{low_bound-gen}}
  \subfloat[KLM\cite{basart2022scaling}]
  {\includegraphics[width=0.28\textwidth]{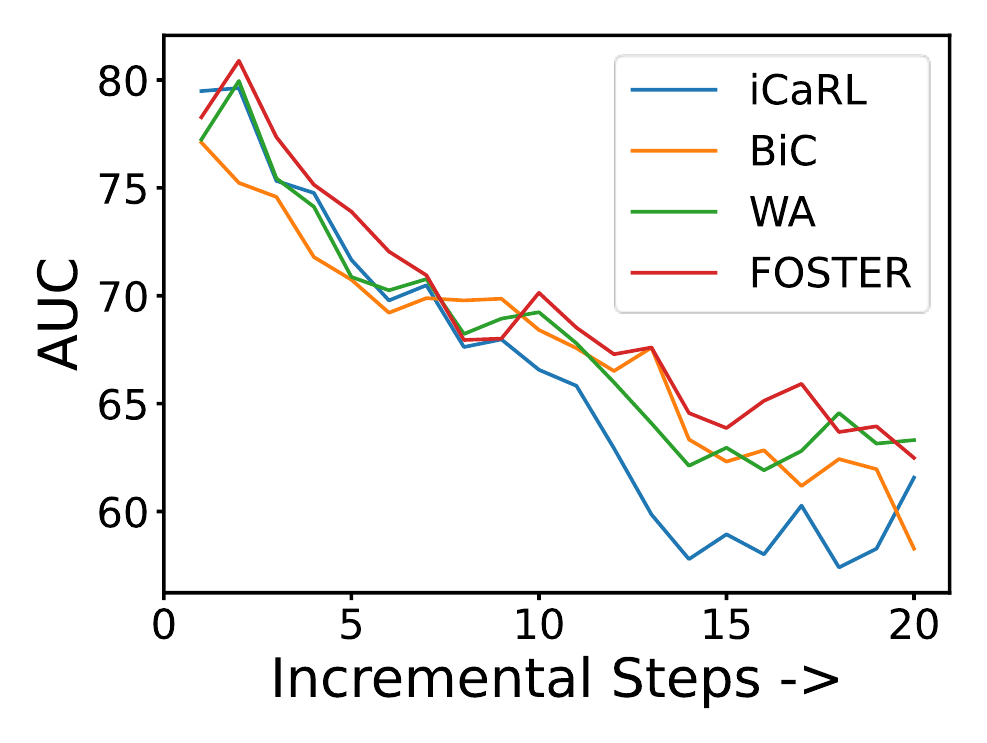}   \label{low_bound-klm}}
  \subfloat[NPOS\cite{tao2023nonparametric}]
  {\includegraphics[width=0.28\textwidth]{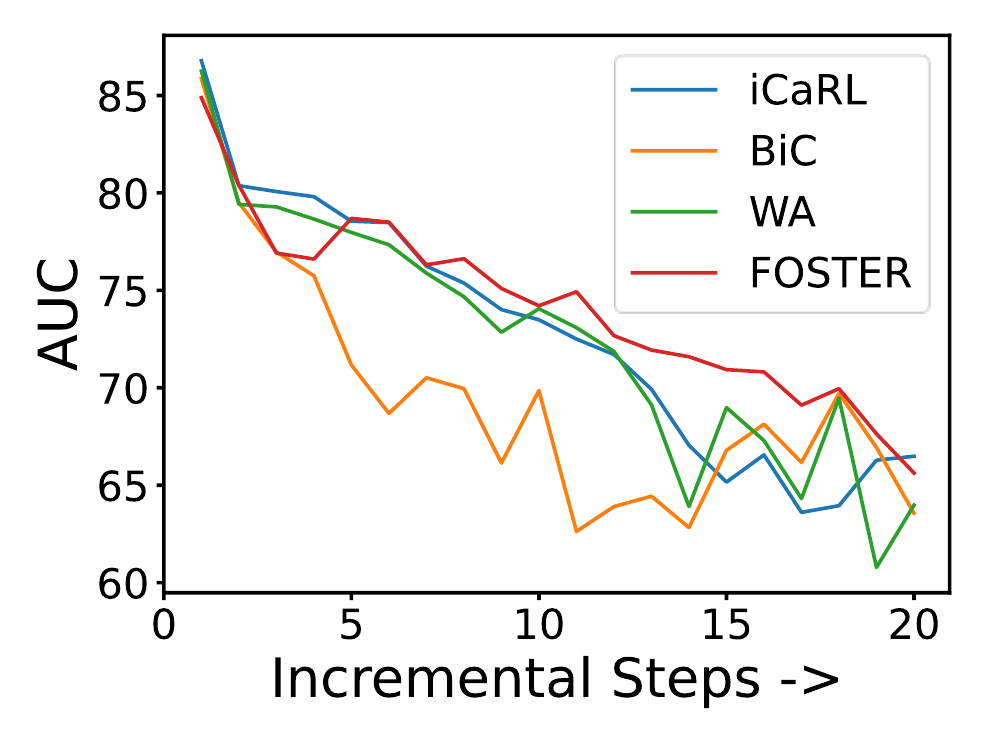}   \label{low_bound-npos}}
  \caption{
  Average performance of CIL models with different OOD detectors on six OOD datasets at each incremental step on CIFAR100.
  }
  \label{low_bound_more}
\end{figure}

In this section, we provide more results for the different baselines w.r.t. increasing incremental steps in our paper.
Particularly, following the results of iCaRL \cite{rebuffi2017icarl} with CIFAR100 \cite{krizhevsky2009learning} in our paper, we provide the results based on Bic \cite{wu2019large} in Fig. \ref{qua_bic}, WA \cite{zhao2020maintaining} in Fig. \ref{qua_wa}, FOSTER\cite{wang2022foster} in Fig. \ref{qua_foster}.
Furthermore, we show relationship between ACC and OOD detection performance on these three CIL models in Fig. \ref{ACC_more}.
We also show the average performance of all four CIL models with more OOD detectors (GEN \cite{liu2023gen}, KLM \cite{basart2022scaling}, and NPOS \cite{tao2023nonparametric}) in Fig. \ref{low_bound_more}.

\section{The OpenCIL Benchmark framework and BER Algorithm}

The full framework of the OpenCIL Benchmark for post-hoc-based OOD methods are given in Algorithm \ref{alg:framework_post}, and for fine-tuning-based OOD methods are given in Algorithm \ref{alg:framework_finetune} below.
Furthermore, the algorithm of our proposed baseline BER are given in Algorithm \ref{alg:BER}.

\begin{algorithm}[ht]
\caption{: Post-hoc-based OOD Methods Framework in OpenCIL Benchmark}
\label{alg:framework_post}
\textbf{Class Incremental Training} \\
\textbf{Data}: A data memory $M$;  A sequence of $c$ tasks ID data $T = \{T_1, T_2,..., T_c\}$, $T_t = (X_t^{train}, X_t^{test}, Y_t)$
\begin{algorithmic}[1] 
\FOR{each task}
\STATE Obtain $t$-th $(1 \le t \le c)$ task training ID data $T_t^{train} = (X_t^{train}, Y_t) \cup M$
\FOR{each iteration} 
\STATE Sample a mini-batch of ID training data from $ T_t^{train} $
\STATE Perform different CIL algorithms
\ENDFOR
\STATE Update data memory $M$       $\;\;\;\;$\textit{//*for replay-based CIL models only*}
\STATE Save current CIL model $\theta_t{(\cdot)}$
\ENDFOR
\end{algorithmic} 
\textbf{Output}: A well-trained CIL model $\theta{(\cdot)} = \{\theta_1{(\cdot)}, \theta_2{(\cdot)},..., \theta_c{(\cdot)}\}$ 

\hrulefill \\
\textbf{Inference} \\
\textbf{Input}: A well-trained CIL model $\theta{(\cdot)} = \{\theta_1{(\cdot)}, \theta_2{(\cdot)},..., \theta_c{(\cdot)}\}$ \\
\textbf{Data}:  A sequence of $c$ tasks OOD data $X^{ood} = \{X_1^{ood}, X_2^{ood},..., X_c^{ood}\} $; A sequence of $c$ tasks ID data $T = \{T_1, T_2,..., T_c\}$, $T_t = (X_t^{train}, X_t^{test}, Y_t)$
\begin{algorithmic}[1] 
\FOR{each task}
\STATE Obtain $t$-th $(1 \le t \le c)$ task testing ID data $T_t^{test} = X_1^{test} \cup X_2^{test} \cup ... \cup X_t^{test}$
\STATE Obtain $t$-th $(1 \le t \le c)$ task testing OOD data $X_t^{ood}$
\STATE Perform ID classification on CIL model $\theta_t{(\cdot)}$ based on $ T_t^{test} $
\STATE Perform different post-hoc OOD detection methods on CIL model $\theta_t{(\cdot)}$ based on $ T_t^{test} \cup X_t^{ood}$
\ENDFOR
\end{algorithmic}
\textbf{Output}: Average incremental accuracy; Average OOD detection performance 
\end{algorithm}

\begin{algorithm}[ht]
\caption{: Fine-tuning-based OOD Methods Framework in OpenCIL Benchmark}
\label{alg:framework_finetune}
\textbf{Class Incremental Training} \\
\textbf{Data}: A data memory $M$;  A sequence of $c$ tasks ID data $T = \{T_1, T_2,..., T_c\}$, $T_t = (X_t^{train}, X_t^{test}, Y_t)$
\begin{algorithmic}[1] 
\FOR{each task}
\STATE Obtain $t$-th $(1 \le t \le c)$ task training ID data $T_t^{train} = (X_t^{train}, Y_t) \cup M$
\FOR{each iteration} 
\STATE Sample a mini-batch of ID training data from $ T_t^{train} $
\STATE Perform different CIL algorithms
\ENDFOR
\STATE Update data memory $M$       $\;\;\;\;$\textit{//*for replay-based CIL models only*}
\STATE Save current CIL model $\theta_t{(\cdot)}$
\ENDFOR
\end{algorithmic} 
\textbf{Output}: A well-trained CIL model $\theta{(\cdot)} = \{\theta_1{(\cdot)}, \theta_2{(\cdot)},..., \theta_c{(\cdot)}\}$ 

\hrulefill \\
\textbf{OOD Method - Fine-tuning} \\
\textbf{Input}: A well-trained CIL model $\theta{(\cdot)} = \{\theta_1{(\cdot)}, \theta_2{(\cdot)},..., \theta_c{(\cdot)}\}$ \\
\textbf{Data}: A data memory $M$;  A sequence of $c$ tasks ID data $T = \{T_1, T_2,..., T_c\}$, $T_t = (X_t^{train}, X_t^{test}, Y_t)$
\begin{algorithmic}[1] 
\FOR{each task}
\STATE Obtain $t$-th $(1 \le t \le c)$ task training ID data $T_t^{train} = (X_t^{train}, Y_t) \cup M$
\STATE Obtain $t$-th CIL model $\theta_t{(\cdot)} = \{ \phi_t{(\cdot)}, h_t(\cdot) \}$ which is composed of a feature extractor $\phi_t{(\cdot)}$ and a original classifier $h_t(\cdot)$
\STATE Initialize an extra classifier $f_t(\cdot)$
\FOR{each iteration} 
\STATE freeze the $\phi_t{(\cdot)}$ and $h_t(\cdot)$
\STATE Sample a mini-batch of ID training data from $ T_t^{train} $
\STATE Perform different training-time OOD detection method to finetune this extra classifier $f_t(\cdot)$ only on top of $\phi_t{(\cdot)}$.
\ENDFOR
\STATE Update data memory $M$       $\;\;\;\;$\textit{//*for replay-based CIL models only*}
\STATE Save current CIL model $\theta_t{(\cdot)}$ with extra classifier $f_t(\cdot)$
\ENDFOR
\end{algorithmic} 
\textbf{Output}: A well-trained CIL model (the same as \textbf{Input}) with a finetuned classifier $ (\theta{(\cdot)}, f(\cdot)) = \{(\theta_1{(\cdot)}, f_1(\cdot)), (\theta_2{(\cdot)}, f_2(\cdot)), ..., (\theta_c{(\cdot), f_c(\cdot))}\}$ 

\hrulefill \\
\textbf{Inference} \\
\textbf{Input}: A finetuned CIL model  $ (\theta{(\cdot)}, f(\cdot)) = \{(\theta_1{(\cdot)}, f_1(\cdot)), (\theta_2{(\cdot)}, f_2(\cdot)), ..., (\theta_c{(\cdot), f_c(\cdot))}\}$  \\
\textbf{Data}:  A sequence of $c$ tasks OOD data $X^{ood} = \{X_1^{ood}, X_2^{ood},..., X_c^{ood}\} $; A sequence of $c$ tasks ID data $T = \{T_1, T_2,..., T_c\}$, $T_t = (X_t^{train}, X_t^{test}, Y_t)$
\begin{algorithmic}[1] 
\FOR{each task}
\STATE Obtain $t$-th $(1 \le t \le c)$ task testing ID data $T_t^{test} = X_1^{test} \cup X_2^{test} \cup ... \cup X_t^{test}$
\STATE Obtain $t$-th $(1 \le t \le c)$ task testing OOD data $X_t^{ood}$
\STATE Perform ID classification on original CIL model $\theta_t{(\cdot)} = \{ \phi_t{(\cdot)}, h_t(\cdot) \}$ based on $ T_t^{test} $
\STATE Perform different post-hoc OOD scoring function on finetuned classifier $\{ \phi_t{(\cdot)}, f_t(\cdot) \}$ based on $ T_t^{test} \cup X_t^{ood}$
\ENDFOR
\end{algorithmic}
\textbf{Output}: Average incremental accuracy; Average OOD detection performance 
\end{algorithm}

\begin{algorithm}[ht]
\caption{: Bi-directional Energy Regularization (BER)}
\label{alg:BER}
\textbf{Input}: A well-trained CIL model $\theta{(\cdot)} = \{\theta_1{(\cdot)}, \theta_2{(\cdot)},..., \theta_c{(\cdot)}\}$ \\
\textbf{Data}: A data memory $M$;  A sequence of $c$ tasks ID data $T = \{T_1, T_2,..., T_c\}$, $T_t = (X_t^{train}, X_t^{test}, Y_t)$
\begin{algorithmic}[1] 
\FOR{each task}
\STATE Obtain $t$-th $(1 \le t \le c)$ task training ID data $T_t^{train} = (X_t^{train}, Y_t) \cup M$
\STATE Obtain $t$-th CIL model $\theta_t{(\cdot)} = \{ \phi_t{(\cdot)}, h_t(\cdot) \}$ which is composed of a feature extractor $\phi_t{(\cdot)}$ and a original classifier $h_t(\cdot)$
\STATE Initialize an extra classifier $f_t(\cdot)$
\FOR{each iteration} 
\STATE Freeze the $\phi_t{(\cdot)}$ and $h_t(\cdot)$
\STATE Sample a mini-batch of current task ID training data $ \left\{ (x_t^i, y_t^i) \right\}_{i=1}^B $ from $  (X_t^{train}, Y_t) $
\STATE Separate the current task batch into two identical parts with equal size
\STATE Conduct mixup on the second part based on Eq.1
\STATE Apply energy regularization on these two parts based on Eq.2 with extra classifier $f_t(\cdot)$
\STATE Sample a mini-batch of old task ID training data $ \left\{ (x_o^i, y_o^i) \right\}_{i=1}^B $ $(1 \le o < t)$ from $ M $
\STATE Conduct mixup between the old task and current task ID training data based on Eq.3
\STATE Apply energy regularization on this mixed data based on Eq.4 with extra classifier $f_t(\cdot)$
\ENDFOR
\STATE Update data memory $M$ 
\STATE Save current CIL model $\theta_t{(\cdot)}$ with extra classifier $f_t(\cdot)$
\ENDFOR
\end{algorithmic} 
\textbf{Output}: A well-trained CIL model (the same as \textbf{Input}) with a finetuned classifier $ (\theta{(\cdot)}, f(\cdot)) = \{(\theta_1{(\cdot)}, f_1(\cdot)), (\theta_2{(\cdot)}, f_2(\cdot)), ..., (\theta_c{(\cdot), f_c(\cdot))}\}$ 
\end{algorithm}
\FloatBarrier

\end{sloppypar}
\end{document}